\pdfoutput=1
\documentclass[11pt]{article}

\usepackage[preprint]{acl}

\usepackage{times}
\usepackage{latexsym}

\usepackage[T1]{fontenc}
\usepackage[utf8]{inputenc}

\usepackage{inconsolata}
\usepackage{enumitem}
\usepackage{tabularx} 
\usepackage{arydshln}
\usepackage{amsmath}
\usepackage{diagbox}
\usepackage{caption}
\usepackage{amsfonts,amssymb}
\usepackage{subfigure}
\usepackage{graphicx}
\usepackage{subcaption}
\usepackage{bm}
\usepackage{multirow}
\usepackage[ruled,lined,commentsnumbered]{algorithm2e}
\usepackage{hyperref}
\usepackage{booktabs}
\usepackage{bbm}
\usepackage[inkscapelatex=false]{svg}
\usepackage{pifont}% http://ctan.org/pkg/pifont
\usepackage{stfloats}
\usepackage{multicol}
\usepackage{float}
\usepackage{array}
\usepackage{amssymb}
\newcommand{\PreserveBackslash}[1]{\let\temp=\\#1\let\\=\temp}
\newcolumntype{C}[1]{>{\PreserveBackslash\centering}p{#1}}
\newcolumntype{R}[1]{>{\PreserveBackslash\raggedleft}p{#1}}
\newcolumntype{L}[1]{>{\PreserveBackslash\raggedright}p{#1}}

\usepackage{mathrsfs}

\title{AlignX: Advancing Multilingual Large Language Models with Multilingual Representation Alignment}
\author{
    Mengyu Bu\textsuperscript{\rm 1,3},
    Shaolei Zhang\textsuperscript{\rm 1,3},
    Zhongjun He\textsuperscript{\rm 4},
    Hua Wu\textsuperscript{\rm 4},
    Yang Feng\textsuperscript{\rm 1,2,3}\footnotemark[2] \\
    \textsuperscript{\rm 1}{Key Laboratory of Intelligent Information Processing,} \\ Institute of Computing Technology, Chinese Academy of Sciences (ICT/CAS) \\
    \textsuperscript{\rm 2} {Key Laboratory of AI Safety, Chinese Academy of Sciences} \\
    \textsuperscript{\rm 3} {University of Chinese Academy of Sciences, Beijing, China} \\
    \textsuperscript{\rm 4} {Baidu Inc.} \\
    \texttt{\{\href{mailto:bumengyu23z@ict.ac.cn}{bumengyu23z}, \href{mailto:zhangshaolei20z@ict.ac.cn}{zhangshaolei20z}, \href{mailto:fengyang@ict.ac.cn}{fengyang}\}@ict.ac.cn}
}

\begin{document}
\maketitle

\renewcommand{\thefootnote}{\fnsymbol{footnote}} %将脚注符号设置为fnsymbol类型，即特殊符号表示
\footnotetext[2]{Corresponding author: Yang Feng.} %对应脚注[1]
% \footnotetext[1]{This paper was done when Shaolei Zhang studied at ICT/CAS.} %对应脚注[2]
\renewcommand{\thefootnote}{\arabic{footnote}}

\begin{abstract}

% 大语言模型具有强大的多语言理解与生成能力。然而，由于训练数据的不均衡分布，大语言模型的多语言能力往往bias到少数高资源语言上，其他语言上性能较差。一个常用解决方案是在大规模的、均衡的多语言数据集上微调，但此类通常方法开销大，并且存在灾难性遗忘问题。在本文中，我们从表示层面进行改进，提出AlignX框架：一种两阶段的多语言表示对齐框架。在第一阶段，我们在平行数据上微调基座模型，利用对比学习和语言匹配对齐多语言表示。在第二阶段，我们利用多语言混合指令微调激发模型的多语言能力。多个基座模型的实验结果表明，我们的方法增强了大模型的多语言理解与跨语言生成能力。进一步分析表明，AlignX拉近了大模型内部的多语言表示，提高了大模型的跨语言对齐水平。

% Multilingual large language models (LLMs) possess impressive multilingual understanding and generation capabilities, though their performance tends to be imbalanced across different languages.
Multilingual large language models (LLMs) possess impressive multilingual understanding and generation capabilities. However, their performance and cross-lingual alignment often lag for non-dominant languages.
% A common solution is to fine-tune LLMs on large-scale and more balanced multilingual corpora, but such approaches often come with high costs and struggle to achieve consistent improvements across languages.
A common solution is to fine-tune LLMs on large-scale and more balanced multilingual corpora, but such approaches often lead to imprecise alignment and suboptimal knowledge transfer, struggling with limited improvements across languages.
% In this paper, we mitigate the multilingual performance gap at the representation level and propose AlignX, a two-stage framework for aligning multilingual representations of pre-trained LLMs. 
In this paper, we propose AlignX to bridge the multilingual performance gap, which is a two-stage representation-level framework for enhancing multilingual performance of pre-trained LLMs. 
In the first stage, we align multilingual representations with multilingual semantic alignment and language feature integration. In the second stage, we stimulate the multilingual capability of LLMs via multilingual instruction fine-tuning. 
Experimental results on several pre-trained LLMs demonstrate that our approach enhances LLMs' multilingual general and cross-lingual generation capability. Further analysis indicates that AlignX brings the multilingual representations closer and improves the cross-lingual alignment.\footnote{The code is available at \url{https://github.com/ictnlp/AlignX}.}

\end{abstract}
\section{Introduction}

% 多语言大模型在大规模多语言上训练，因此有很强的多语言理解与生成能力。然而，多语言LLM的能力偏向高资源语言（主要是英语），在低资源语言上能力和跨语言对齐是inferior的。
Multilingual large language models (LLMs), trained on extensive multilingual corpora, demonstrate impressive capabilities across a wide range of NLP tasks~\cite{10.5555/3495724.3495883, touvron2023llama1, ustun-etal-2024-aya}. However, they still exhibit a strong language bias towards high-resource languages, predominantly English, resulting in inferior performance and cross-lingual alignment for other languages~\cite{qi-etal-2023-cross, chen2023breaking, zhu-etal-2024-multilingual, chua2024crosslingual}.

% 为缓解这一问题，当前的主流方法基于数据隐式注入跨语言对齐信息，例如在大规模多语言数据集上继续预训练、多语言通用指令微调、在翻译数据上做跨语言指令微调等等。但是这类方法对语义空间的影响是不可控的，不同语言存在不同的语义子空间，知识迁移效果不佳。
To alleviate this problem, current mainstream methods implicitly inject cross-lingual alignment information at data level, such as continual pre-training on large-scale multilingual corpora~\cite{cui2023efficient, yang2023bigtranslate, xu2024a}, multilingual general instruction fine-tuning~\cite{li2023bactrian, zhang-etal-2024-enhancing-multilingual}, and cross-lingual instruction fine-tuning on translation pairs~\cite{zhang2023bayling, zhu2023extrapolating}. Such data-level methods adjust multilingual semantic representations by implicitly injecting alignment information through translation pairs or large-scale multilingual corpora. However, the impact of such methods on the semantic space is uncontrollable and inefficient, often resulting in imprecise alignment and suboptimal knowledge transfer.

% 语言的一个经典假设是，对齐的多语言表示让知识迁移更容易。Wendler等人发现，英文中心的LLM以英文作为内部中心语言，这是LLM内部知识迁移的一个证据。我们进一步分析LLM逐层处理多语言数据的过程，发现了“对齐-then-发散”的处理多语言数据的机制（见图）。从低层到中间层，模型逐渐拉近多语言表示，促进知识共享；从中间层到高层，模型逐渐推远多语言表示，最终输出不同语言。按照这一假设和发现，一个直观的增强多语言能力的想法是，在中间层对齐多语言语义空间，促进知识共享。为避免单纯对齐干扰语言生成，在高层保持语言特征，实现精确的语言生成。
A classic assumption in multilingual NLP is that more consistent multilingual representations facilitate easier knowledge transfer~\cite{pan2021contrastive, tang2022align}. \citet{wendler-etal-2024-llamas} provide evidence by showing that English-centric LLMs internally pivot through English during processing.
To investigate this further, we explore how LLMs process multilingual data across layers, and reveal an align-then-diverge pattern (Figure \ref{visualization_selected_layer}). From the lower to intermediate layers, the model aligns multilingual representations to enable knowledge sharing, while from the intermediate to upper layers, it gradually diverges these representations to produce language-specific outputs. Building on this assumption and finding, an intuitive approach to enhancing multilingual capabilities is to align multilingual semantic spaces at the intermediate layer for better knowledge sharing, while preserving language-specific features in higher layers for accurate generation.

% 为实现这一目标，我们提出AlignX，一个两阶段、在表示层面增强预训练LLM多语言性能的框架。我们的方法利用翻译数据和多语言表示对齐增强多语言通用能力和跨语言生成能力。第一阶段，我们利用翻译指令数据进行继续预训练，高效向LLM注入对齐信息。我们在中间层进行语义对齐，促进知识共享，这通过指令对比学习任务实现；在输出层整合语言特征，更准确地输出目标语言，这通过语言匹配任务实现；此外，我们通过标准语言模型任务提供整体约束。第二阶段，我们用多语言指令微调模型，包括多语言翻译指令和多语言通用指令，从而激发通用能力、并保持第一阶段建立的对齐信息。
To achieve this, we propose AlignX, a two-stage and representation-level framework for enhancing the multilingual performance of pre-trained LLMs.
% To achieve this, we propose AlignX, a two-stage framework for aligning multilingual representations of pre-trained LLMs. (这里的“对齐”不准确)
% AlignX utilizes translation data and multilingual representation alignment to enhance multilingual general and cross-lingual generation capabilities. 
In the first phase, we efficiently align multilingual representations during continual pre-training. Specifically, we perform multilingual semantic alignment at the intermediate layer using the instruction contrastive learning task to promote knowledge sharing, and integrate language-specific features at the output layer via the language matching task to ensure accurate target language generation. In addition, we provide overall constraints through the standard language modeling task. In the second stage, we fine-tune the model with multilingual instruction data, comprising both translation and multilingual general instruction data, to stimulate the general capability while preserving the multilingual alignment information established in the first stage.

% 我们在5个预训练LLM上实验，并评估了5个广泛使用的、多语言通用和跨语言生成的benchmark。结果表明我们的方法有效增强了多语言通用能力和跨语言生成能力。此外，扩展到51种语言还能进一步提升性能，这表明增加对齐语言的数量可以进一步促进语言间知识共享。
We experiment on five pre-trained LLMs and evaluate performance across five widely used multilingual general and generation benchmarks. The results indicate that AlignX effectively enhances multilingual general and cross-lingual generation capabilities. Extending AlignX to 51 languages further improves performance, highlighting the benefits of increasing the number of aligned languages in enhancing knowledge sharing.

\begin{figure*}[t!]
  \centering
  \includegraphics[width=1.0\linewidth]{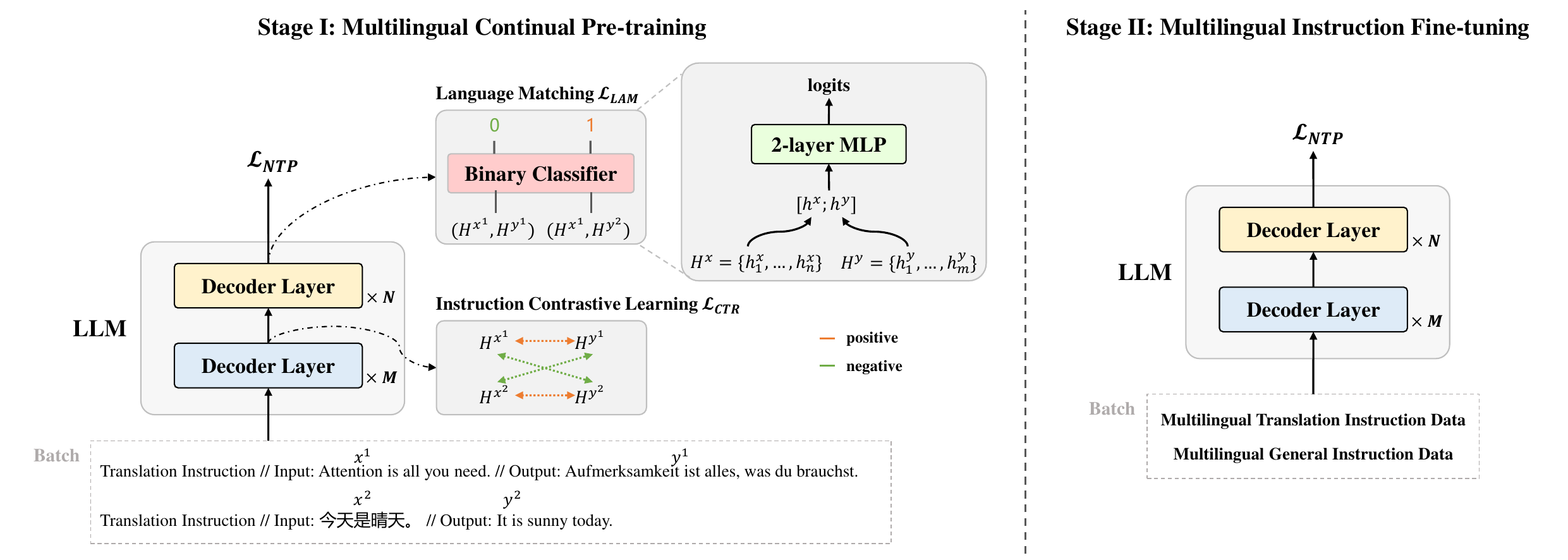}
  \caption{The framework of our AlignX, which consists of two stages. The first stage, multilingual continual pre-training, aligns multilingual representations within LLMs. The second stage, multilingual instruction fine-tuning, stimulates the multilingual capabilities of LLMs and maintains pre-established multilingual alignment.}
  \label{fig_framework}
\end{figure*}

% 我们在多个预训练LLMs上实验，包括以英语为中心的和非英语中心的模型。多语言翻译任务上的结果表明，和基线方法相比，我们的模型在训练语向和非训练语向上都获得了显著提升。考虑到我们使用的指令数据量和基线方法相近或显著更少，这表明表示层面的方法比数据层面的方法在提升跨语言生成能力上更高效。此外，我们还在多个多语言通用测试集上测试，获得XNLI上+1.78，m_Helleswag上+1.34，m_truthfulQA上+3.59的平均提升，证明AlignX提升了多语言通用能力。最新跨语言对齐评测框架表明，AlignX有效缓解了多语言能力不均衡的问题，提升了跨语言对齐水平。
% We experiment on five pre-trained LLMs and evaluate performance across five widely used multilingual general and generation benchmarks. The results indicate that AlignX effectively enhances multilingual general and cross-lingual generation capabilities. Extending AlignX to 51 languages further improves performance, highlighting the benefits of increasing the number of aligned languages in enhancing knowledge sharing. Further analysis reveals that AlignX brings the multilingual representations closer and enhances the cross-lingual alignment.

\section{Related Work}

\paragraph{Data-level Cross-lingual Alignment in Multilingual LLMs} Many works aim to enhance LLMs' multilingual capabilities through data-level approaches. Since translation pairs inherently contain language alignment information, researchers often use translation corpora to boost language proficiency~\cite{zhang2023bayling, alvestower, zhu-etal-2024-question}. \citet{xu2024a} first utilize massive monolingual data to enhance language generation and then leverage a small but high-quality translation dataset to inject alignment information. \citet{zhu2023extrapolating} leverage both multilingual translation instruction data and general task instruction data to build semantic alignment across languages. \citet{zhang-etal-2024-enhancing-multilingual} propose a self-distillation method that transfers high-resource language capabilities to enhance multilingual performance using translation and code-switched pairs. Compared to these approaches, our representation-level AlignX aligns cross-lingual representations more efficiently.

\paragraph{Representation-level Cross-lingual Alignment in Multilingual LLMs} Representation engineering provides a powerful lens for analyzing internal representations of LLMs~\cite{zhang2024truthx, yu2024truth}. Following this, many recent works investigate internal cross-lingual capability of LLMs~\cite{zhong2024beyond, zhao2025lens}. \citet{wendler-etal-2024-llamas} suggest that English-centric LLMs use English as an internal pivot language. Given the existence of an internal pivot language, it is intuitive that aligning multilingual representations facilitates more efficient knowledge transfer. \citet{li-etal-2024-improving-context} bridge the multilingual representation gap through multilingual contrastive learning and cross-lingual instruction tuning. \citet{li2024prealign} establish word-level multilingual alignment before pre-training and then pre-train on code-switched text. AlignX differs by performing multilingual semantic alignment at the intermediate layer and language feature integration at the output layer. This integration is crucial for preserving language-specific features and enhancing accurate language generation.

% \section{Preliminary Analysis: How LLMs Process Multilingual Data Layer by Layer?}
\section{Preliminary Analysis}

\begin{figure*}[htbp]
    \centering
    \subfigure[Layer 1]{\includegraphics[width=1.2in]{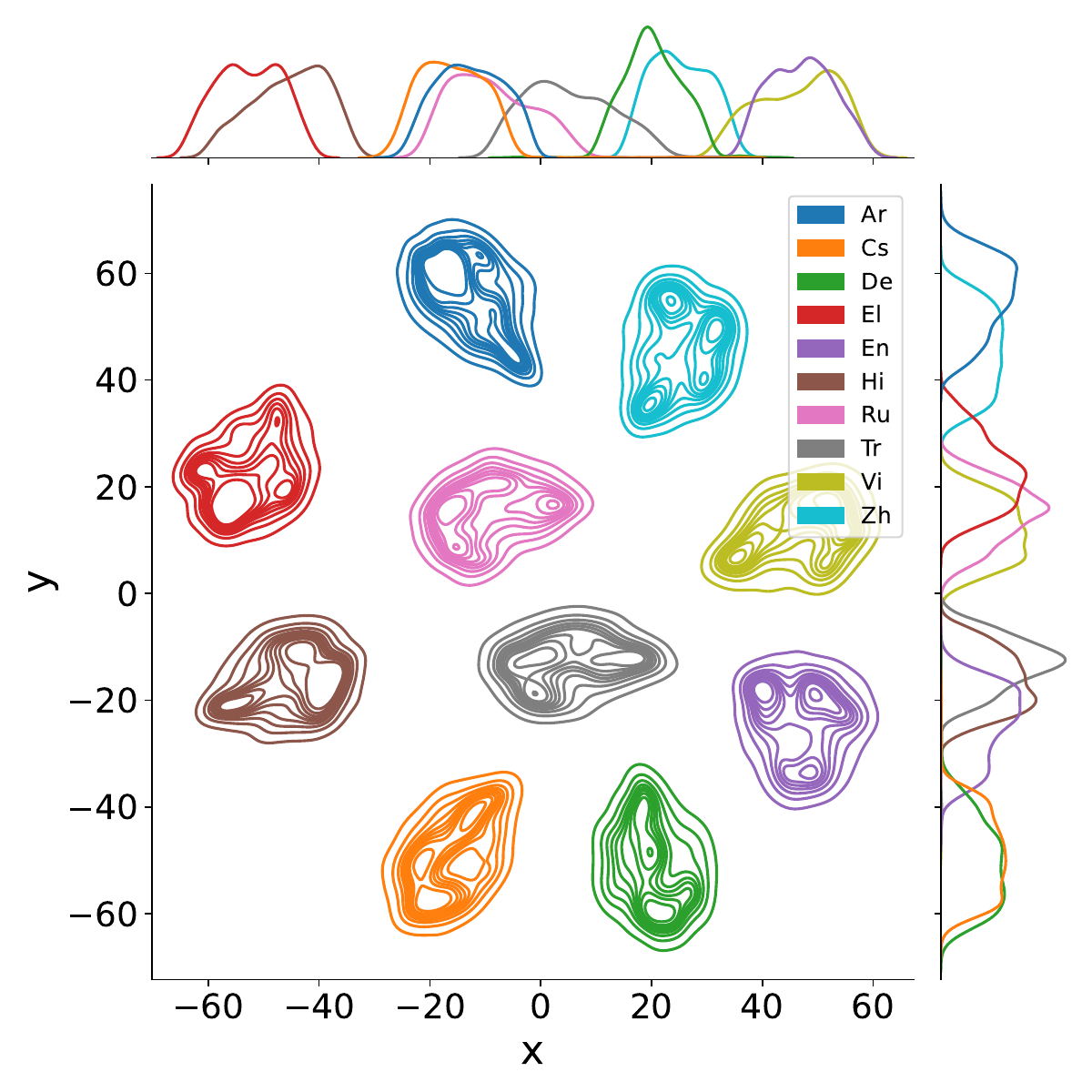}} 
    \subfigure[Layer 6]{\includegraphics[width=1.2in]{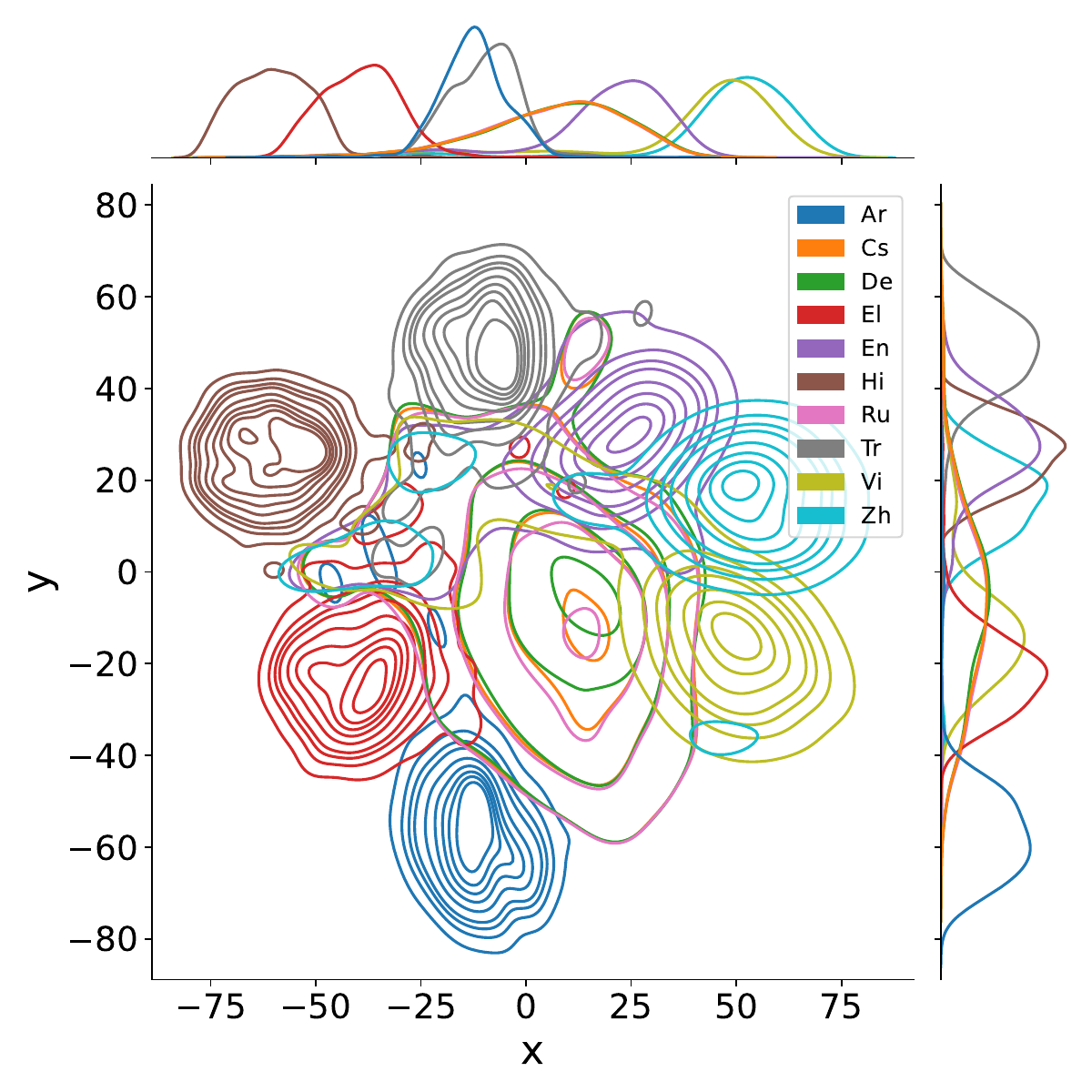}} 
    \subfigure[Layer 16]{\includegraphics[width=1.2in]{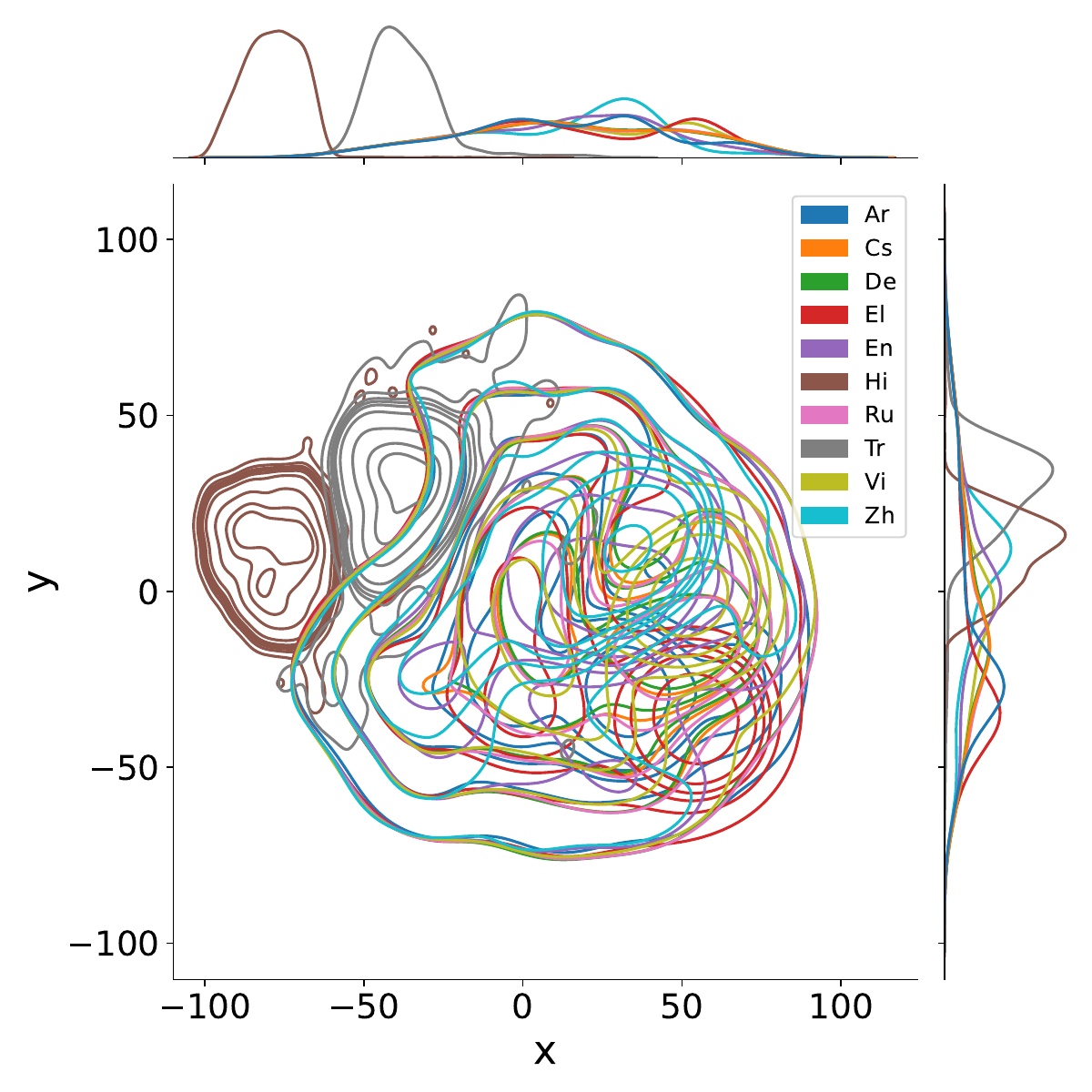}}
    \subfigure[Layer 20]{\includegraphics[width=1.2in]{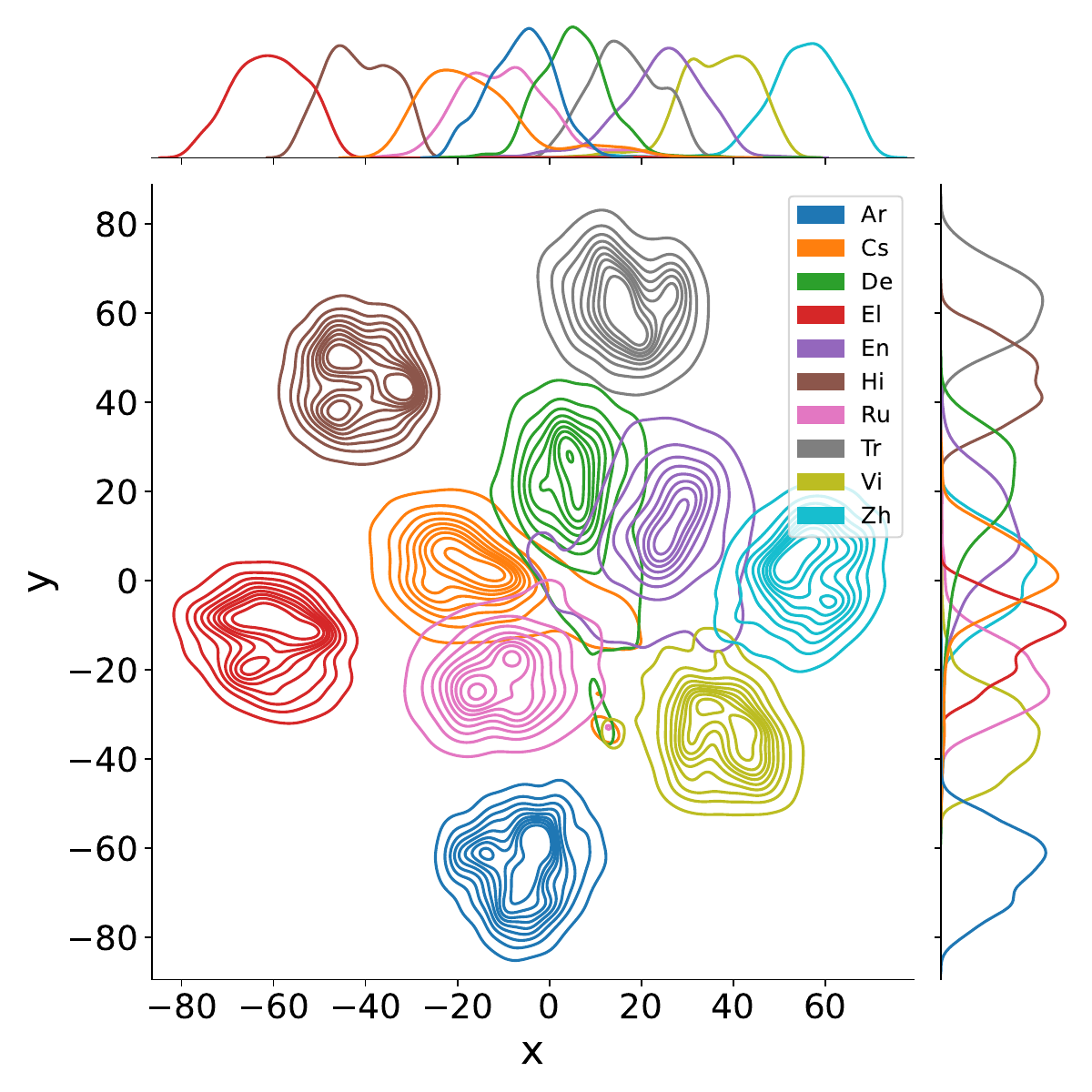}}
    \subfigure[Layer 32]{\includegraphics[width=1.2in]{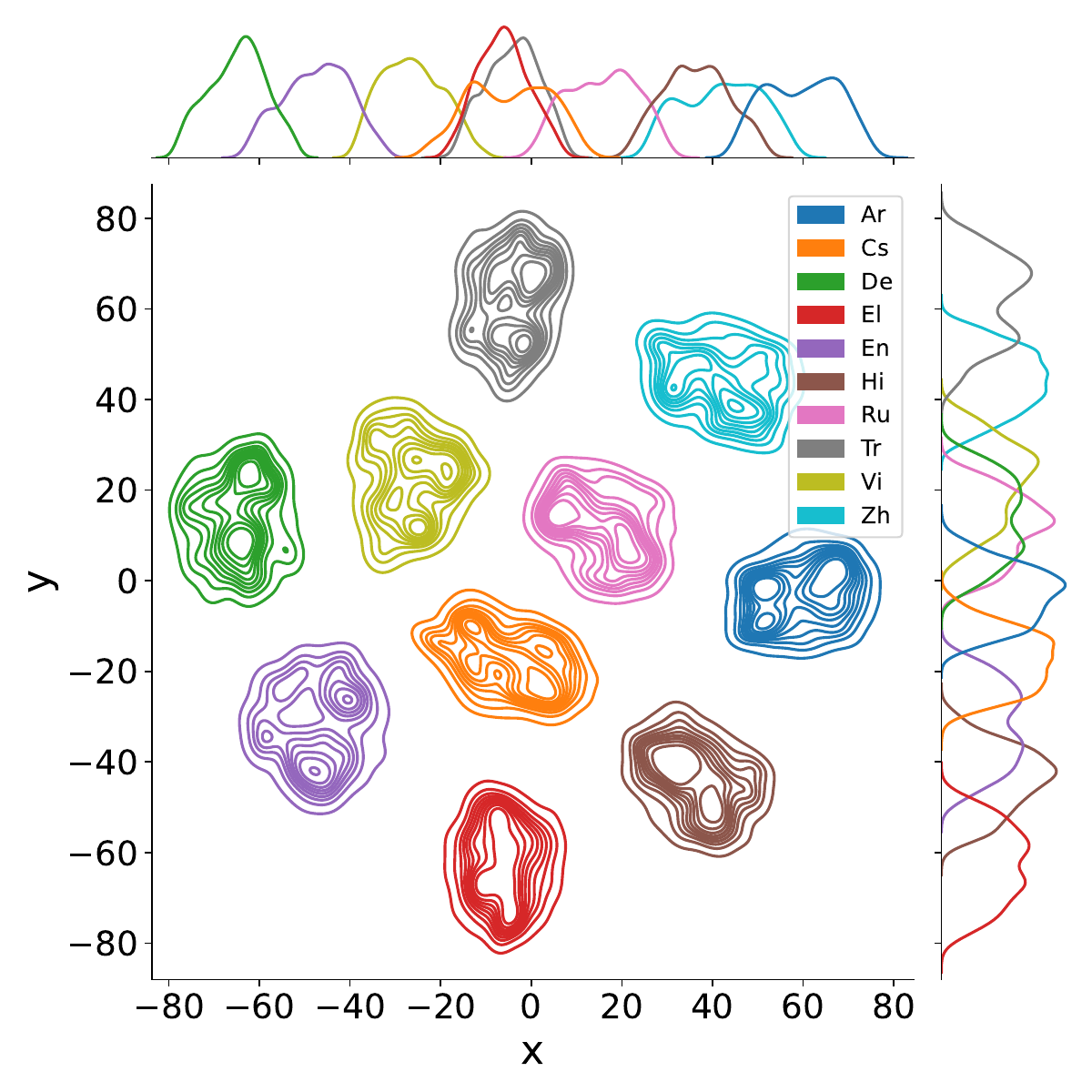}}
    \caption{Preliminary visualization of multilingual representations from some representative LLaMA3 layers after dimension reduction. We leverage the FLORES-101 dev set, which is multi-way parallel, and use average-pooled hidden states for dimension reduction. Appendix \ref{appendix_detailed_visualization} presents detailed visualization process and results.}
    \label{visualization_selected_layer}
\end{figure*}

We visualize the multilingual representations across different layers of LLaMA3 after dimension reduction. Specifically, we use the FLORES-101 dev set, perform mean-pooling on hidden states of different layers, and reduce dimension to 2-dim using t-SNE~\cite{van2008visualizing}. Figure \ref{visualization_selected_layer} presents the results of some representative layers, revealing a clear pattern: from the lower to intermediate layers, the LLM progressively aligns multilingual representations, facilitating knowledge transfer. After the intermediate layers, multilingual representations gradually diverge into distinct language-specific subspaces, ultimately leading to different language outputs. This reveals a pattern in how LLMs handle multilingual data: the LLM exhibits a natural tendency to process multilingual representations in an align-then-diverge manner, though the alignment remains suboptimal. Following this, we propose AlignX to strengthen this pattern.

\section{Method}

% To align multilingual representations within LLMs, we first explore how LLMs process multilingual data layer by layer. We discover that LLMs have a pattern of processing multilingual data in an aligned manner. 
To enhance this align-then-diverge pattern of LLMs, we propose AlignX, a two-stage and representation-level framework for enhancing the multilingual performance of pre-trained LLMs. AlignX contains two stages: 1) multilingual continual pre-training, which explicitly aligns multilingual representations via multilingual semantic alignment and language feature integration; and 2) multilingual instruction fine-tuning, which combines multilingual general instruction data with multilingual translation instruction data to stimulate general capabilities while preserving the pre-established alignment. Figure \ref{fig_framework} illustrates the framework of AlignX.

% \subsection{Preliminary Analysis: How LLMs Process Multilingual Data Layer by Layer?}

% \input{results/visualization_base_layer}

% We visualize the multilingual representations at different layers of LLaMA3 after dimension reduction. We use the FLORES-101 dev set, perform mean-pooling on hidden states of different layers, and apply t-SNE~\cite{van2008visualizing} for dimension reduction to 2-dim. Figure \ref{visualization_base_layer} indicates that, from low layers to the intermediate layer, the LLM progressively brings multilingual representations closer, after which the representations gradually diverge. This reveals a pattern in how LLMs handle multilingual data: the LLM exhibits a natural tendency to process multilingual representations in an align-then-diverge manner, though the alignment remains sub-optimal. Following this, we propose AlignX to strengthen this pattern.

% \subsection{Multilingual Representation Alignment}
\subsection{Multilingual Continual Pre-training}

In the first stage, we conduct multilingual continual pre-training and efficiently align multilingual representations through multilingual semantic alignment and language feature integration, guided by multilingual instruction contrastive learning $\mathcal{L}_{CTR}$ and language matching learning $\mathcal{L}_{LAM}$, respectively. This stage leverages English-centric translation instruction data.
% , both of which have been proven effective in aligning representations in prior studies~\cite{pan2021contrastive, Radford2021LearningTV, 10.5555/3618408.3619222}.

% \paragraph{Instruction Contrastive Learning} 
\paragraph{Multilingual Semantic Alignment}
Our preliminary visualization reveals that LLMs tend to align multilingual representations during processing. To enhance this trend, we explicitly introduce a contrastive learning task at the intermediate layer, encouraging the model to produce similar representations for translation pairs.

Formally, given the dataset $\mathcal{D}$ and a translation instruction data $I^i=\{\mathbf{x}^i, \mathbf{y}^i\}$, we compute its $l$-th layer hidden states in model $f(\theta)$ as follows:
\begin{equation}
    f_l(I^i) = \{f_l(I^i)_1, \dots, f_l(I^i)_n\}
\end{equation}

We extract the hidden states of $\mathbf{x}^i$ and $\mathbf{y}^i$ based on their respective token index ranges and get $I^i_{\mathbf{x}^i}$ and $I^i_{\mathbf{y}^i}$, and subsequently apply mean pooling to obtain the sentence representation $h^{\mathbf{x}^i}$ and $h^{\mathbf{y}^i}$:
\begin{equation}
    h^{\mathbf{x}^i} = g(f_l(I^i_{\mathbf{x}^i})), h^{\mathbf{y}^i} = g(f_l(I^i_{\mathbf{y}^i}))
\end{equation}
where $g(\cdot)$ denotes mean pooling operation.

We take $(\mathbf{x}^i, \mathbf{y}^i)$ as the positive example and randomly select $\mathbf{y}^j$ to form the negative example $(\mathbf{x}^i, \mathbf{y}^j)$. Then, the objective of multilingual instruction contrastive learning is:
\begin{equation}
    % \resizebox{1.0\hsize}{!}{$
        \mathcal{L}_{CTR} = -\mathop{\mathbb{E}}\limits_{\mathbf{x}^i, \mathbf{y}^i \in \mathcal{D}} \log \frac{e^{sim(h^{\mathbf{x}^i}, h^{\mathbf{y}^i})/\tau}}{\sum_{\mathbf{y}^j} e^{sim(h^{\mathbf{x}^i}, h^{\mathbf{y}^j})/\tau}}
    % $}
\end{equation}
where $sim(\cdot)$ calculates the similarity of different sentences, and we use cosine similarity in our work. $\tau$ is a temperature hyper-parameter. To simplify the implementation, we sample negative examples within every training batch.

% 直觉上，在中间层对齐多语言语义空间后，模型输出时更难区分语言。受BLIP-2工作启发，我们引入一个语言匹配任务对齐输出语言。具体来说，我们训练一个匹配分类器，预测给定句对表示是否属于同种语言。分类器输入最后一层隐状态，利用一个二分类任务进行优化。
% \paragraph{Language Matching}
\paragraph{Language Feature Integration}
Intuitively, after aligning the multilingual semantic space in the intermediate layer, it becomes more difficult for the model to distinguish between output languages. To bridge this gap, we leverage a language matching task to inject language features. Specifically, we train a language matching classifier that predicts whether a given sentence pair is in the same language (matched) or different languages (unmatched). The classifier takes the hidden states from the model's final layer as input and is optimized using a binary classification task.

% 形式上，给定平均输出的句子表示，我们拼接它们并输入匹配分类器得到logits。ground truth是x和y是否属于同一语言，用y表示。语言匹配任务用CE损失优化。
Formally, given mean-pooled sentence representations $h^{\mathbf{x}}$ and $h^{\mathbf{y}}$, we concatenate and feed them into the matching classifier to get a logit. The ground truth is whether $\mathbf{x}$ and $\mathbf{y}$ belong to the same language, denoted as $y$. The language matching task is optimized with the cross-entropy loss:
\begin{equation}
    \mathcal{L}_{LAM} = -\mathop{\mathbb{E}}\limits_{\mathbf{x}, \mathbf{y}\in \mathcal{D}} \log M([h^{\mathbf{x}};h^{\mathbf{y}}], y)
\end{equation}
where $M(\theta)$ denotes the language matching classifier, a 2-layer MLP in our work. To simplify the implementation, we sample sentence pairs within every training batch.

\paragraph{Final Loss} Finally, we optimize the representation alignment objectives $\mathcal{L}_{CTR}$ and $\mathcal{L}_{LAM}$ with the standard next token prediction loss of LLMs, using multilingual translation instruction data. The next token prediction loss $\mathcal{L}_{NTP}$ is:
\begin{equation}
    \mathcal{L}_{NTP} = -\mathop{\mathbb{E}}\limits_{I^i \in \mathcal{D}} \sum_j \log P(I^i_j|I^i_{<j})
\label{equat_NTP}
\end{equation}
The final loss of multilingual continual pre-training stage is:
\begin{equation}
    \mathcal{L} = \mathcal{L}_{NTP} + \alpha_1\mathcal{L}_{CTR} + \alpha_2\mathcal{L}_{LAM}
\end{equation}
where $\alpha_1$ and $\alpha_2$ are hyper-parameters to balance the losses.

\subsection{Multilingual Instruction Fine-tuning}

% 多语言表示对齐后，我们利用多语言混合指令微调，来激发通用能力并保持一阶段建立的对齐。我们混合多语言翻译指令和多语言通用指令，前者隐式提供对齐信息，后者高效激发通用能力。
After multilingual representation alignment, we leverage multilingual instruction fine-tuning to effectively stimulate general capabilities while maintaining the alignment established in the first stage. We mix multilingual translation instruction data, which implicitly provides alignment information~\cite{zhu2023extrapolating}, and multilingual general instruction data, which efficiently stimulates multilingual general capabilities~\cite{chen-etal-2024-monolingual}. To ensure diversity and multilingual balance, we sample English-centric translation pairs uniformly across languages, pairing each with a randomly selected translation instruction. General instruction data are also sampled uniformly across languages. Empirically, we maintain a 1:3 ratio of translation data to general instruction data and a 1:5 ratio of second-stage data to first-stage data. In this stage, we optimize the model with the next token prediction loss in Equation \eqref{equat_NTP}.

\section{Experiment}

\begin{figure*}[t!]
    \centering
    % \subfigure[Multilingual TruthfulQA]{\includegraphics[width=0.85\linewidth]{images/m_truthfulqa.pdf}\label{a}} \\
    % \subfigure[Multilingual HellaSwag]{\includegraphics[width=0.85\linewidth]{images/m_hellaswag.pdf}\label{b}} \\
    % \subfigure[XNLI]{\includegraphics[width=0.85\linewidth]{images/xnli.pdf}\label{c}} \\
    % \subfigure[XStoryCloze]{\includegraphics[width=0.85\linewidth]{images/xstorycloze.pdf}\label{d}}
    \includegraphics[width=1.0\linewidth]{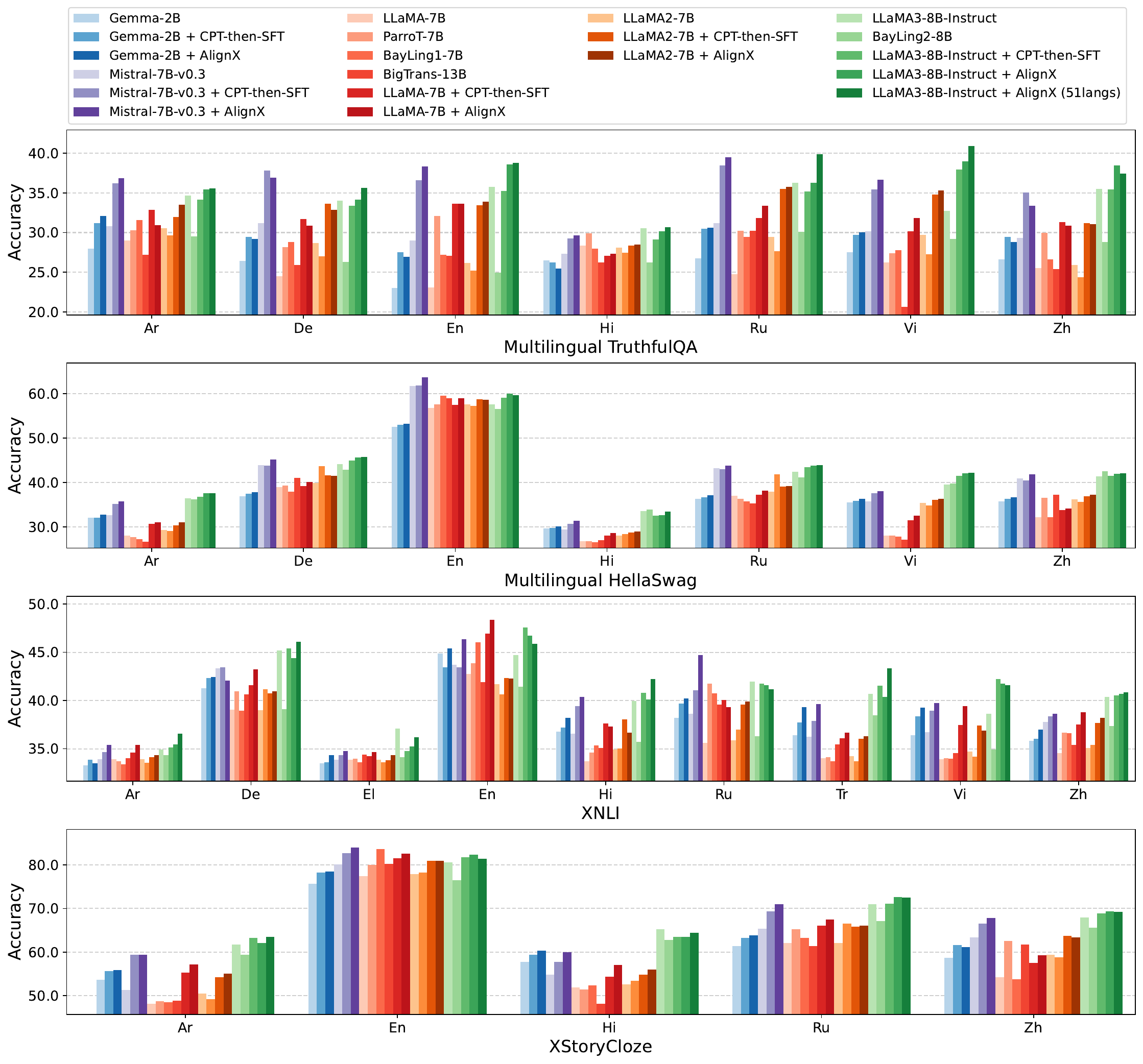}
    \caption{Overall performance on multilingual general benchmarks. All the benchmarks are evaluated using Accuracy metric. For comparison purposes, models with the same base LLM have the same color scheme, with the lightest being the base LLM and the darkest being AlignX.}
    \label{figure_main_general}
\end{figure*}

\begin{table*}[t]
\centering
\small
\begin{tabular}{cccccccccccc} \hline
\centering \textbf{Methods}              & \textbf{X-En}  & \textbf{X-Ar}  & \textbf{X-Cs}  & \textbf{X-De}  & \textbf{X-El}  & \textbf{X-Hi}  & \textbf{X-Ru}  & \textbf{X-Tr}  & \textbf{X-Vi}  & \textbf{X-Zh}  & \textbf{Avg.}   \\ \hline
\multicolumn{12}{c}{\textbf{Gemma-2B Based}}                                                                        \\ \hline
\textbf{Gemma-2B}           & 30.46 & 3.77 & 6.39 & 8.53 & 4.67 & 3.41 & 4.84 & 4.11 & 8.22 & 7.19 & 8.16 \\
\textbf{CPT-then-SFT}           & 31.62 & \textbf{7.36} & 7.94 & 13.40 & \textbf{8.99} & 7.16 & \textbf{11.05} & 7.24 & 14.01 & 10.55 & 11.93   \\
\textbf{AlignX}          & \textbf{31.68} & 6.94 & \textbf{8.75} & \textbf{13.84} & 8.83 & \textbf{7.40} & 10.63 & \textbf{7.94} & \textbf{15.87} & \textbf{11.02} & \textbf{12.29} 
 \\ \hline
 \multicolumn{12}{c}{\textbf{Mistral-7B-v0.3 Based}}                                                                        \\ \hline
\textbf{Mistral-7B-v0.3} & 30.96 & 3.89 & 14.29 & \textbf{17.88} & 2.26 & 3.49 & 15.68 & 5.78 & 9.71 & 10.50 & 11.44 \\
\textbf{CPT-then-SFT} & \textbf{34.29} & \textbf{8.13} & \textbf{15.05} & 17.35 & \textbf{6.84} & 6.17 & \textbf{16.88} & 9.94 & 13.86 & 12.56 & \textbf{14.11}   \\
\textbf{AlignX} & 33.85 & 8.00 & 14.58 & 16.54 & 6.77 & \textbf{6.74} & 16.59 & \textbf{10.34} & \textbf{14.18} & \textbf{12.78} & 14.04 \\ \hline
\multicolumn{12}{c}{\textbf{LLaMA-7B Based}}                                                                        \\ \hline
\textbf{LLaMA-7B}           & 20.81 & 0.74  & 5.70  & 9.05  & 0.80  & 0.97  & 6.60  & 1.22  & 1.10  & 1.60  & 4.86  \\
\textbf{ParroT-7B}          & 17.91 & 0.22 & 3.28 & 4.65 & 0.45 & 0.26 & 1.86 & 1.59 & 1.67 & 1.81 & 3.37 \\
\textbf{BayLing1-7B}           & 21.64 & 0.59  & 5.09  & 5.67  & 0.78  & 0.54  & 3.83  & 1.78  & 1.87  & 3.62  & 4.54  \\
\textbf{BigTrans-13B}           & 21.30 & 2.10 & 5.27 & 5.31 & 1.38 & \textbf{4.03} & 4.66 & 3.80 & 2.54 & 4.95 & 5.53 \\
\textbf{CPT-then-SFT}           & \textbf{27.02} & 3.21 & \textbf{9.38} & \textbf{11.90} & 3.34 & 2.66 & \textbf{10.48} & 4.79 & 6.26 & \textbf{5.64} & 8.47  \\
\textbf{AlignX}             & 25.46 & \textbf{3.85} & 9.01 & 11.39 & \textbf{4.51} & 3.36 & 9.96 & \textbf{5.60} & \textbf{7.84} & 5.30 & \textbf{8.63} 
  \\ \hline
\multicolumn{12}{c}{\textbf{LLaMA2-7B Based}}                                                                       \\ \hline
\textbf{LLaMA2-7B}          & 25.74 & 1.37  & 8.99  & 12.28 & 1.27  & 1.62  & 8.68  & 2.41  & 9.69  & 5.48  & 7.76  \\
\textbf{Tower-7B} & 30.48  & 1.51  & 7.74  & 12.75  & 1.33  & 1.79  & 10.15  & 2.26  & 8.56  & 9.26  & 8.58 \\
\textbf{CPT-then-SFT}          & 31.61 & 3.46 & \textbf{13.50} & \textbf{17.73} & 2.74 & 2.71 & 14.39 & 5.37 & 13.53 & 10.52 & 11.55    \\
\textbf{AlignX}             & \textbf{31.99} & \textbf{5.06}  & 13.44 & 17.62 & \textbf{4.22}  & \textbf{4.17}  & \textbf{15.19} & \textbf{6.30}  & \textbf{15.55} & \textbf{11.64} & \textbf{12.52} \\ \hline
\multicolumn{12}{c}{\textbf{LLaMA3-8B-Instruct Based}}                                                              \\ \hline
\textbf{LLaMA3-8B-Instruct} & 34.52 & 14.16 & 19.40 & 22.99 & 16.45 & 16.42 & 21.18 & 16.37 & 24.65 & 17.04 & 20.32 \\
\textbf{BayLing2-8B} & 35.21 & 13.83 & 15.09 & 18.33 & 14.26 & 16.02 & 18.14 & 14.09 & 22.41 & 15.94 & 18.33 \\
\textbf{CPT-then-SFT} & 36.50 & 14.82 & 19.68 & 23.98 & \textbf{17.82} & 16.09 & 21.52 & 15.26 & 24.65 & 19.08 & 20.94  \\
\textbf{AlignX}             & 37.36 & 14.98 & 19.66 & 23.58 & 17.73 & 16.26 & 21.31 & 14.72 & 23.51 & 18.40 & 20.75\\ 
\textbf{AlignX (51langs)}& \textbf{38.70} & \textbf{16.12} & \textbf{20.71} & \textbf{24.58} & 17.63 & \textbf{17.77} & \textbf{22.00} & \textbf{16.90} & \textbf{25.88} & \textbf{19.08} & \textbf{21.94} \\ \hline
\end{tabular}
% \caption{The overall performance on FLORES-101 benchmark. We report averaged BLEU~\cite{PapineniRWZ02} scores for each target language. "X" denotes all other training languages except the target language. "Avg." denotes average scores on all translation directions. The "AlignX (51langs)" model is fine-tuned under the 51-language setting, and evaluated under the 10-language setting for comparison. We bold the highest scores.}
\caption{The overall performance on the FLORES-101 benchmark. In each model group, we fine-tune the base LLM (first row) to obtain CPT-then-SFT and AlignX. We report averaged BLEU scores for each target language. "X" denotes all other nine languages except the target language. "Avg." denotes average scores on all translation directions. We bold the highest scores.}
\label{table_main_translation}
\end{table*}

% In this section, we present the experiment setup and the main results, including performance on multilingual general and generation benchmarks.

\subsection{Experiment Setup}
% base LLM, baselines, training dataset, evaluation benchmarks, configuration
\paragraph{Base LLMs} We leverage LLaMA-7B~\cite{touvron2023llama1}, LLaMA2-7B~\cite{touvron2023llama2}, LLaMA3-8B-Instruct~\cite{dubey2024llama}, Gemma-2B~\cite{team2024gemma} and Mistral-7B-v0.3~\cite{jiang2023mistral} as base LLMs.

\paragraph{Language Setup}
% In the main experiment, we focus on ten languages: Arabic (Ar), Czech (Cs), German (De), Greek (El), English (En), Hindi (Hi), Russian (Ru), Turkish (Tr), Vietnamese (Vi) and Chinese (Zh). These languages span diverse language families and range from low- to high-resource. For this 10-language setup, we refer to our method as AlignX. To evaluate the scalability of our approach, we further extend our experiments to a 51-language setup using the LLaMA3-8B-Instruct model and denote this configuration as AlignX (51langs), to distinguish it from the original 10-language setup.
In the main experiment, we focus on ten languages: Arabic (Ar), Czech (Cs), German (De), Greek (El), English (En), Hindi (Hi), Russian (Ru), Turkish (Tr), Vietnamese (Vi) and Chinese (Zh), spanning diverse families and resource levels, and refer to this setup as \textbf{AlignX}. To validate scalability, we extend to 51 languages using LLaMA3-8B-Instruct, denoting this configuration as \textbf{AlignX (51langs)} to distinguish it from the original 10-language setup.

\paragraph{Baselines} We compare with these baselines: (1) \textbf{CPT-then-SFT} follows the same two-stage multilingual instruction tuning as AlignX but without $\mathcal{L}_{CTR}$ and $\mathcal{L}_{LAM}$; (2) \textbf{ParroT-7B}~\cite{jiao-etal-2023-parrot} leverages chat data to improve translation ability; (3) \textbf{BayLing1-7B}~\cite{zhang2023bayling} and \textbf{BayLing2-8B}~\cite{zhang2024bayling} use multi-turn interactive translation instruction data for cross-lingual alignment; (4) \textbf{BigTrans-13B}~\cite{yang2023bigtranslate} applies continual pre-training on a large-scale multilingual corpus; (5) \textbf{Tower-7B}~\cite{alvestower} first trains on multilingual data then fine-tunes on translation instruction data.
% ParroT-7B and BayLing1-7B are fine-tuned on LLaMA-7B, BigTrans-13B is fine-tuned on LLaMA-13B, and BayLing2-8B is fine-tuned on LLaMA3-8B-Instruct.

\paragraph{Training Dataset} We construct the multilingual translation instruction data from OPUS-100~\cite{zhang-etal-2020-improving} and extract multilingual general instruction data from Bactrian-X~\cite{li2023bactrian}. Appendix \ref{appendix_data_processing} and \ref{appendix_statistics_on_corupus_size_and_language} present details about data processing and data statistics, respectively. 
% We conduct experiments across multiple language group setups, including 51, 10, 5, and 3 languages. These languages span diverse language families and range from low- to high-resource languages. 
% For multilingual translation instruction data, we randomly select English-centric translation pairs from the OPUS-100~\cite{zhang-etal-2020-improving} corpora and sample a translation instruction from the translation instruction set for each pair. For multilingual general instruction data, we primarily draw from the Bactrian-X~\cite{li2023bactrian} dataset, a multilingual version of the Alpaca~\cite{alpaca} dataset available in 52 languages\footnote{Since Greek (El) is not included in Bactrian-X, we obtain the Greek Alpaca dataset from \url{https://github.com/NJUNLP/x-LLM}.}. The instructions are translated using Google Translate, and the responses are generated with GPT-3.5-Turbo, with off-target responses filtered out using \textit{langid} toolkit~\cite{lui-baldwin-2012-langid}. In the first stage, we sample 50k translation pairs per direction, resulting in 0.9M translation instruction data in total. In the second stage, we sample 2.5k translation instruction data per direction and 10k general instruction data per language, resulting in 145k mixed instruction data in total. Detailed information for our training dataset is illustrated in Appendix \ref{appendix_statistics_of_dataset}.

\paragraph{Evaluation Benchmarks} We evaluate multilingual general and cross-lingual generation capabilities on five benchmarks: multilingual HellaSwag~\cite{zellers2019HellaSwag}, multilingual TruthfulQA~\cite{lin-etal-2022-truthfulqa}, XNLI~\cite{conneau2018xnli}, XStoryCloze~\cite{lin2021few}, evaluated with Accuracy metric; and FLORES-101~\cite{costa2022no}, evaluated with BLEU~\cite{PapineniRWZ02} and COMET~\cite{rei-etal-2020-comet} metrics. See Appendix \ref{appendix_evaluation_benchmarks} for details.
% \begin{itemize}[leftmargin=0pt]
%     \item \textbf{Cross-lingual Natural Language Inference (XNLI)} ~\cite{conneau2018xnli} This assesses multilingual reasoning capability.
%     \item \textbf{Multilingual HellaSwag} ~\cite{zellers2019HellaSwag} This assesses commonsense reasoning and contextual understanding capabilities.
%     \item \textbf{Multilingual TruthfulQA} ~\cite{lin-etal-2022-truthfulqa} This assesses knowledge and truthfulness capabilities.
%     \item \textbf{FLORES-101} ~\cite{costa2022no} This is a multilingual machine translation benchmark and assesses the cross-lingual generation capability.
% \end{itemize}
% The multilingual HellaSwag and TruthfulQA benchmarks are obtained from Okapi~\cite{lai-etal-2023-okapi}, translated by ChatGPT. We evaluate the trained languages.

% language matching classifier structure, training details (optimizer, lr, epoch, hyper-params, bsz), evaluation framework (MMT, lm-evaluation-harness)
\paragraph{Configuration} In AlignX, the language matching classifier is a 2-layer MLP with an intermediate dimension of 128 and an output dimension of 2. For training, we optimize using AdamW optimizer with a learning rate of 2e-6, training for 2 epochs per stage with a batch size of 128. We empirically set $\alpha_1=0.3$, $\alpha_2=0.4$, and $\tau=0.1$. For evaluation, we use the \textit{MMT-LLM} framework~\cite{zhu-etal-2024-multilingual} for FLORES-101 and the \textit{lm-evaluation-harness} framework~\cite{eval-harness} for other general benchmarks. All tasks are evaluated in a 1-shot setup.

\subsection{Main Results}
% \subsection{Main Results on Multilingual General Benchmarks}
\paragraph{AlignX improves multilingual general capability.}
% 图1展示了XNLI、multilingual Hellaswag、multilingual TruthfulQA上的结果。尽管这些benchmark在我们的训练分布之外，但AlignX仍然在LLaMA-7B,LLaMA2-7B,LLaMA3-8B-Instruct上取得一致的提升，这表明AlignX的通用性。
Figure \ref{figure_main_general} shows the results on multilingual TruthfulQA, multilingual Hellaswag, XNLI, and XStoryCloze, with detailed results in Appendix \ref{appendix_multilingual_general}. While these benchmarks are out of our training distribution, AlignX achieves improvements on all five base LLMs, demonstrating the generalizability of AlignX.
% 相比之下，data-level的方法对通用性能产生不同程度的影响。具体来说，BayLing-7B在主要训练语言（En，Zh，De）上获得显著提升，而其他语言表现和base模型相近或略好，这证明混合多语言翻译指令和通用指令的训练方式能够同时增强翻译能力和通用能力，但data-level方法的效果依赖于数据分布。BigTrans-13B则在多个语言上表现出严重的遗忘问题，这表明在大规模平行语料上继续预训练和小规模翻译指令微调的训练方式学习的跨语言对齐是task-specific的，难以泛化到其他任务上。结果表明，多语言表示对齐对提升多语言能力和泛化性有重要作用，并且多语言混合指令微调有效激发通用能力。
% Specifically, BayLing1-7B shows significant improvement in its primary training languages (English, Chinese, German), while performance in other languages remains comparable to or slightly better than the base LLM. This demonstrates that mixing multilingual translation instruction data and general instruction data enhances multilingual general capabilities, though the effect is influenced by the data distribution. 
% BigTrans-13B, on the other hand, suffers from severe forgetting across multiple languages, including its primary training language, Chinese. Another data-level approach ParroT-7B encounters similar problems. This suggests that the cross-lingual alignment learned through continual pre-training on massive parallel datasets, followed by instruction fine-tuning on small-scale translation datasets, is task-specific and struggles to generalize across other tasks. 
% The results demonstrate that multilingual representation alignment is crucial for enhancing multilingual capabilities and generalizability, while multilingual instruction fine-tuning effectively stimulates general capabilities.
In contrast, data-level methods exhibit varying degrees of forgetting across languages, indicating that cross-lingual alignment achieved solely from continual pre-training on large parallel datasets struggles to generalize beyond translation tasks. 
These findings demonstrate the importance of multilingual representation alignment for enhancing multilingual capability and generalizability.

% \subsection{Main Results on Multilingual Translation Benchmark}
\paragraph{AlignX significantly enhances cross-lingual generation capability.}
% 表1展示了FLORES-101测试集上的结果。为展示方便，我们对同一目标语言的得分做平均，细节的翻译结果见附录。结果表明，AlignX在3个基座上获得了最高分数，在LLaMA-7B, LLaMA2-7B, LLaMA3-8B-Instruct上分别获得平均+3.96，+5.76，+0.43BLEU分数。这表明我们方法的通用性，对多语言能力较弱和较强的基座模型均能增强，并且包括base模型和指令微调后的模型。
Table \ref{table_main_translation} presents results on the FLORES-101 benchmark, with statistical significance test, COMET scores, and additional details in Appendix \ref{appendix_detailed_results} and a case study in Appendix \ref{appendix_case_study}. The results show that AlignX achieves the highest scores on all base LLMs, with average +4.13, +2.6, +3.77, +4.76, and +0.43 BLEU scores on Gemma-2B, Mistral-7B-v0.3, LLaMA-7B, LLaMA2-7B, and LLaMA3-8B-Instruct, respectively. 
This demonstrates AlignX's versatility, as it improves across LLMs with varying multilingual capabilities.
Notably, BigTrans-13B uses a larger base LLM, 300M translation pairs, and some non-English translation directions (e.g., Hi-Zh). In contrast, AlignX relies solely on less than 1M English-centric translation pairs yet still achieves higher translation scores, indicating that representation alignment effectively enhances LLMs' cross-lingual capability.

\begin{figure*}[t!]
    \centering
    \subfigure[Multilingual General Benchmarks]{\includegraphics[height=6.8cm]{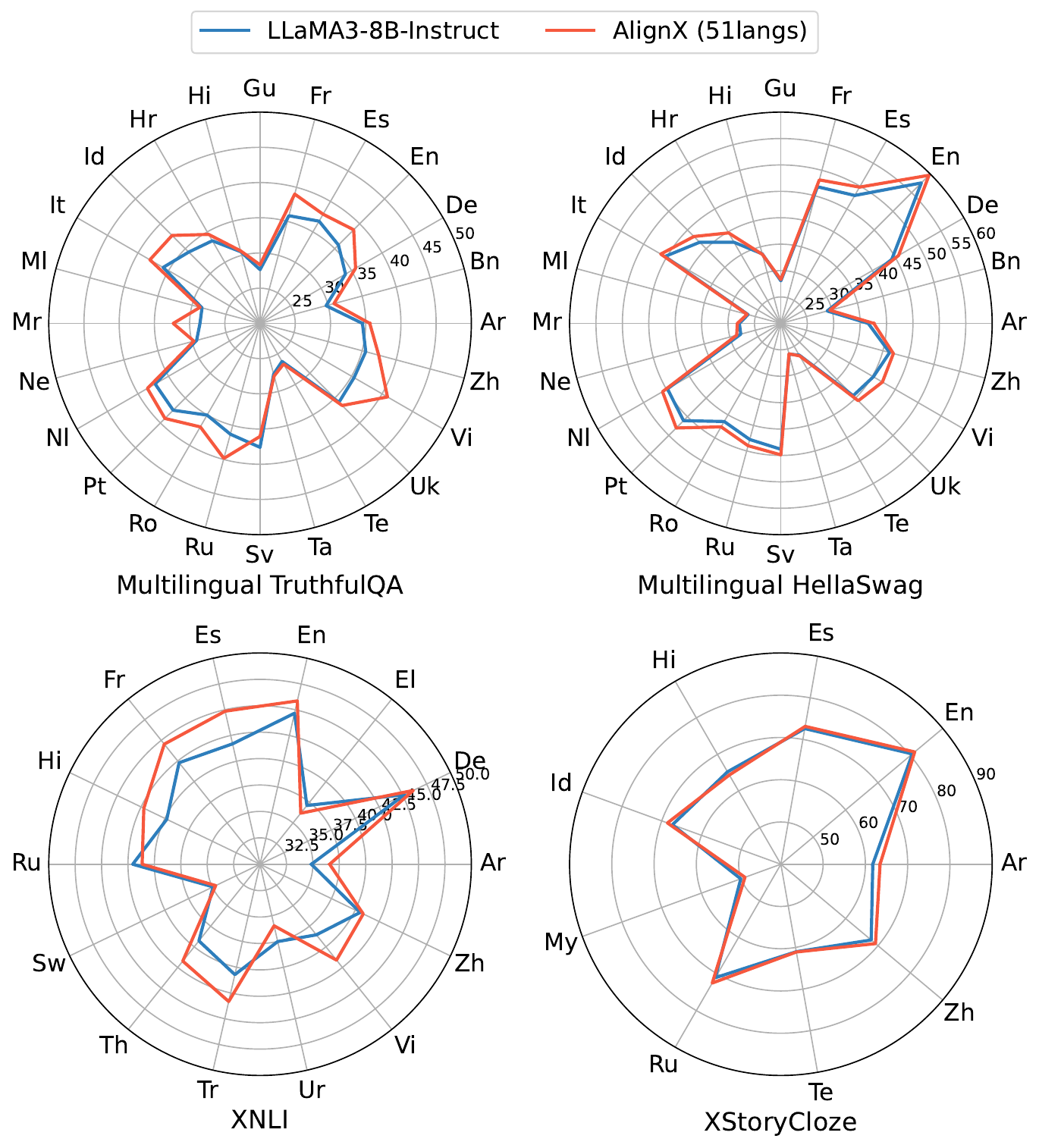}}
    \subfigure[FLORES-101 Benchmark]{\includegraphics[height=7cm]{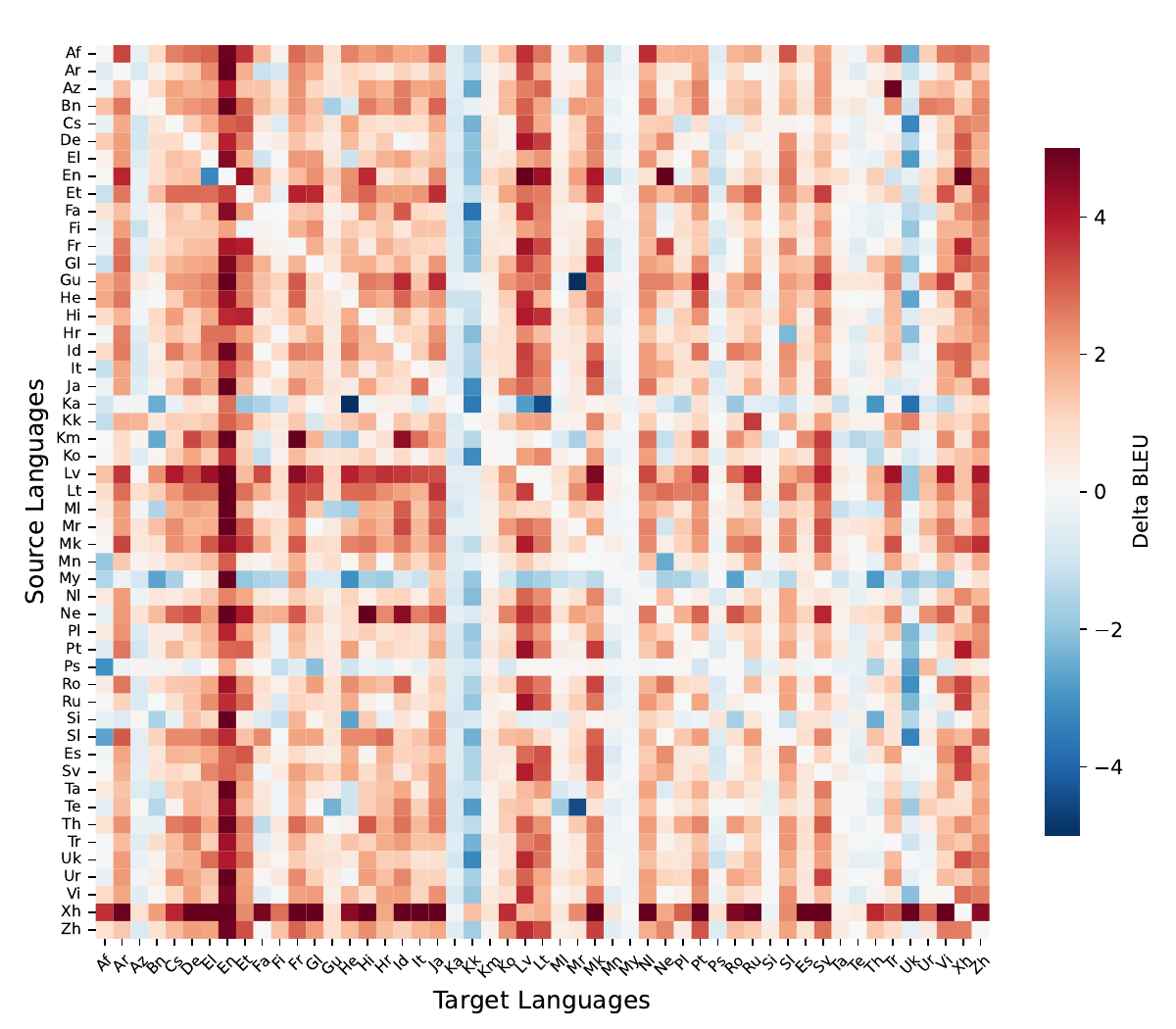}}
    \caption{Performance comparison of LLaMA3-8B-Instruct (blue) and AlignX (red) under the 51-language setup. (a): Accuracy metric results on multilingual general benchmarks. (b): BLEU differences on FLORES-101 benchmark, with red showing better AlignX performance and blue showing better LLaMA3-8B-Instruct performance.}
    \label{figures_main_results_51langs}
\end{figure*}

\begin{table*}[t]
\centering
\begingroup
\small
% \begin{tabular}{ccccc}
% \hline
% \multirow{2}{*}{\textbf{Methods}}  & \multirow{2}{*}{\textbf{X-En}} & \multirow{2}{*}{\textbf{En-X}} & \multicolumn{2}{c}{\textbf{Non-En}}  \\
%                                  &                                &                                & \textbf{BLEU}  & \textbf{OTR} \\ \hline
% \textbf{w/o MRA}                 & 36.02                          & 22.02                          & 16.34          & 8.67                \\
% \textbf{w/o $\mathcal{L}_{LAM}$} & \textbf{36.81}                 & 29.39                          & 18.59          & 12.59               \\
% \textbf{AlignX}                  & 36.25                          & \textbf{29.88}                 & \textbf{21.21} & \textbf{4.16}       \\ \hline
% \end{tabular}
\begin{tabular}{cccccccc}
\hline
\multirow{2}{*}{\textbf{Methods}} & \multirow{2}{*}{\textbf{TruthfulQA}} & \multirow{2}{*}{\textbf{HellaSwag}} & \multirow{2}{*}{\textbf{XNLI}} & \multicolumn{4}{c}{\textbf{IWSLT2017}}                            \\
                                  &                                      &                                     &                                & \textbf{X-En}  & \textbf{En-X}  & \textbf{Non-En} & \textbf{OTR}  \\ \hline
\textbf{LLaMA-7B}                 & 24.77                                & 41.79                               & 35.70                          & 31.20          & 16.08          & 10.90           & 30.92         \\
\textbf{w/o $\mathcal{L}_{CTR}$}            & 31.34                                & 44.03                               & 36.07                          & 36.13          & 28.68          & 20.01           & 4.89          \\
\textbf{w/o $\mathcal{L}_{LAM}$}            & \textbf{32.22}                       & 43.99                               & \textbf{36.27}                 & \textbf{36.81} & 29.39          & 18.59           & 12.59         \\
\textbf{AlignX}                   & 32.12                                & \textbf{44.11}                      & 36.00                          & 36.25          & \textbf{29.88} & \textbf{21.21}  & \textbf{4.16} \\ \hline
\end{tabular}
\endgroup
\caption{Averaged results of variant models on LLaMA for multilingual general and IWSLT2017 benchmarks. "w/o $\mathcal{L}_{CTR}$" and "w/o $\mathcal{L}_{LAM}$" remove the instruction contrastive learning and language matching in the first stage, respectively. "Non-En" indicates translation between four non-English languages. "OTR" (off-target ratio) represents the proportion of outputs in incorrect target languages, thus the lower the better. We bold the best results.}
\label{table_ablation_5langs}
\end{table*}

\begin{figure*}[t]
    \centering
    \subfigure[Base Model]{\includegraphics[width=1.5in]{images/vis_flores_10langs_layer16.pdf}\label{d}} 
    \subfigure[xSFT]{\includegraphics[width=1.5in]{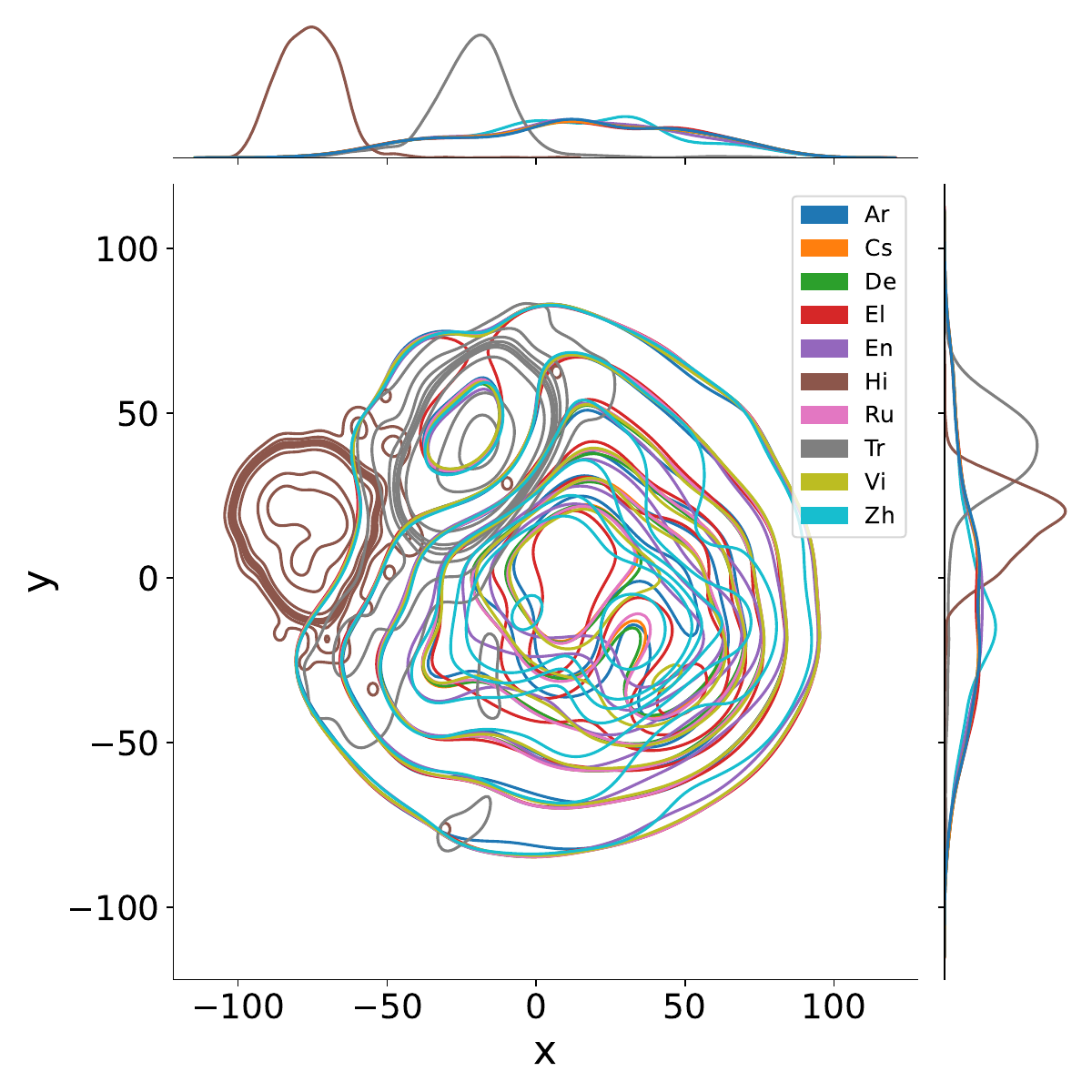}\label{e}}
    \subfigure[CPT]{\includegraphics[width=1.5in]{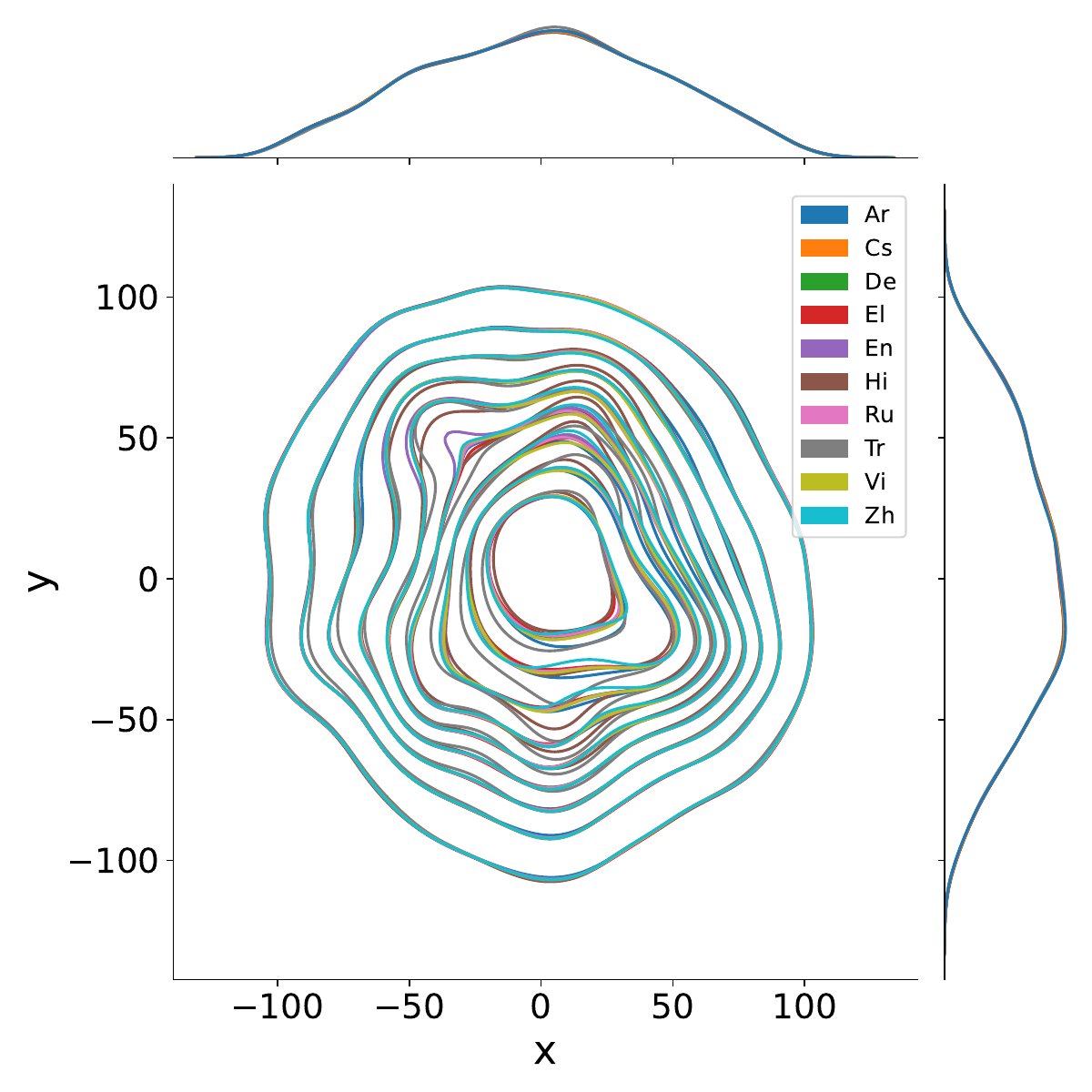}\label{f}}
    \subfigure[AlignX]{\includegraphics[width=1.5in]{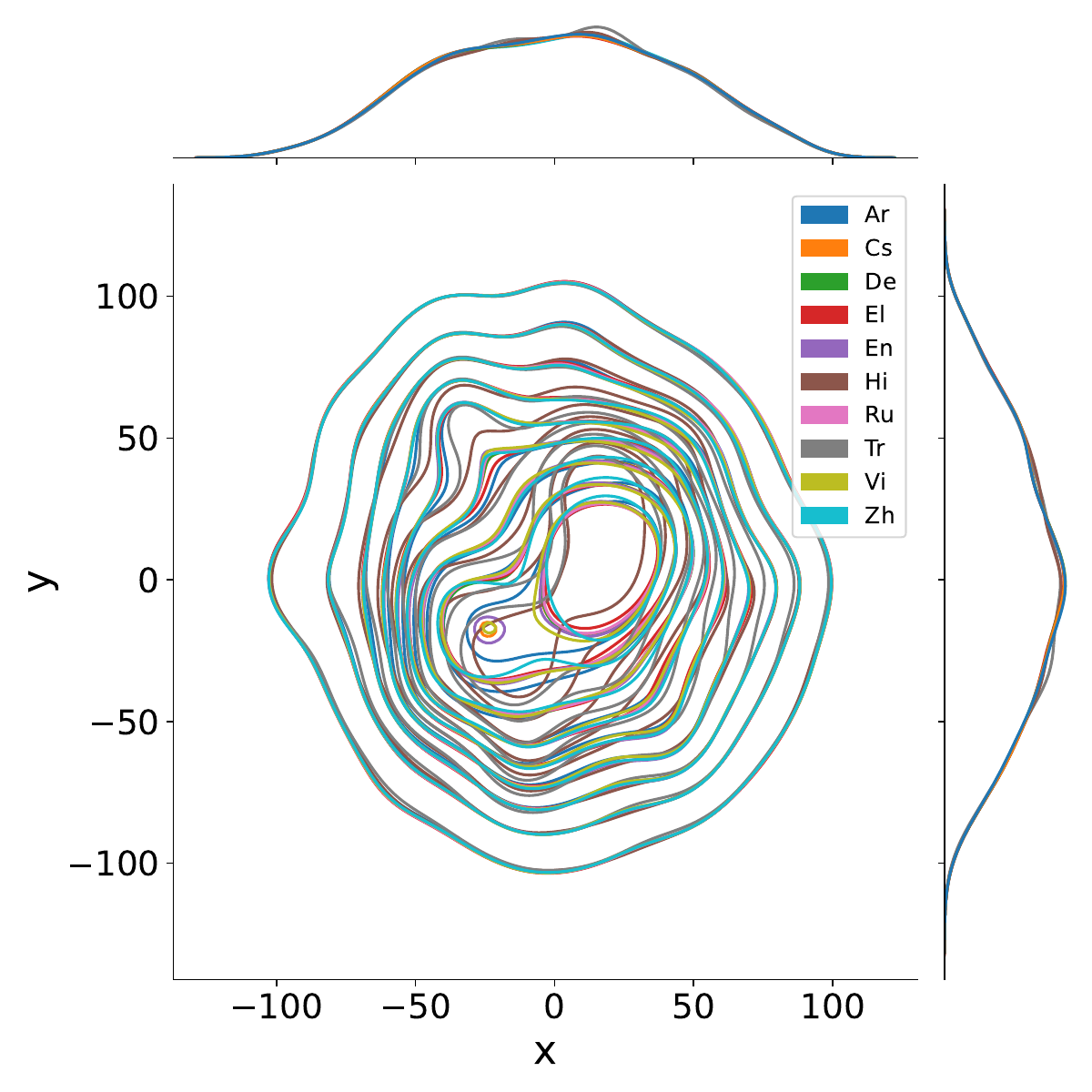}\label{g}}
    \caption{Visualization of multilingual representations from the base model, xSFT model, CPT model, and AlignX model after dimension reduction. The CPT and xSFT models train only the first and second stages of AlignX, respectively. The visualization follows the same steps as in Figure \ref{visualization_selected_layer}.}
    \label{visualization}
\end{figure*}

% \subsection{Scaling Up to 51 Languages}
\paragraph{Scaling AlignX to 51 languages further improves multilingual general and cross-lingual generation capabilities.}
To further validate the effectiveness of AlignX across a broader range of languages, we scale AlignX to 51 languages, listed in Appendix \ref{appendix_51_languages}. Figure \ref{figures_main_results_51langs} presents the results on multilingual general and translation benchmarks, with detailed results in Appendix \ref{appendix_detailed_results}. These results indicate that AlignX performs effectively under the 51-language setup, enhancing both multilingual general and cross-lingual generation capabilities.
To illustrate the impact of language scaling, we compare AlignX (51-langs) and AlignX (10-langs) under the 10-language setup, as shown in Figure \ref{figure_main_general} and Table \ref{table_main_translation}. Surprisingly, increasing the number of languages from 10 to 51 leads to further gains in multilingual general and generation tasks, without exhibiting the "curse of multilinguality"~\cite{zhu2024multilingual}. This highlights the efficacy of multilingual representation alignment across a large number of languages.

\section{Analysis}

% In this section, we present further analysis of AlignX. We first conduct an ablation study to evaluate the contribution of each module. Then, we visualize the representation alignment, revealing the implicit alignment of the data-level approach and the explicit alignment of AlignX. Finally, we evaluate AlignX with the latest cross-lingual alignment evaluation framework CLiKA~\cite{gao-etal-2024-multilingual}, which shows that AlignX achieves more balanced multilingual capabilities and improves the cross-lingual alignment of LLMs.
% In this section, we provide an in-depth analysis to understand the impact of AlignX on LLMs. We leverage additional benchmarks, detailed in Appendix \ref{appendix_evaluation_benchmarks}. 

In this section, we provide an in-depth analysis of AlignX, examining the following aspects: the role of two auxiliary objectives, improvements in multilingual representation alignment, enhancements in cross-lingual generation and knowledge transfer, the impact of corpus size, and gains in cross-lingual alignment. Appendix~\ref{additional_analysis} provides detailed comparisons with representation-level AFP, along with efficiency analysis. Additional evaluation benchmarks are reported in Appendix~\ref{appendix_evaluation_benchmarks}.

% \subsection{Ablation Study}
\paragraph{Instruction contrastive learning facilitates knowledge sharing, and language matching promotes more accurate cross-lingual generation.}

We conduct experiments on several variant models to investigate the effectiveness of training objectives. We experiment with German, English, Italian, Dutch, and Romanian. These five languages belong to the Indo-European language family, which are relatively similar. We extract the training and test sets from the IWSLT2017 dataset, follow the same data construction process as the main experiment, and experiment on LLaMA-7B.
% We evaluate multilingual translation benchmarks in a zero-shot setup.
% Table \ref{table_ablation_5langs} indicates the results, showing the effectiveness of multilingual representation alignment, $\mathcal{L}_{CTR}$ and $\mathcal{L}_{LAM}$. Specifically, the instruction contrastive learning task significantly enhances the En-X direction performance, suggesting that the alignment semantic space mainly benefits relatively low-resource languages. However, aligning the semantic space alone can hinder the model's ability to distinguish between output languages, leading to an increase in incorrect language outputs in cross-lingual tasks like multilingual translation. To mitigate this off-target issue, the language matching task serves as a regularizer for output languages, improving the model's ability to distinguish output languages. The language matching task further enhances the model's ability to distinguish output languages. Overall, the multilingual representation alignment effectively enhances the model's translation performance in non-English generation.
Table \ref{table_ablation_5langs} presents the results, highlighting the effectiveness of multilingual representation alignment. Specifically, instruction contrastive learning facilitates knowledge sharing, primarily benefiting relatively low-resource languages, but increases incorrect linguistic output in cross-lingual generation. To mitigate this off-target issue, language matching serves as a regularizer for output languages, enhancing LLMs' ability to distinguish output languages.

\begin{table}[t]
\centering
\small
\begin{tabular}{ccc}
\hline
\textbf{W/T/L} & \textbf{Grammar}  & \textbf{Language Fit} \\ \hline
\textbf{X-El}             & 173 / 174 / 103   & 192 / 138 / 120                   \\
\textbf{All}              & 1098 / 2901 / 501 & 1223 / 2611 / 666                 \\ \hline
\end{tabular}
\caption{Automatic evaluation results using GPT-4o on cross-lingual generation. We compare paired outputs of AlignX and LLaMA3-8B-Instruct, where “W/T/L” indicates that AlignX produces better, comparable, or worse translations, respectively. "X" denotes all other nine languages except the target language.}
\label{gpt4o_evaluation}
\end{table}

% \subsection{Can AlignX Bring the Multilingual Representations Closer?}
\paragraph{AlignX effectively brings the multilingual representations closer.}
% 具体句子数量、怎么取的表示
% 为进一步理解不同方法对表示空间的作用，我们可视化模型内部的隐状态表示，包括data-level的xSFT和representation-level的AlignX。我们使用FLORES-101 dev set，并在LLaMA3-8B上做实验。我们对中间层隐状态做mean-pool操作，使用t-SNE将表示降维到2维。
To intuitively understand the effectiveness of multilingual representation alignment, we visualize multilingual representations of LLaMA3, data-level xSFT, and representation-level CPT, AlignX.
% 图3展示了可视化结果，证明了xSFT隐式对齐表示和AlignX显示对齐表示。图a和b的对比表明，xSFT在没有显式辅助任务的情况下初步对齐了多语言表示，尽管仍有一些距离较远的语言。我们认为这种对齐能力来自翻译对，这天然地蕴含对齐信息。图b和c的对比表明，第一阶段多语言表示对齐在拉近多语言表示上发挥重要作用。
Figure \ref{visualization} presents visualization results, demonstrating the implicit multilingual alignment of xSFT and explicit multilingual alignment of AlignX. Comparing Figures \ref{d} and \ref{e}, xSFT preliminarily aligns multilingual representations without explicit auxiliary tasks, although some distant languages still exist. This alignment capability comes from translation pairs, which naturally contain alignment information. The comparison in Figures \ref{e} and \ref{g} shows that multilingual representation alignment in the first stage is important in bringing multilingual representations closer. Appendix \ref{appendix_detailed_visualization} presents detailed results for AlignX.

\paragraph{AlignX achieves better grammatical correctness and more accurate target language according to GPT-4o evaluation.} 
To further examine how AlignX enhances cross-lingual generation, particularly in low-resource scenarios, we employ GPT-4o as an automatic evaluator focusing on the \textit{grammar} and \textit{language fit} dimensions. Evaluations are conducted under the 10-language setup. For each translation direction, we randomly sample 50 instances and compare the outputs from LLaMA3-8B-Instruct and AlignX. 
As shown in Table~\ref{gpt4o_evaluation}, detailed in Appendix~\ref{details_gpt4o_evaluation}, AlignX demonstrates stronger grammatical correctness and more appropriate target language across languages, especially low-resource languages (e.g., Greek).

\begin{figure}[t]
  \centering
  \includegraphics[width=1.0\linewidth]{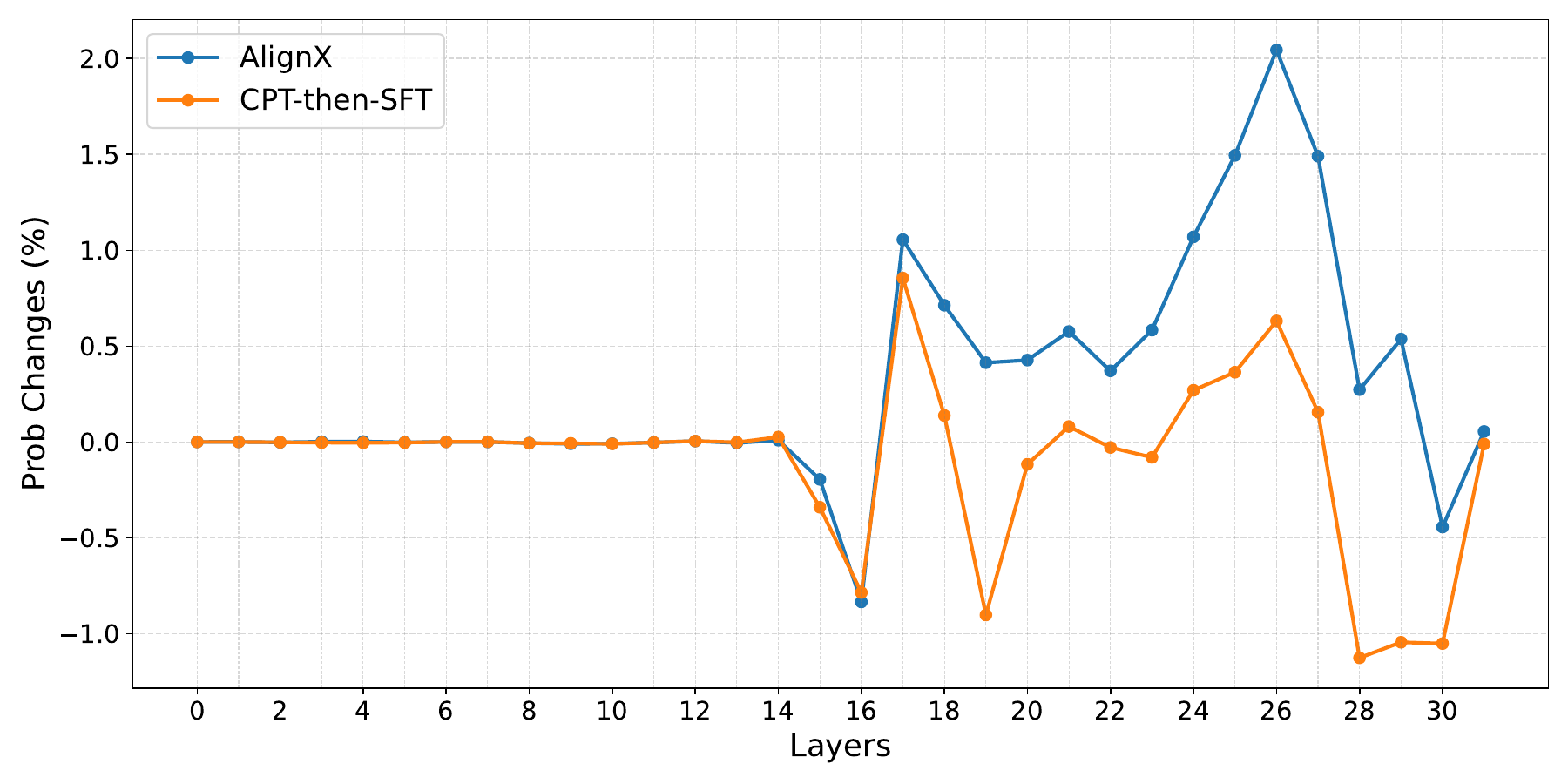}
  \caption{Layer-wise probability changes of unembedding the correct English answer token at non-final layers during non-English translation among German, Russian, and Chinese (e.g., De-Ru). Probabilities are averaged across samples.}
  \label{fig_delta_probs_cmp}
\end{figure}

\paragraph{AlignX facilitates knowledge transfer via cross-lingual concept alignment in intermediate layers.}
Following~\citet{wendler-etal-2024-llamas}, we analyze how AlignX enhances knowledge transfer and influences cross-lingual representations. Specifically, we measure the likelihood that non-English inputs activate English-centric representations in intermediate layers through German, Russian, and Chinese translation tasks (e.g., German–Chinese). For each direction, we compute the probability of unembedding the correct English token at non-final layers, averaged across samples, and report the change relative to LLaMA2 (e.g., AlignX – LLaMA2).
Figure \ref{fig_delta_probs_cmp} reveals that in the concept space~\citep{wendler-etal-2024-llamas}, approximately layers 16 to 28, associated with language-agnostic processing, AlignX exhibits higher probabilities of retrieving the English token, whereas CPT-then-SFT fluctuates around zero. This indicates that AlignX better aligns cross-lingual representations with the English-centric concept space, leveraging English as an interlingua to improve cross-lingual transfer.

\begin{figure*}[t]
    \centering
    \includegraphics[width=1.0\linewidth]{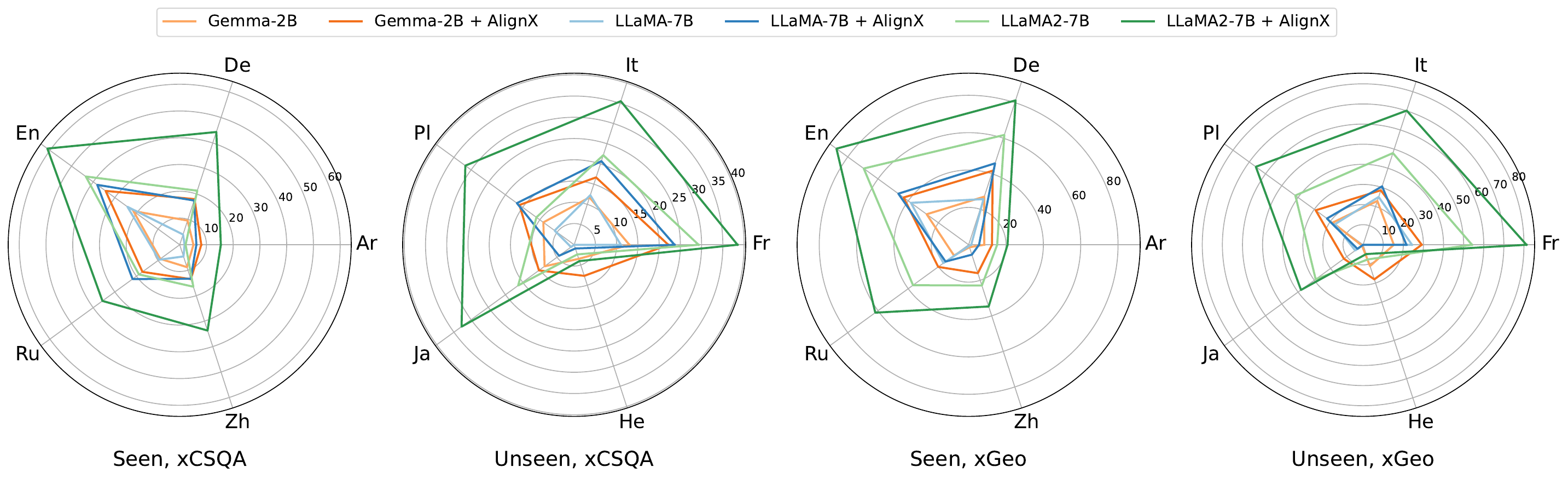}
    \caption{Results of the general cross-lingual knowledge alignment evaluation with CLiKA~\cite{gao-etal-2024-multilingual}. We present the re-scaled accuracy (RA) scores, where higher values (greater than 0) indicate better performance. "Seen" and "Unseen" denote whether these languages are included during AlignX training. xCSQA and xGeo evaluate \textit{Basic} knowledge and \textit{Factual} knowledge, respectively.}
    \label{clika}
\end{figure*}

% \usepackage{multirow}
\begin{table}[t]
\centering
\begingroup
\small
\begin{tabular}{ccccc} \hline
\multirow{2}{*}{\textbf{Methods}} & \multicolumn{2}{c}{\textbf{WMT14}} & \multicolumn{2}{c}{\textbf{WMT17}} \\
                                & \textbf{De-En}   & \textbf{En-De}  & \textbf{Zh-En}   & \textbf{En-Zh}  \\ \hline
\multicolumn{5}{c}{\textbf{LLaMA-7B Based}}                                                               \\ \hline
\textbf{LLaMA-7B}               & 28.95 & 18.30 & 15.75 & 10.58               \\
\textbf{xSFT}                    & 28.83            & 19.15           & 16.16            & 18.69           \\
\textbf{AlignX}                 & 31.45            & 22.25           & 18.37            & 24.86           \\
\textbf{AlignX$^\dag$}                 & \textbf{32.16}   & \textbf{24.79}  & \textbf{19.43}   & \textbf{29.15}  \\ \hline
\multicolumn{5}{c}{\textbf{LLaMA2-7B Based}}                                                              \\ \hline
\textbf{LLaMA2-7B}              & 31.25 & 20.32 & 21.41 & 20.97               \\
\textbf{xSFT}                    & 30.81            & 20.42           & 18.88            & 24.51           \\
\textbf{AlignX}                 & 32.04            & 23.59           & 19.76            & 29.13           \\
\textbf{AlignX$^\dag$}                 & \textbf{33.94}   & \textbf{26.70}  & \textbf{22.00}   & \textbf{32.18}  \\ \hline
\multicolumn{5}{c}{\textbf{LLaMA3-8B Based}}                                                              \\ \hline
\textbf{LLaMA3-8B}              & 33.66 & 23.70 & 25.03 & 31.60                \\
\textbf{xSFT}                    & 34.77            & 26.45           & \textbf{25.06}            & 35.77           \\
\textbf{AlignX}                 & \textbf{36.09}                & 27.23               & 24.84                & 36.26               \\
\textbf{AlignX$^\dag$}                 & 35.70   & \textbf{28.33}  & 24.32            & \textbf{36.46} \\  \hline
\end{tabular}
\endgroup
\caption{The results for the WMT14EnDe and WMT17EnZh benchmarks. The xSFT model only trains the second stage of multilingual instruction fine-tuning. AlignX$^\dag$ is a variant that increases the training data in the first stage from 50K to 250K examples per language direction. We bold the highest scores.}
\label{table_ablation_3langs}
\end{table}

% 第二组的3种语言距离相对较远，并且包含LLaMA未支持的语言中文。表中展示了结果，表明AlignX主要增强非英语语向性能，这与表2的结果一致。此外，扩展一阶段数据可以持续增强性能，尤其是非英语语向。
% \paragraph{Ablation on Corpora Size}
\paragraph{Scaling up the dataset of multilingual representation alignment consistently enhances non-English language generation.}
We investigate the impact of scaling the first-stage training dataset for multilingual representation alignment on language generation, using German (De), English (En), and Chinese (Zh) with LLaMA-7B. Training and test sets are drawn from WMT14EnDe and WMT17EnZh. In Stage 1, we vary the corpus size for each language pair from 50K to 250K, while Stage 2 training follows the same settings as in the main experiment.
As presented in Table \ref{table_ablation_3langs}, AlignX mainly enhances the performance in the non-English translation directions, consistent with Table \ref{table_ablation_5langs}. Moreover, extending corpora size in the first stage further enhances performance, especially in non-English generation.

% \subsection{Cross-lingual Alignment Evaluation}
\paragraph{AlignX achieves better cross-lingual alignment.}
To validate AlignX's impact on cross-lingual alignment, we use the CLiKA framework~\cite{gao-etal-2024-multilingual}, evaluating \textit{Basic} and \textit{Factual} knowledge with the xCSQA and xGeo benchmarks. We compute re-scaled accuracy scores, which exclude the interference of random baseline and question difficulty for better cross-lingual comparison.
Results in Figure \ref{clika} show that AlignX improves cross-lingual alignment across multiple languages, reducing multilingual performance gaps. Remarkably, AlignX also generalizes well to unseen languages, with cross-lingual alignment correlating strongly with similar languages from training, suggesting generalization is influenced by language family similarities.

\section{Conclusion}

% 在本文中，我们提出了一个高效的两阶段多语言表示对齐框架，AlignX。多个基座LLM和多个benchmark上的结果表明，AlignX有效增强了多语言通用能力和跨语言生成能力。分析实验进一步证明了AlignX在拉近多语言表示、提升跨语言对齐、增强多语言能力的作用。
In this paper, we propose AlignX, an efficient two-stage representation-level framework for enhancing multilingual performance of multilingual LLMs. Results on several pre-trained LLMs and multiple widely used benchmarks show that AlignX effectively enhances multilingual general capabilities and cross-lingual generation capabilities. The analysis further demonstrates the impact of AlignX on LLMs, including bringing multilingual representations closer and improving the cross-lingual alignment of LLMs.
\section*{Limitations}

% AlignX有效缓解了多语言能力不均衡，但距离balanced多语言能力还有很长距离。
Given LLMs with uneven multilingual capabilities, AlignX enhances both the multilingual general capability and cross-lingual generation capability, and alleviates the imbalance of multilingual capabilities, which matches our motivation. However, the multilingual performance of the final model is the outcome of the combined influence of the base model and AlignX, meaning that the imbalance remains unavoidable. This suggests that we are still far from achieving fully balanced multilingual capabilities.
\section*{Acknowledgements}

We thank all the anonymous reviewers for their insightful and valuable comments on this paper. This work was supported by the grant from the National Natural Science Foundation of China (No. 62376260).

\bibliography{custom}

\newpage
% \onecolumn
\appendix

% \section{Example Appendix}
% \label{sec:appendix}

% This is an appendix.

\section{Detailed Visualization across Layers on LLaMA3}
\label{appendix_detailed_visualization}

\begin{figure*}[htbp]
    \centering
    % \subfigure[Embedding]{\includegraphics[width=1.5in]{images/vis_flores_10langs_layer0.pdf}} 
    % \subfigure[Layer 1]{\includegraphics[width=1.5in]{images/vis_flores_10langs_layer1.pdf}} 
    \subfigure[Layer 2]{\includegraphics[width=1.5in]{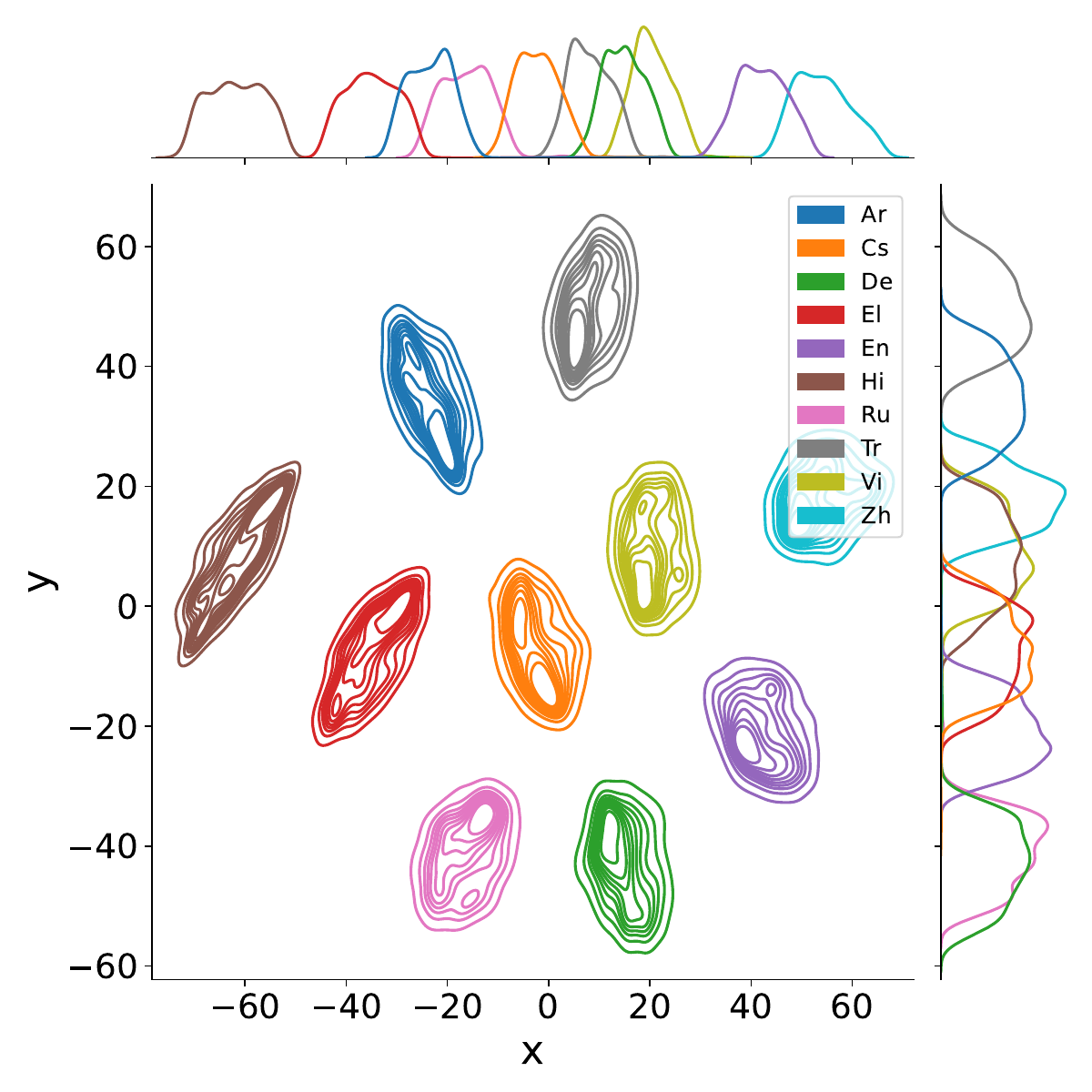}}
    % \subfigure[Layer 3]{\includegraphics[width=1.5in]{images/vis_flores_10langs_layer3.pdf}}
    \subfigure[Layer 4]{\includegraphics[width=1.5in]{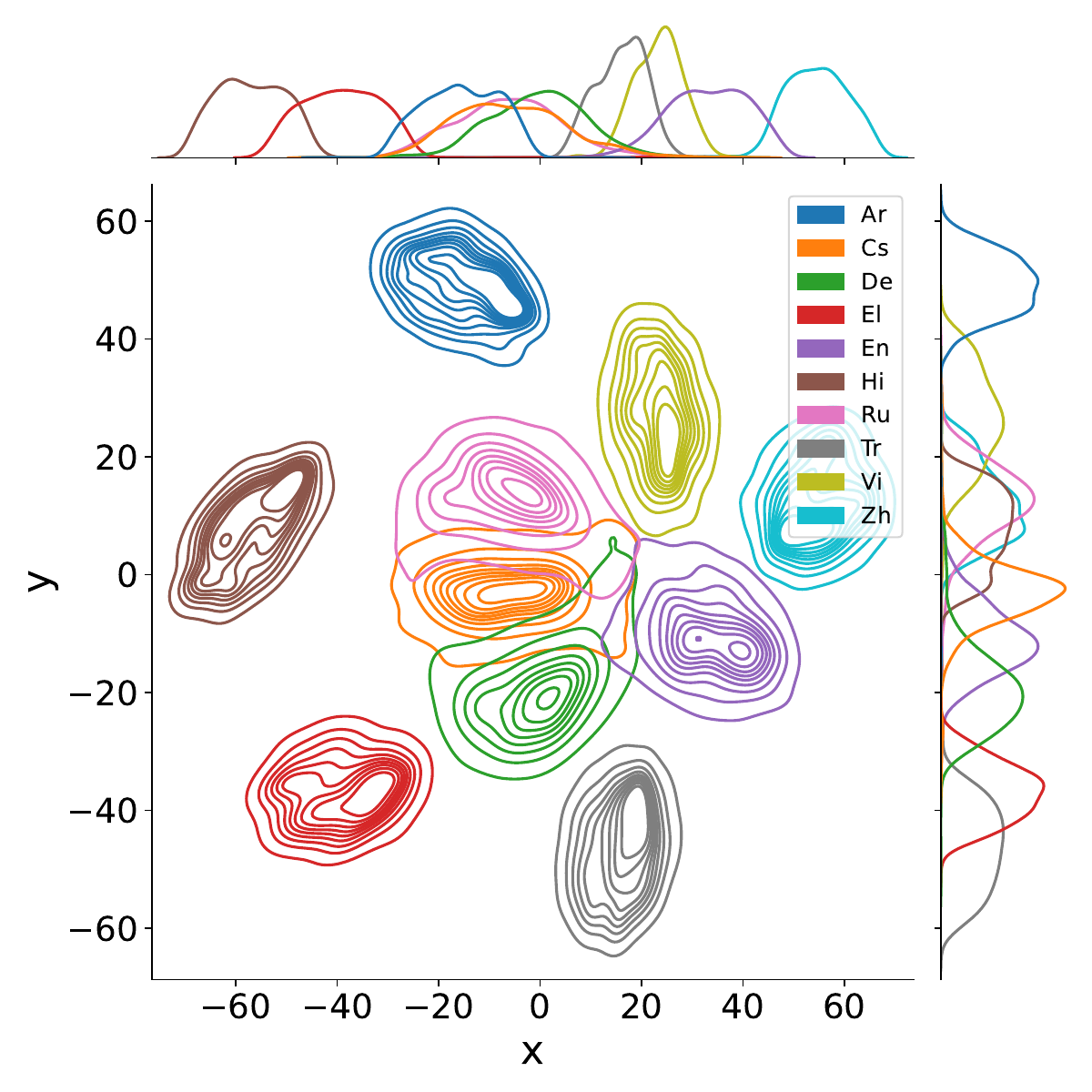}}
    % \subfigure[Layer 5]{\includegraphics[width=1.5in]{images/vis_flores_10langs_layer5.pdf}} 
    \subfigure[Layer 6]{\includegraphics[width=1.5in]{images/vis_flores_10langs_layer6.pdf}} 
    % \subfigure[Layer 7]{\includegraphics[width=1.5in]{images/vis_flores_10langs_layer7.pdf}}
    \subfigure[Layer 8]{\includegraphics[width=1.5in]{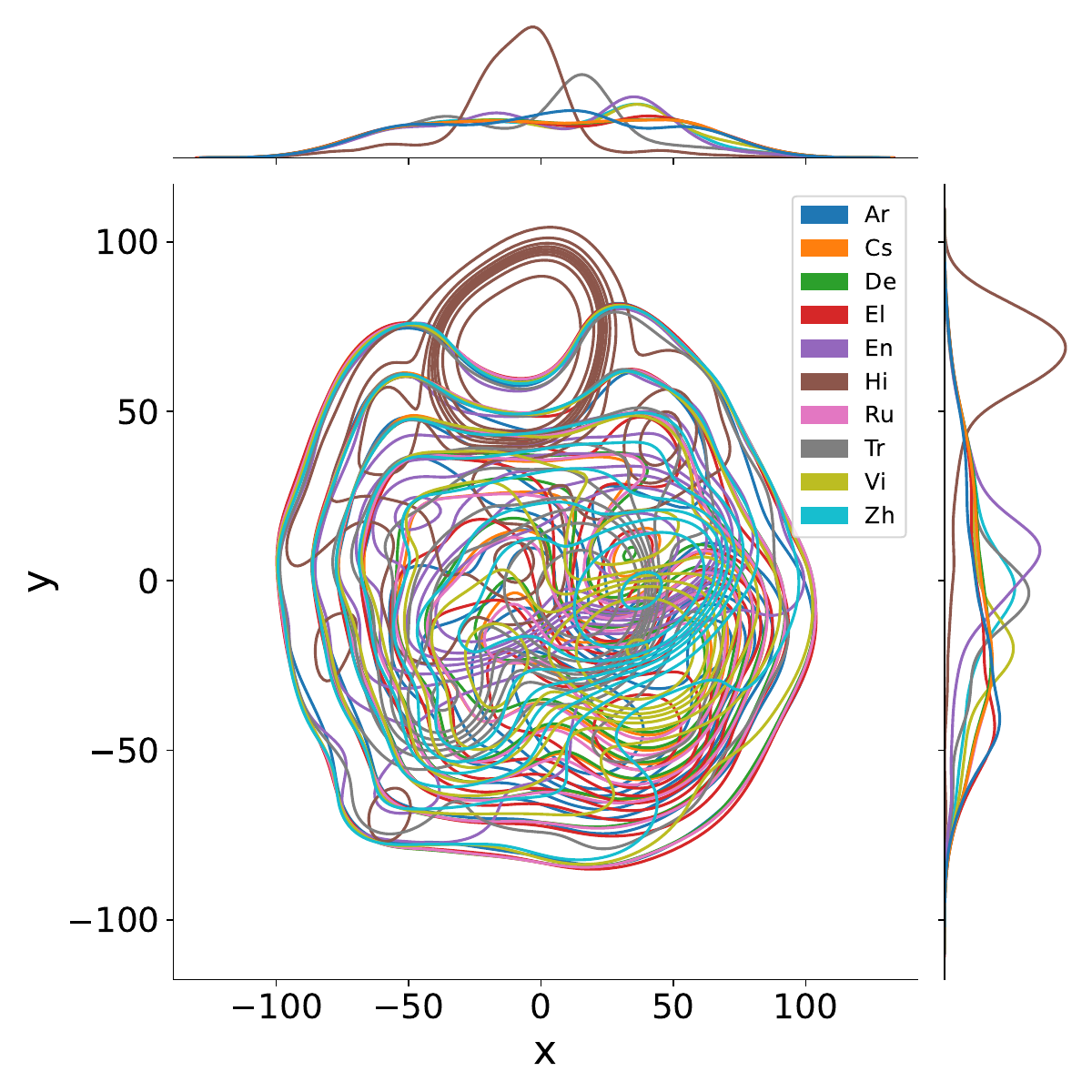}} \\
    % \subfigure[Layer 9]{\includegraphics[width=1.5in]{images/vis_flores_10langs_layer9.pdf}} \\
    \subfigure[Layer 10]{\includegraphics[width=1.5in]{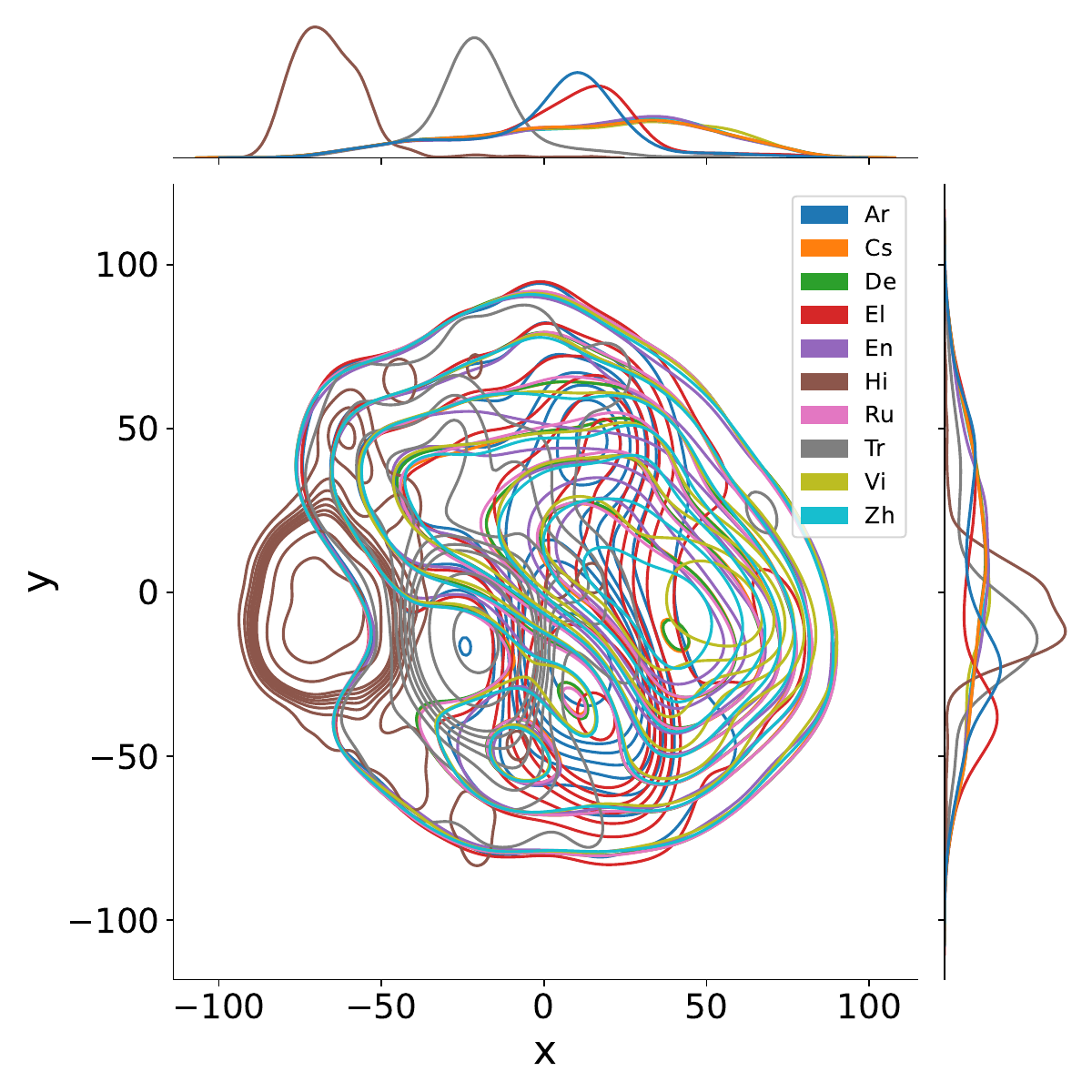}} 
    % \subfigure[Layer 11]{\includegraphics[width=1.5in]{images/vis_flores_10langs_layer11.pdf}} 
    \subfigure[Layer 12]{\includegraphics[width=1.5in]{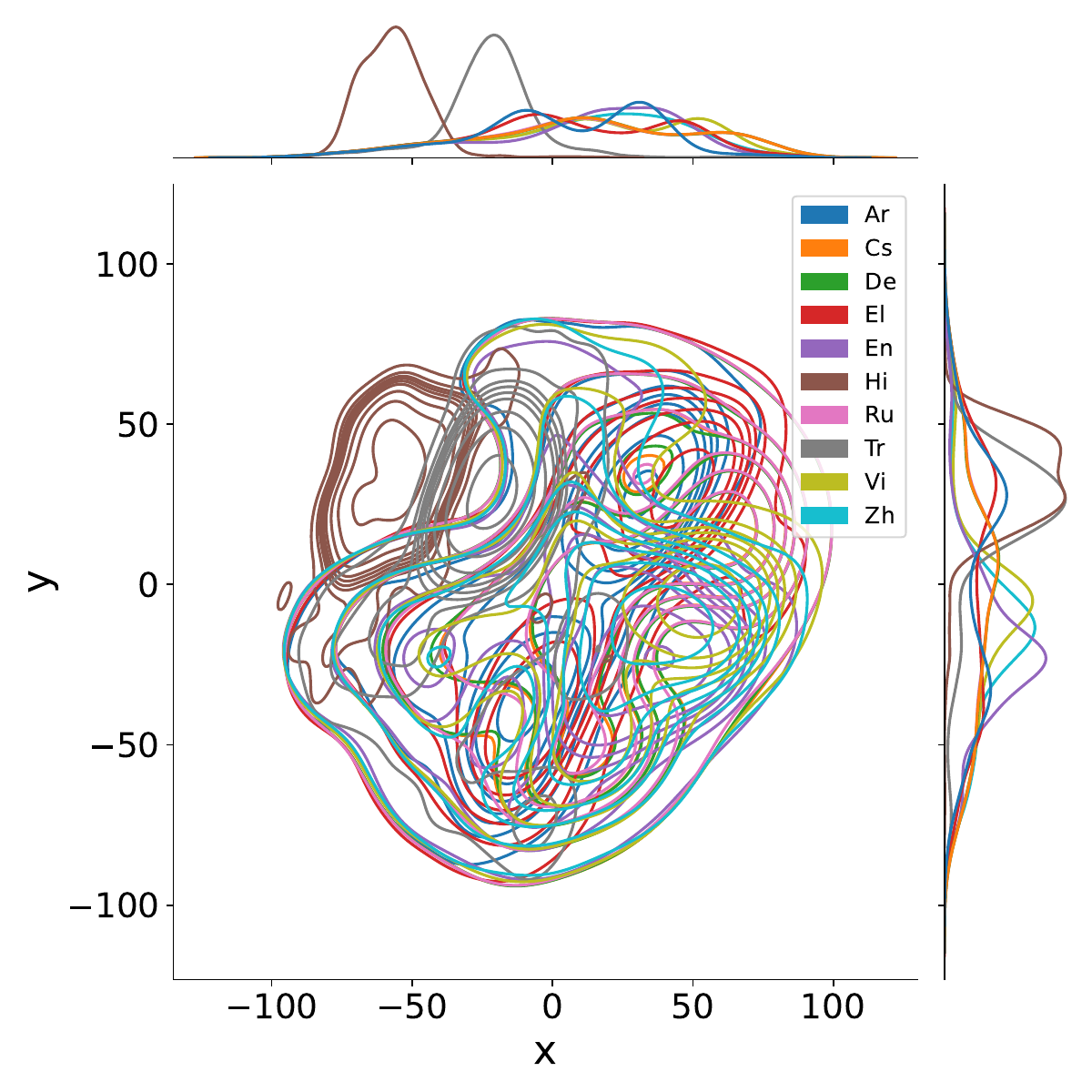}}
    % \subfigure[Layer 13]{\includegraphics[width=1.5in]{images/vis_flores_10langs_layer13.pdf}}
    \subfigure[Layer 14]{\includegraphics[width=1.5in]{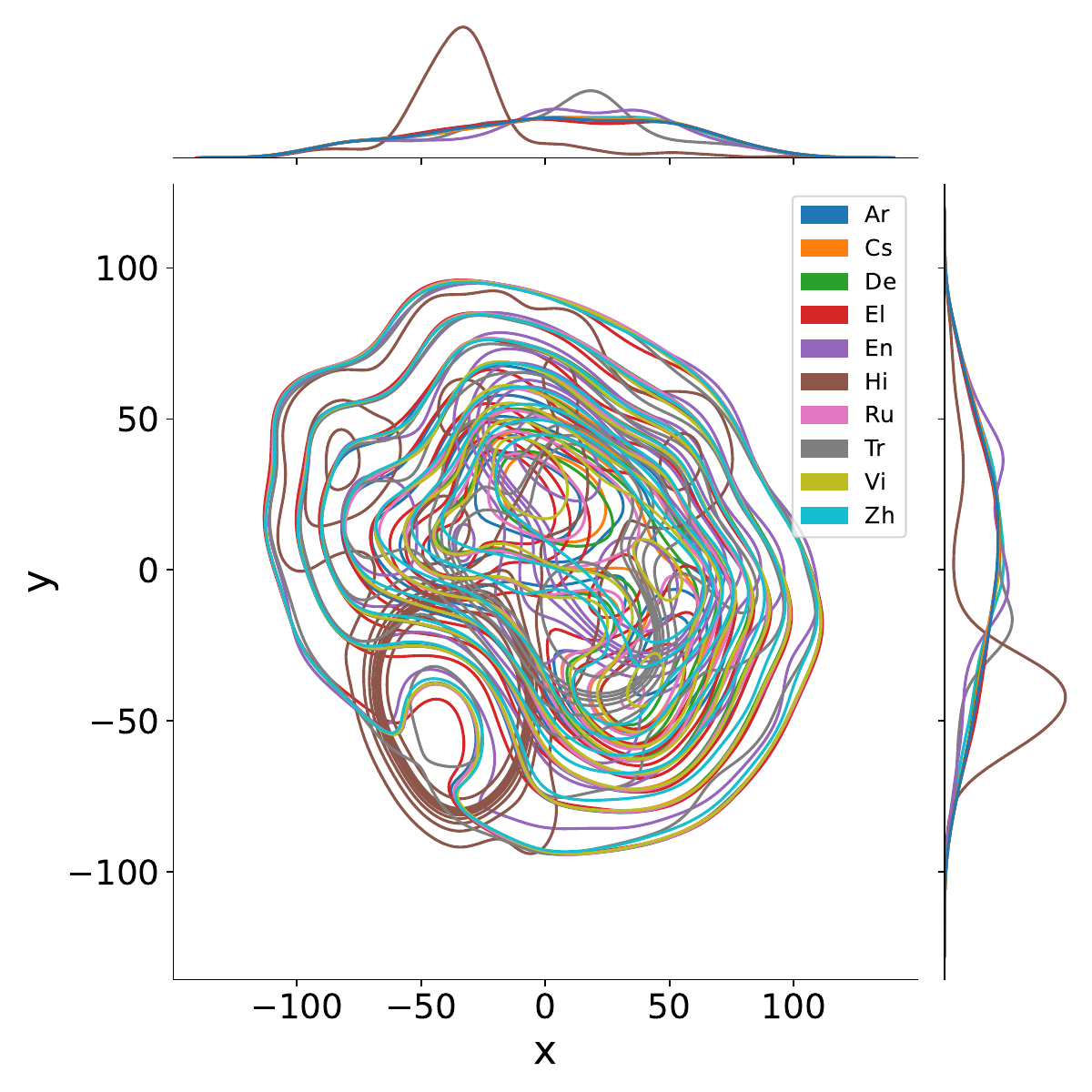}}
    % \subfigure[Layer 15]{\includegraphics[width=1.5in]{images/vis_flores_10langs_layer15.pdf}} 
    \subfigure[Layer 16]{\includegraphics[width=1.5in]{images/vis_flores_10langs_layer16.pdf}} \\
    % \subfigure[Layer 17]{\includegraphics[width=1.5in]{images/vis_flores_10langs_layer17.pdf}}
    \subfigure[Layer 18]{\includegraphics[width=1.5in]{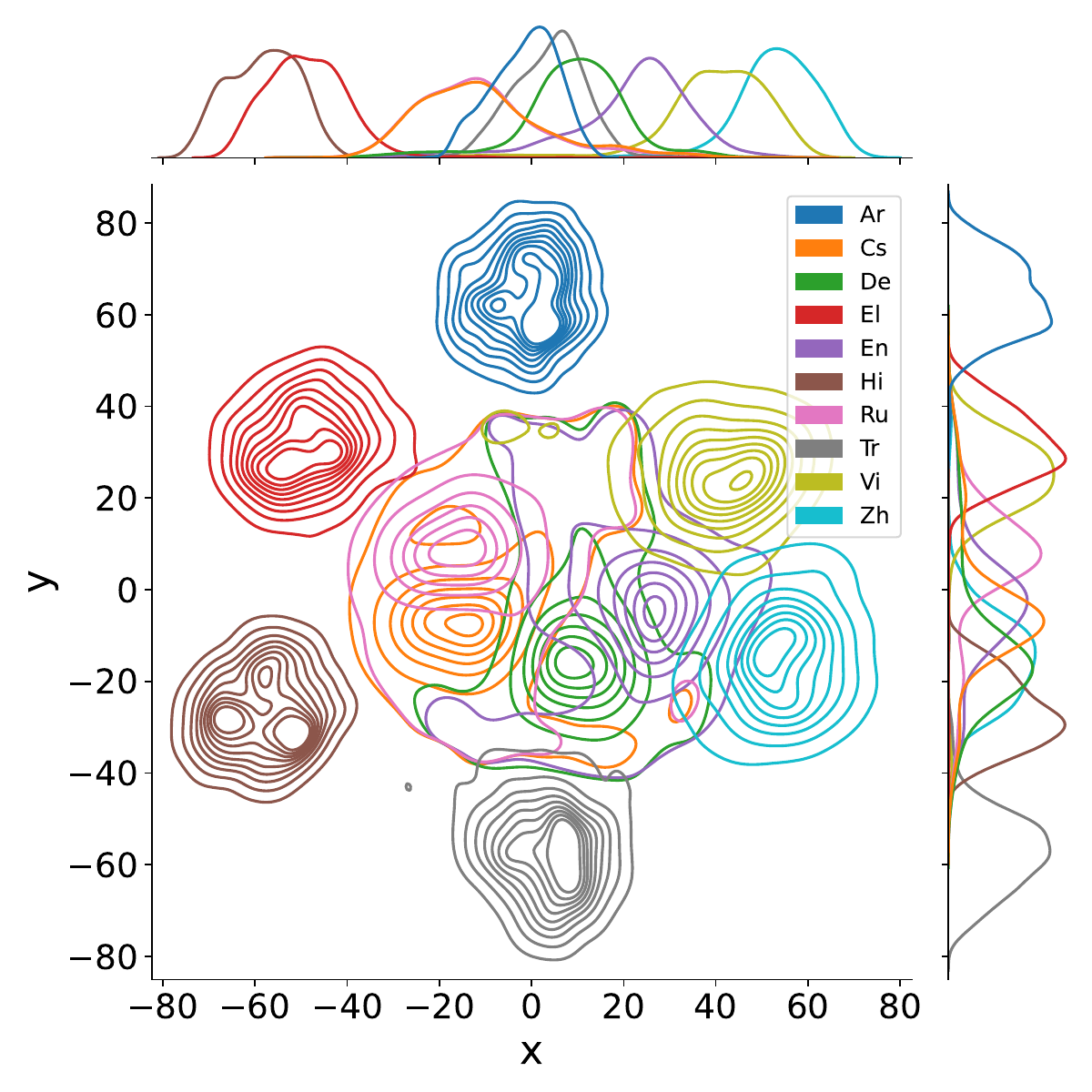}}
    % \subfigure[Layer 19]{\includegraphics[width=1.5in]{images/vis_flores_10langs_layer19.pdf}}
    \subfigure[Layer 20]{\includegraphics[width=1.5in]{images/vis_flores_10langs_layer20.pdf}} 
    % \subfigure[Layer 21]{\includegraphics[width=1.5in]{images/vis_flores_10langs_layer21.pdf}} 
    \subfigure[Layer 22]{\includegraphics[width=1.5in]{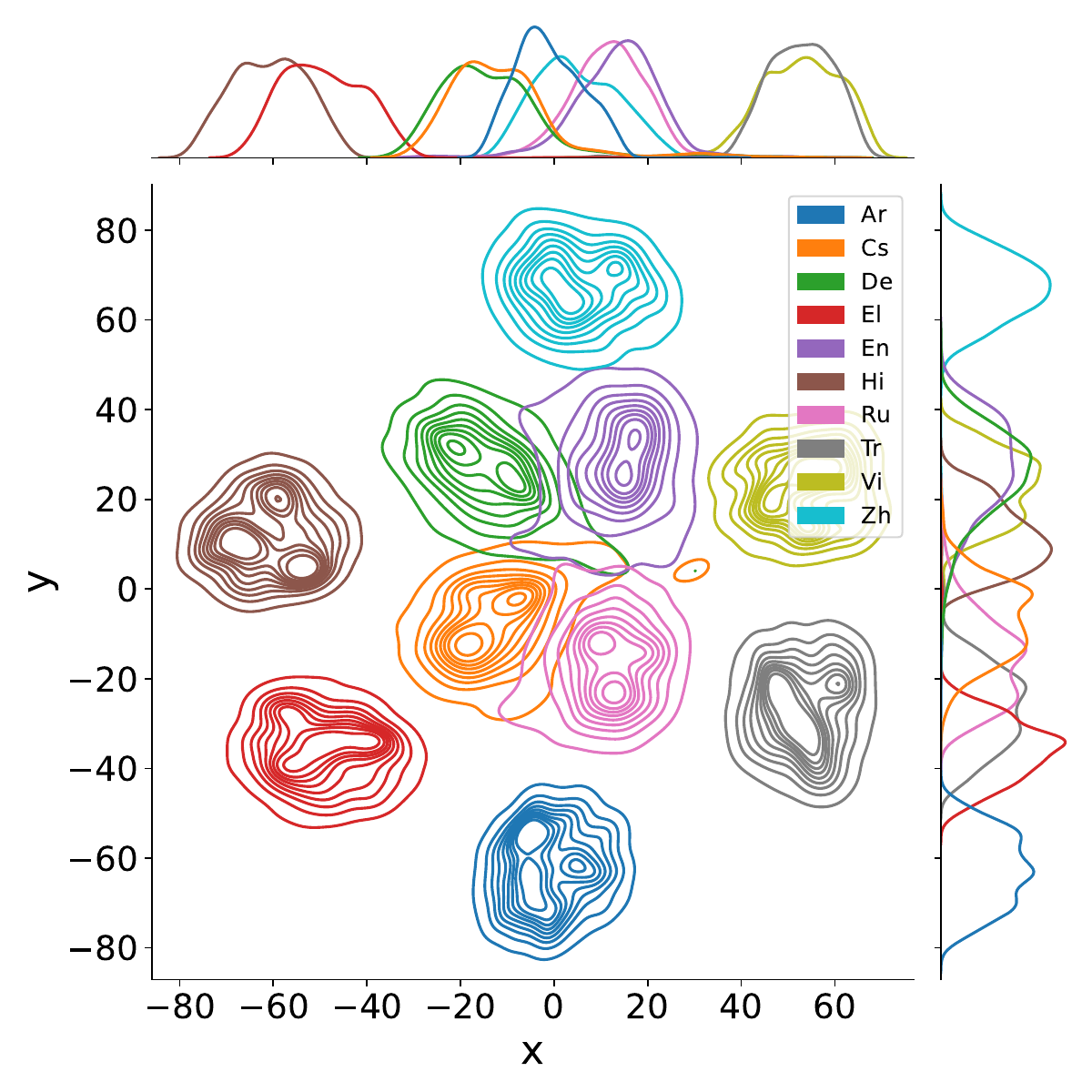}}
    % \subfigure[Layer 23]{\includegraphics[width=1.5in]{images/vis_flores_10langs_layer23.pdf}}
    \subfigure[Layer 24]{\includegraphics[width=1.5in]{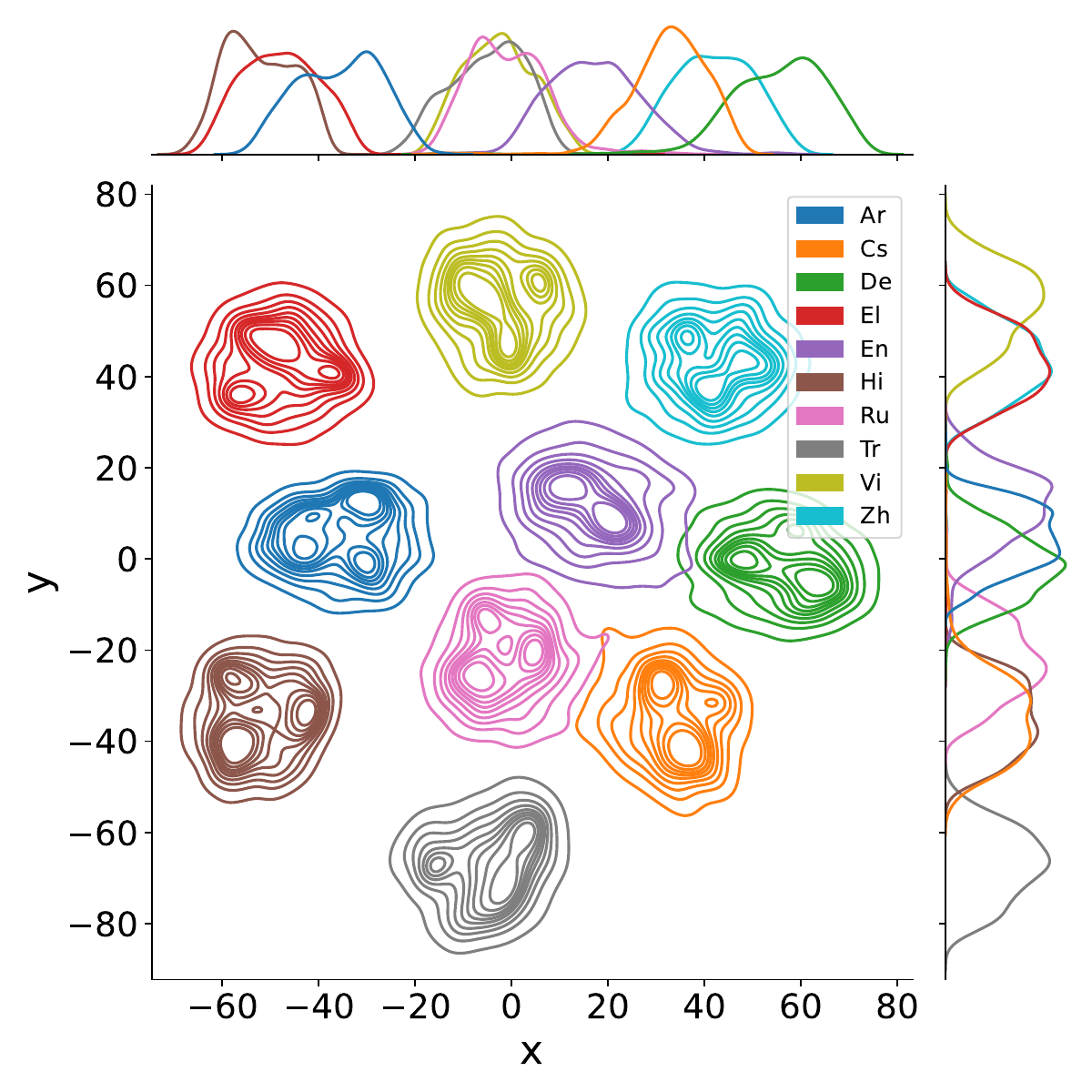}} \\
    % \subfigure[Layer 25]{\includegraphics[width=1.5in]{images/vis_flores_10langs_layer25.pdf}} 
    \subfigure[Layer 26]{\includegraphics[width=1.5in]{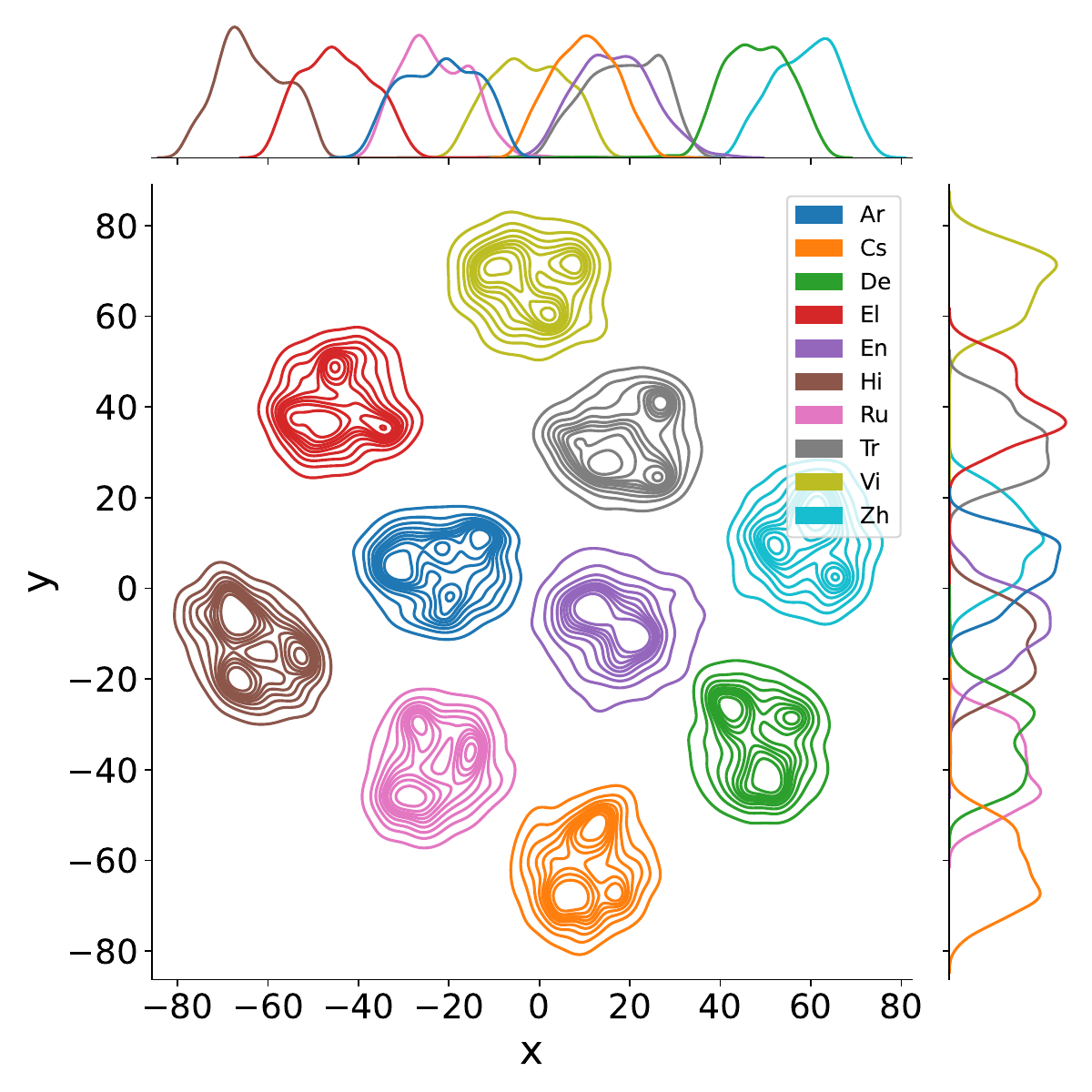}} 
    % \subfigure[Layer 27]{\includegraphics[width=1.5in]{images/vis_flores_10langs_layer27.pdf}}
    \subfigure[Layer 28]{\includegraphics[width=1.5in]{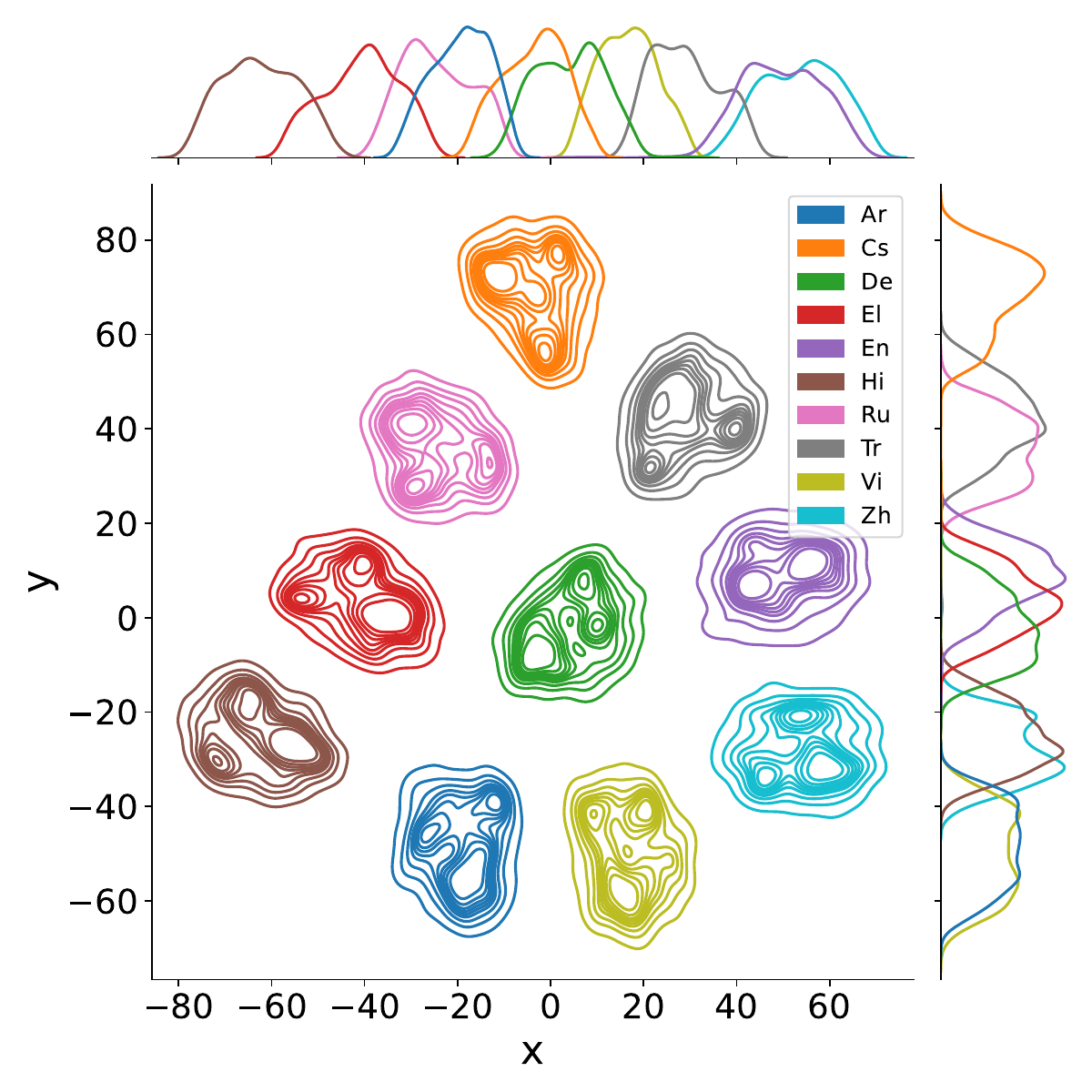}}
    % \subfigure[Layer 29]{\includegraphics[width=1.5in]{images/vis_flores_10langs_layer29.pdf}}
    \subfigure[Layer 30]{\includegraphics[width=1.5in]{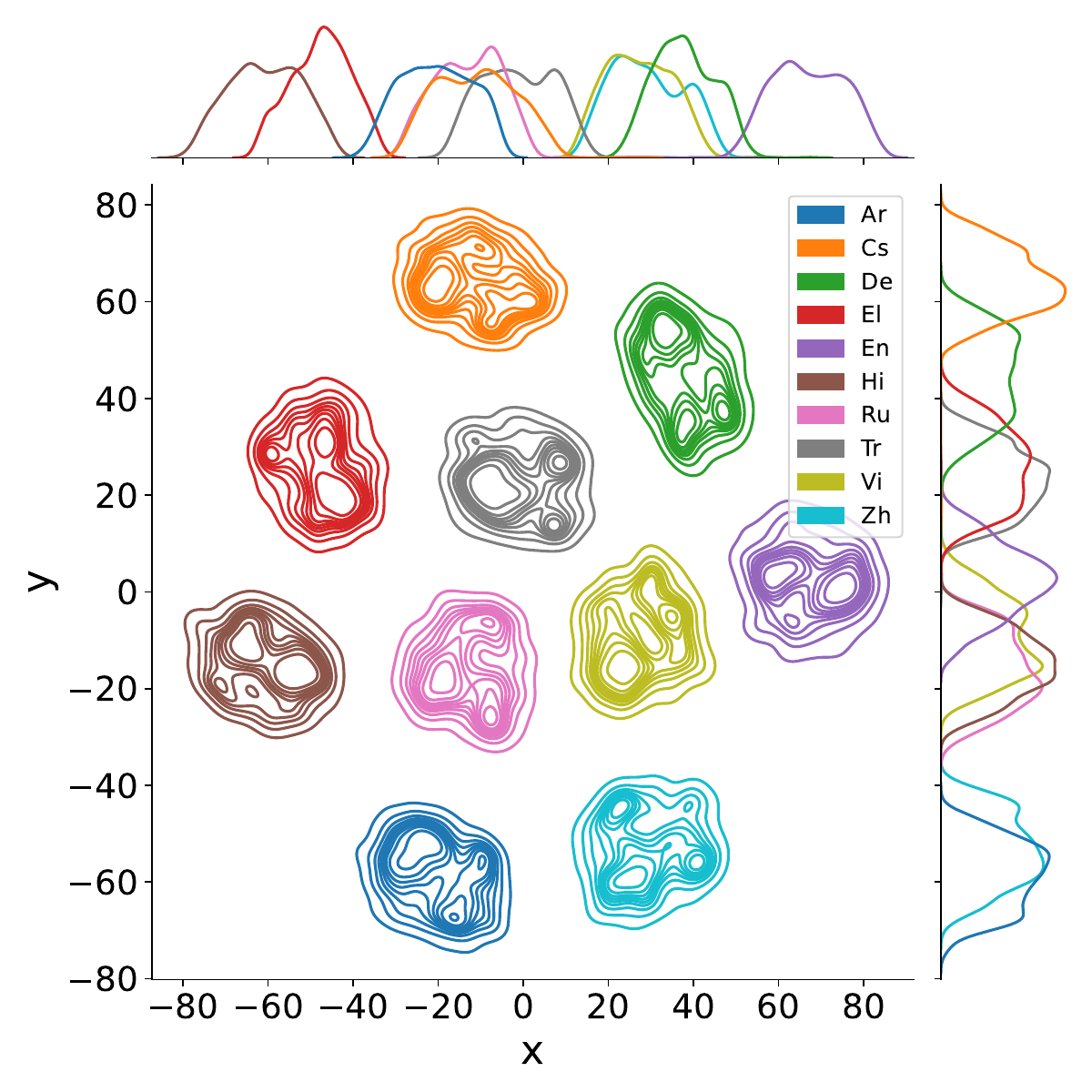}}
    % \subfigure[Layer 31]{\includegraphics[width=1.5in]{images/vis_flores_10langs_layer31.pdf}}
    \subfigure[Layer 32]{\includegraphics[width=1.5in]{images/vis_flores_10langs_layer32.pdf}}
    \caption{Preliminary visualization of the multilingual representations across layers in the LLaMA3 after dimension reduction. We leverage the FLORES-101 dev set, which is multi-way parallel.}
    \label{visualization_all_layer}
\end{figure*}

\begin{figure*}[htbp]
    \centering
    % \subfigure[Embedding]{\includegraphics[width=1.5in]{images/ours-vis_flores_10langs_layer0.pdf}} 
    % \subfigure[Layer 1]{\includegraphics[width=1.5in]{images/ours-vis_flores_10langs_layer1.pdf}} 
    \subfigure[Layer 2]{\includegraphics[width=1.5in]{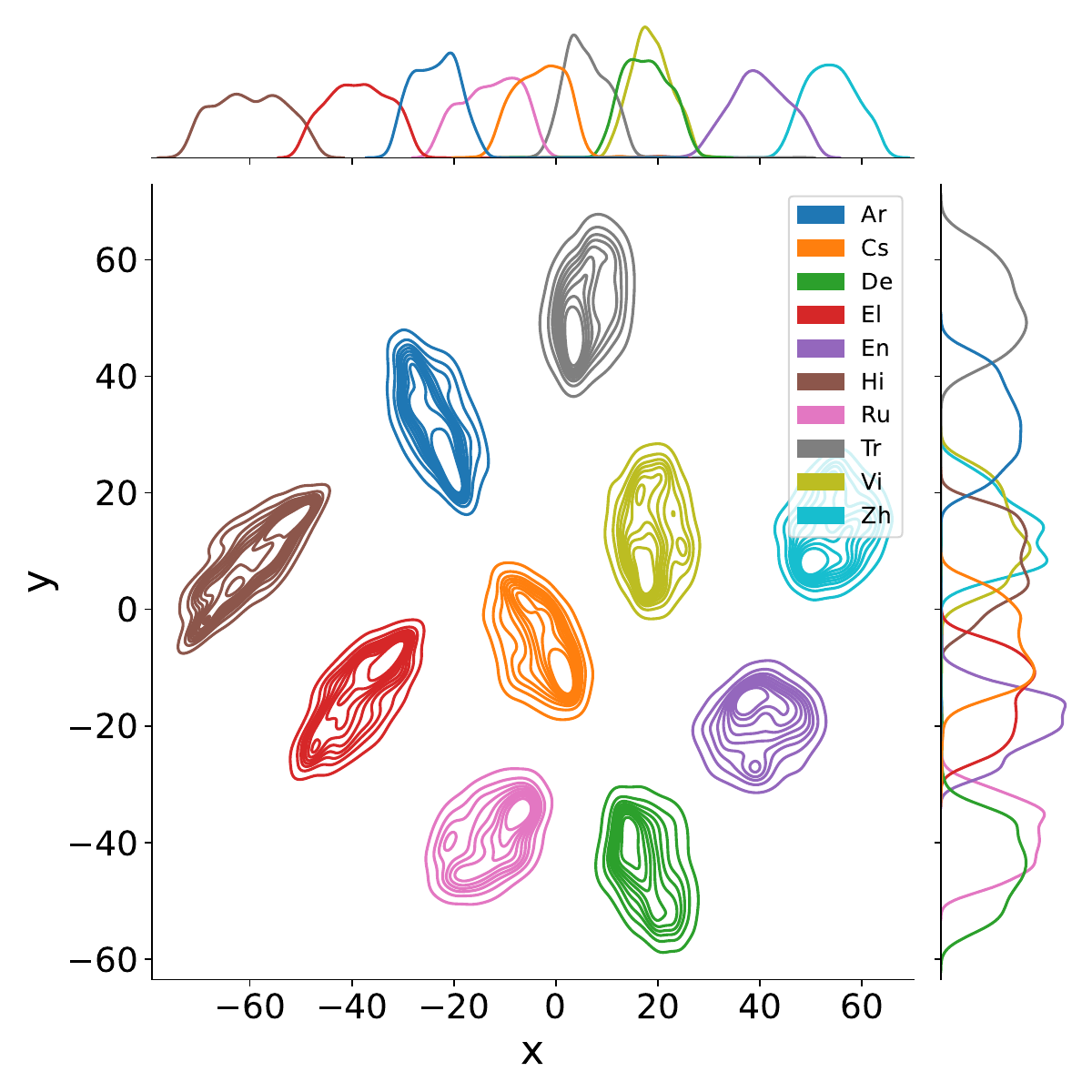}}
    % \subfigure[Layer 3]{\includegraphics[width=1.5in]{images/ours-vis_flores_10langs_layer3.pdf}}
    \subfigure[Layer 4]{\includegraphics[width=1.5in]{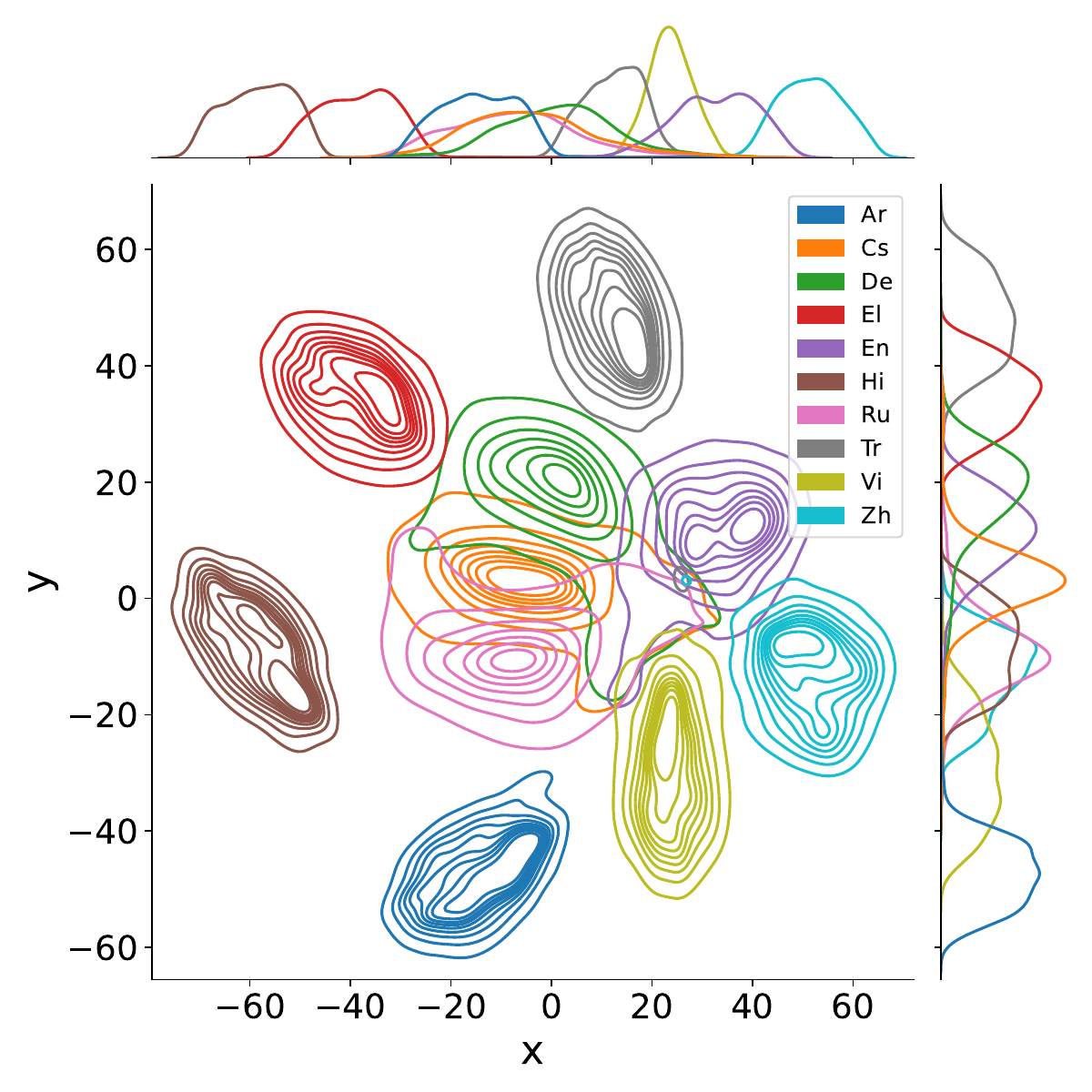}}
    % \subfigure[Layer 5]{\includegraphics[width=1.5in]{images/ours-vis_flores_10langs_layer5.pdf}} 
    \subfigure[Layer 6]{\includegraphics[width=1.5in]{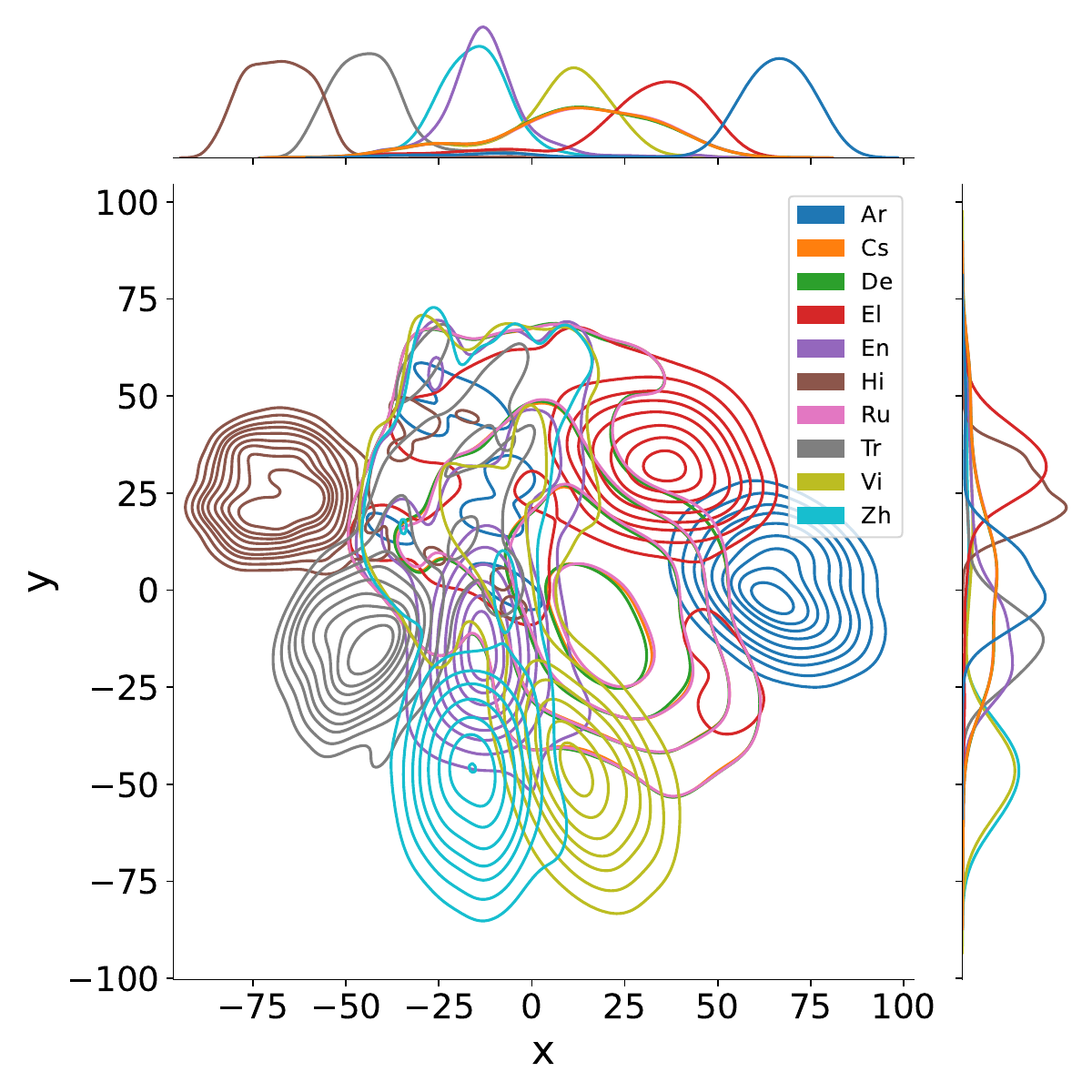}} 
    % \subfigure[Layer 7]{\includegraphics[width=1.5in]{images/ours-vis_flores_10langs_layer7.pdf}}
    \subfigure[Layer 8]{\includegraphics[width=1.5in]{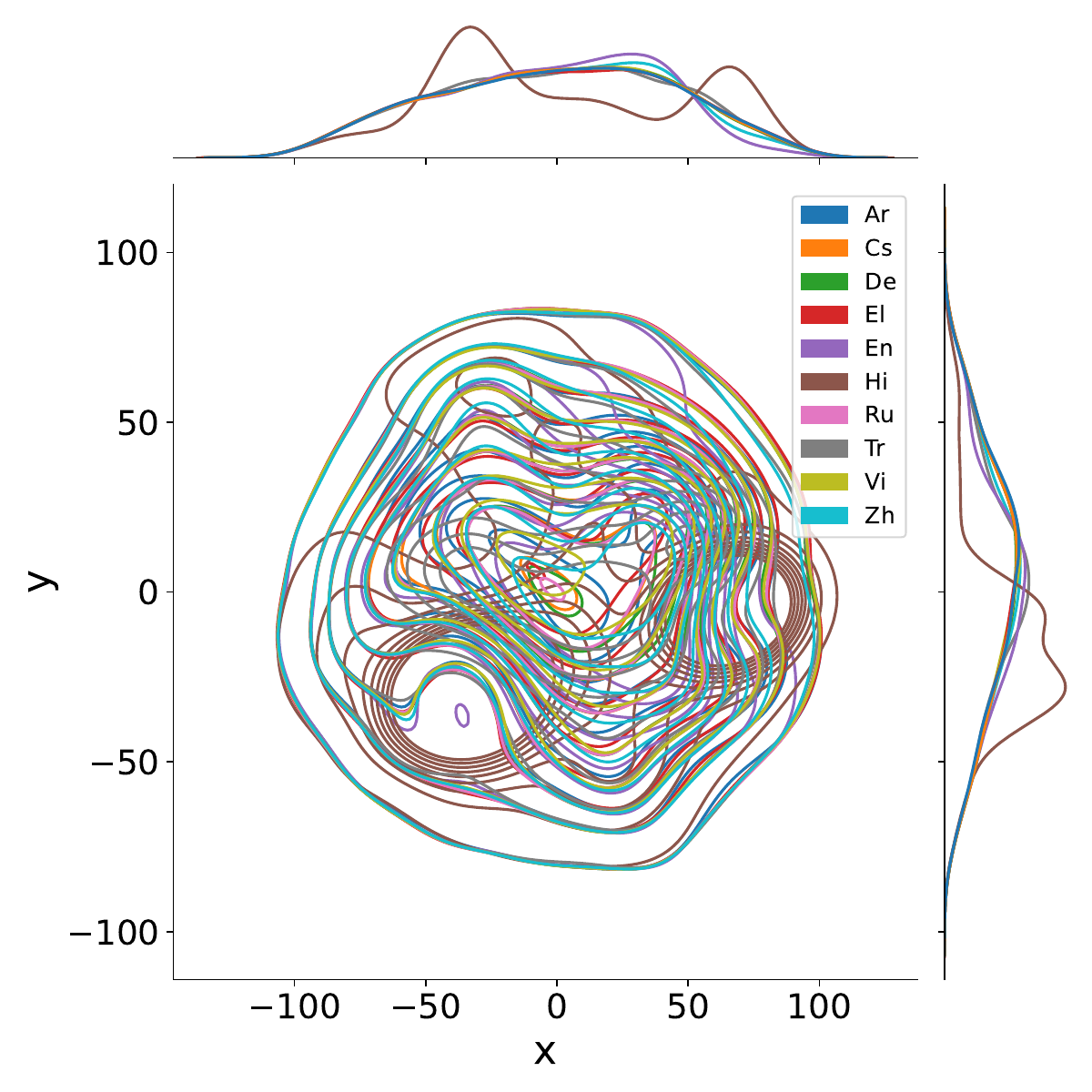}} \\
    % \subfigure[Layer 9]{\includegraphics[width=1.5in]{images/ours-vis_flores_10langs_layer9.pdf}} \\
    \subfigure[Layer 10]{\includegraphics[width=1.5in]{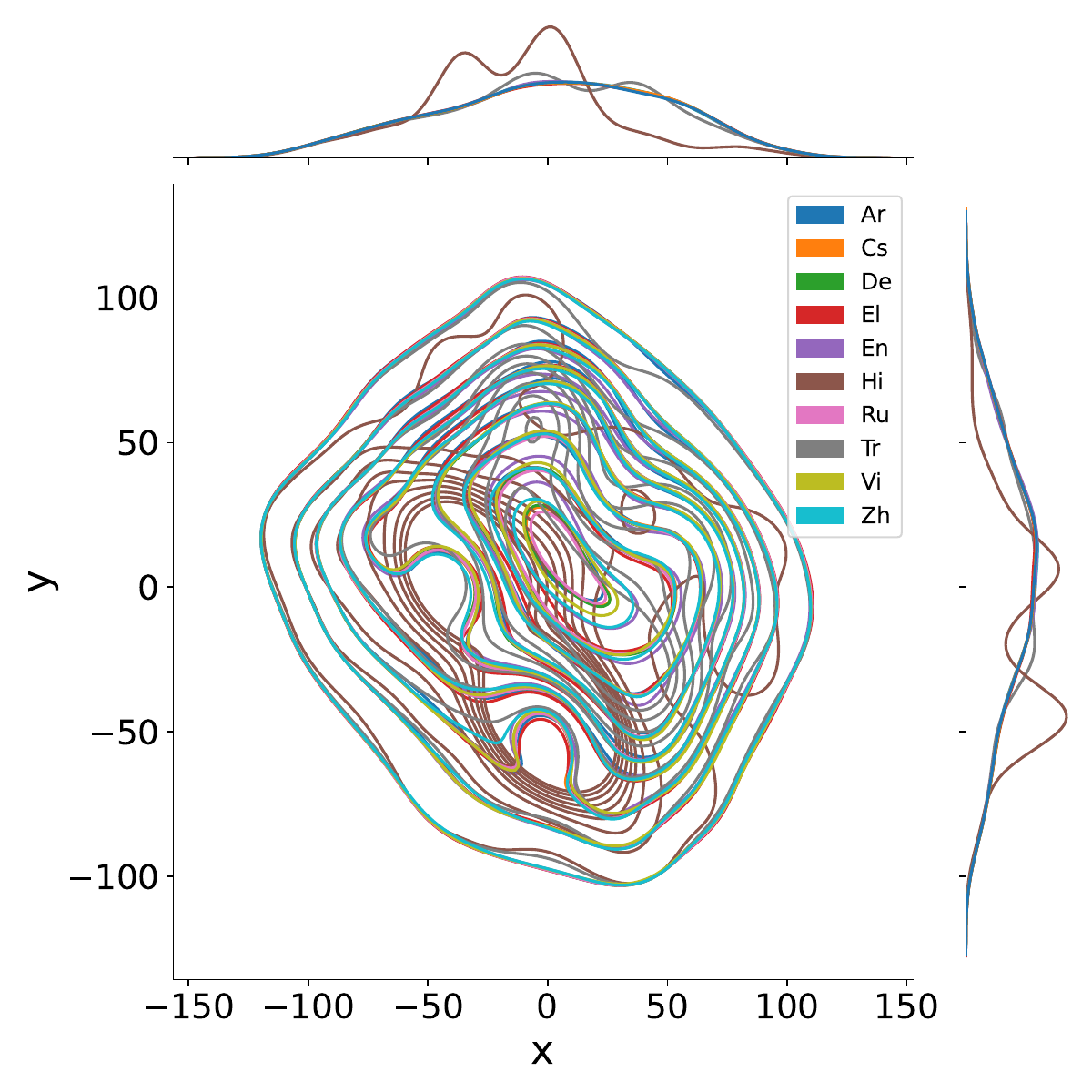}} 
    % \subfigure[Layer 11]{\includegraphics[width=1.5in]{images/ours-vis_flores_10langs_layer11.pdf}} 
    \subfigure[Layer 12]{\includegraphics[width=1.5in]{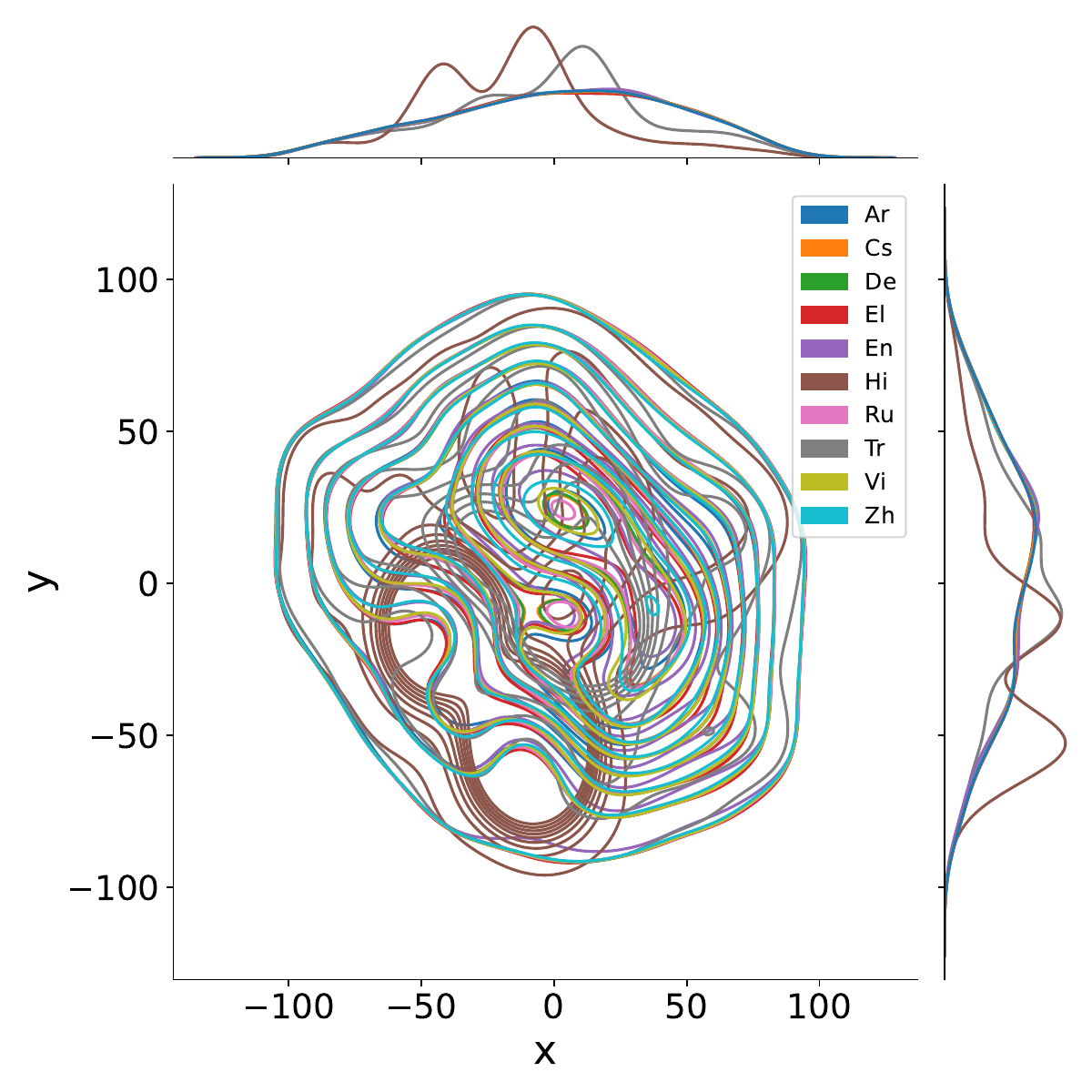}}
    % \subfigure[Layer 13]{\includegraphics[width=1.5in]{images/ours-vis_flores_10langs_layer13.pdf}}
    \subfigure[Layer 14]{\includegraphics[width=1.5in]{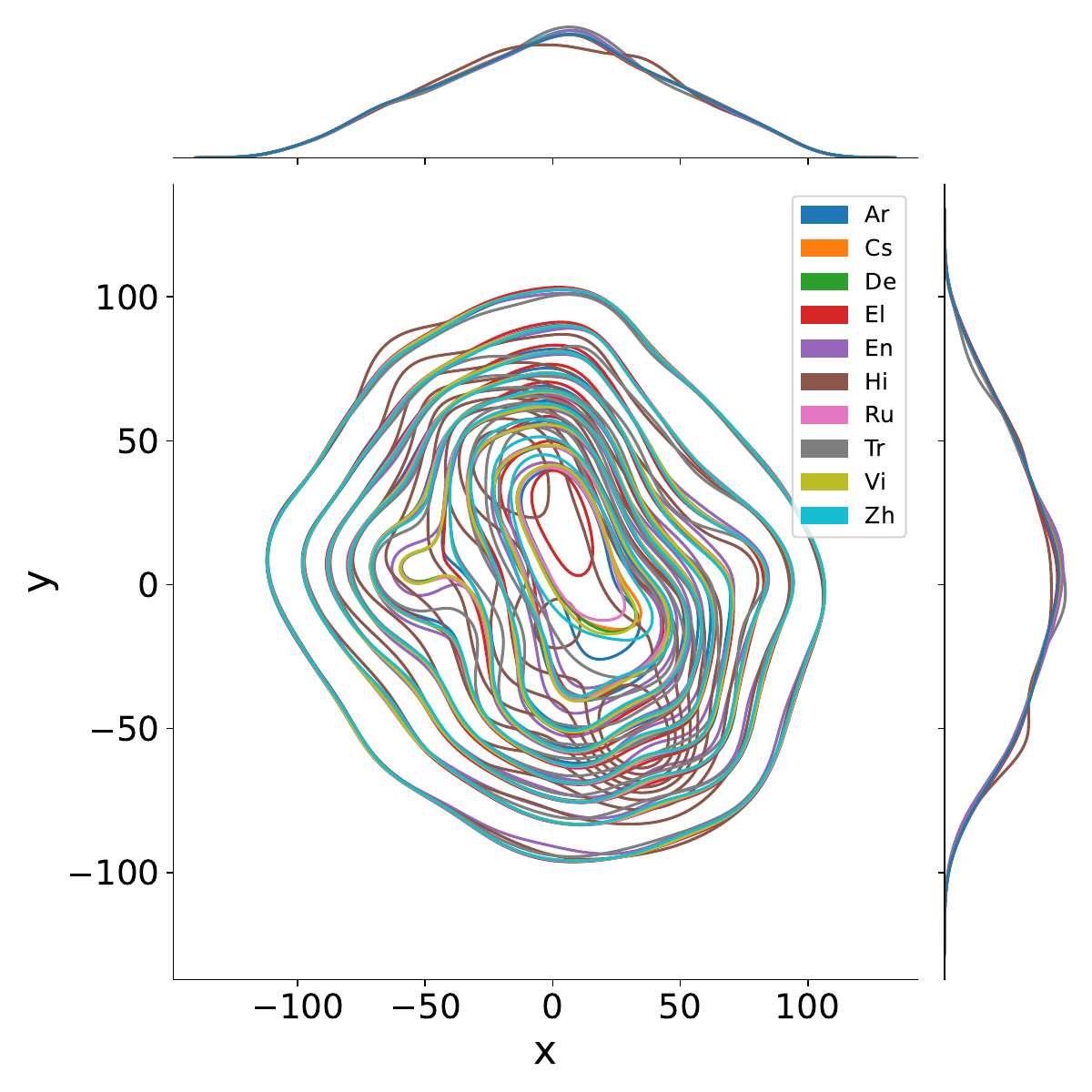}}
    % \subfigure[Layer 15]{\includegraphics[width=1.5in]{images/ours-vis_flores_10langs_layer15.pdf}} 
    \subfigure[Layer 16]{\includegraphics[width=1.5in]{images/ours-vis_flores_10langs_layer16.pdf}} \\
    % \subfigure[Layer 17]{\includegraphics[width=1.5in]{images/ours-vis_flores_10langs_layer17.pdf}}
    \subfigure[Layer 18]{\includegraphics[width=1.5in]{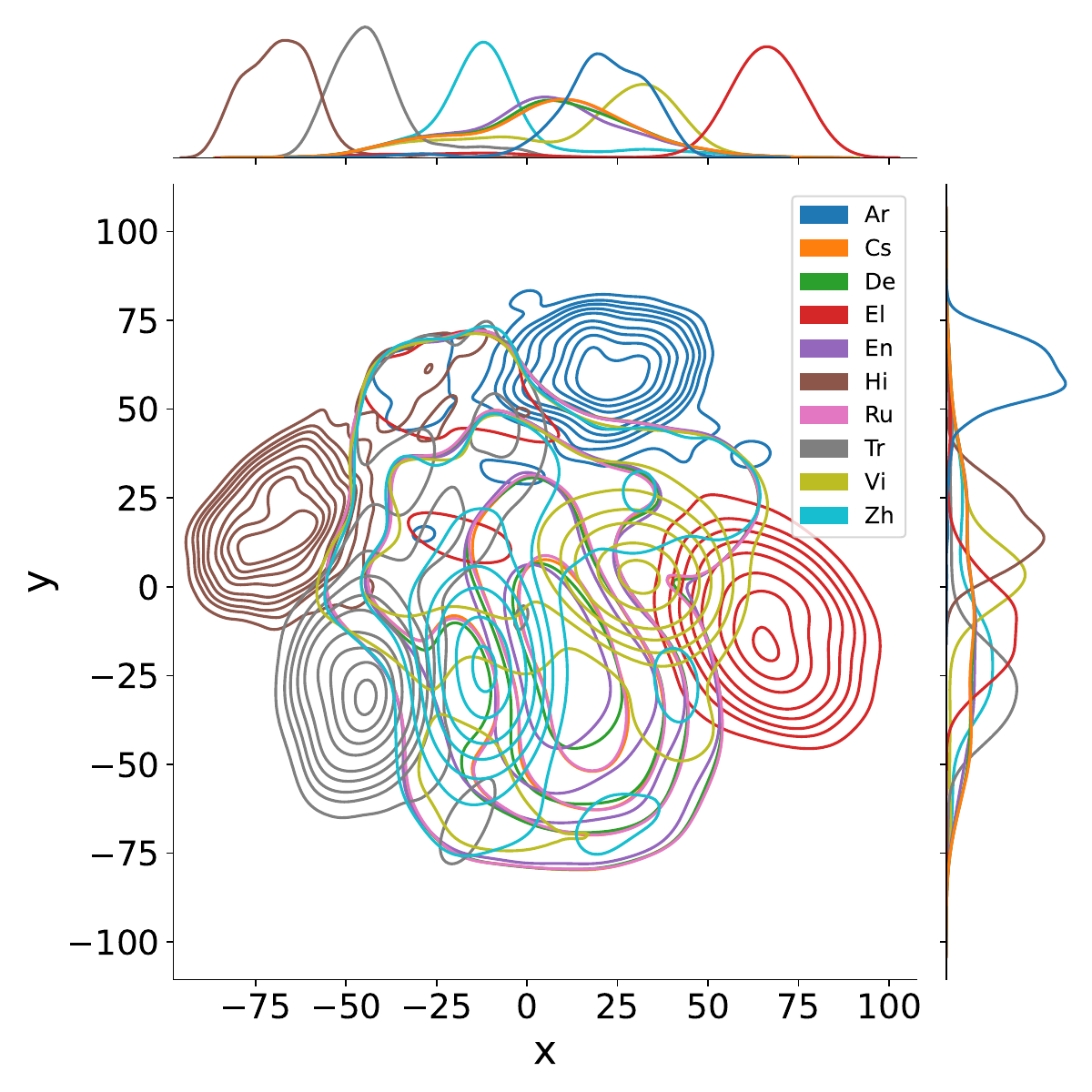}}
    % \subfigure[Layer 19]{\includegraphics[width=1.5in]{images/ours-vis_flores_10langs_layer19.pdf}}
    \subfigure[Layer 20]{\includegraphics[width=1.5in]{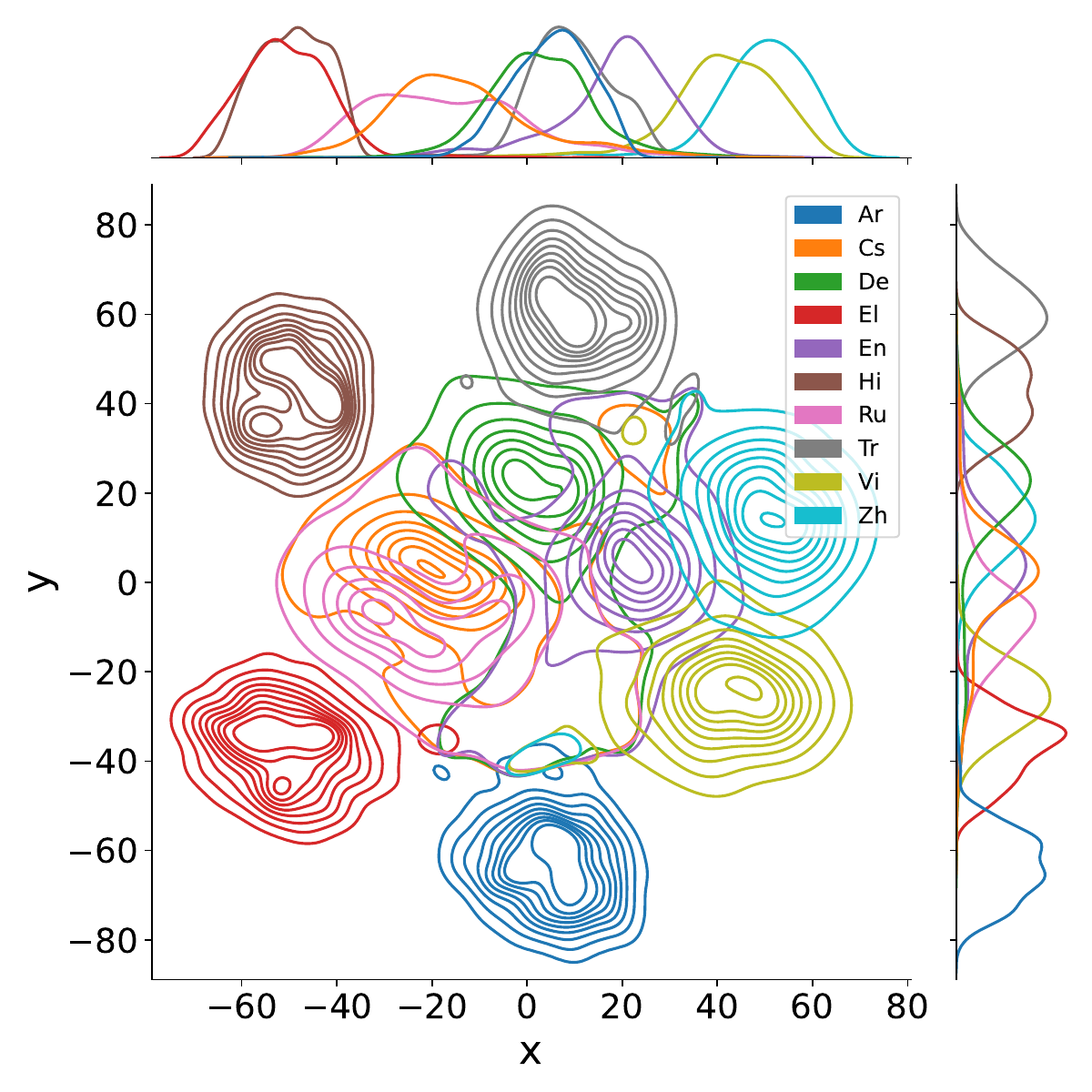}} 
    % \subfigure[Layer 21]{\includegraphics[width=1.5in]{images/ours-vis_flores_10langs_layer21.pdf}} 
    \subfigure[Layer 22]{\includegraphics[width=1.5in]{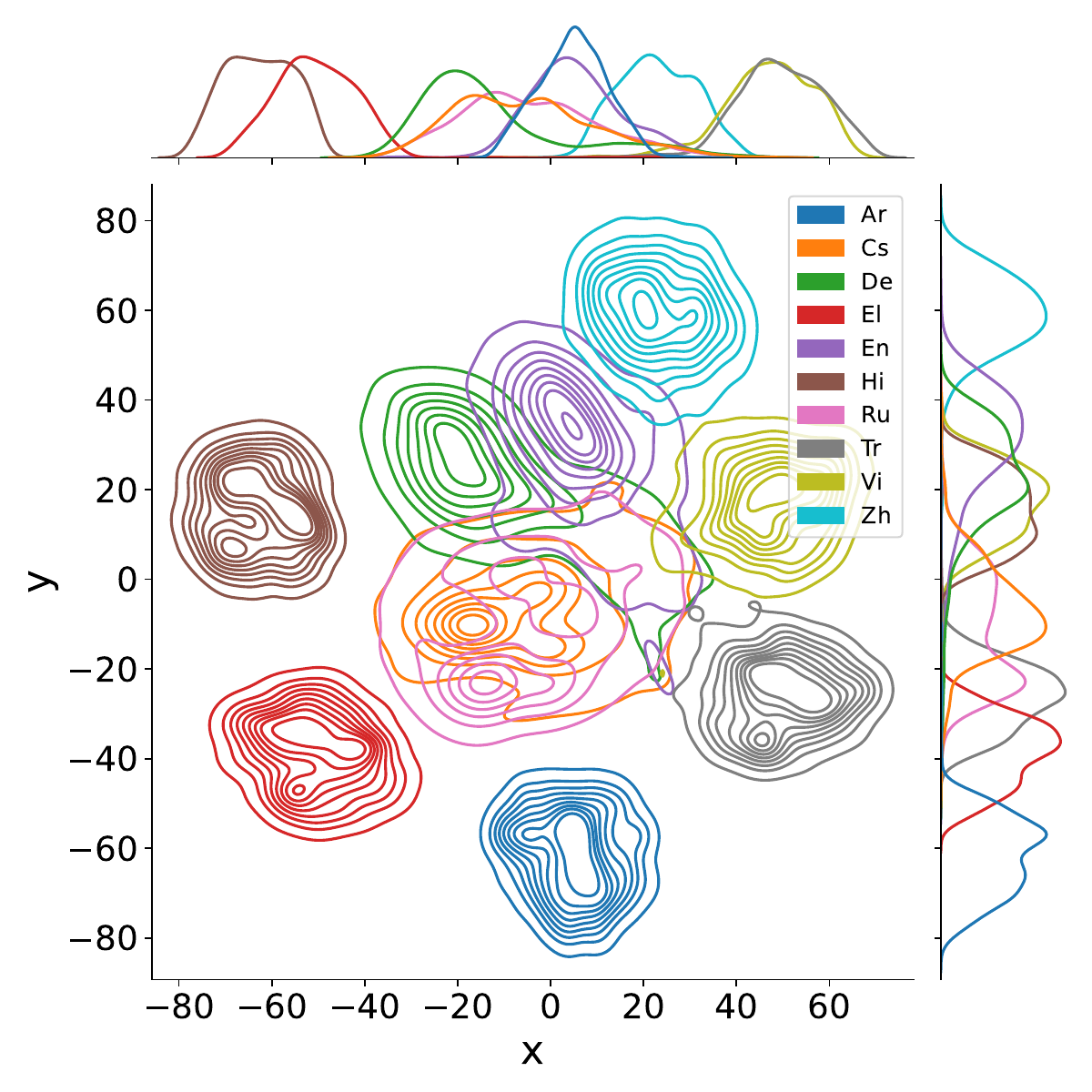}}
    % \subfigure[Layer 23]{\includegraphics[width=1.5in]{images/ours-vis_flores_10langs_layer23.pdf}}
    \subfigure[Layer 24]{\includegraphics[width=1.5in]{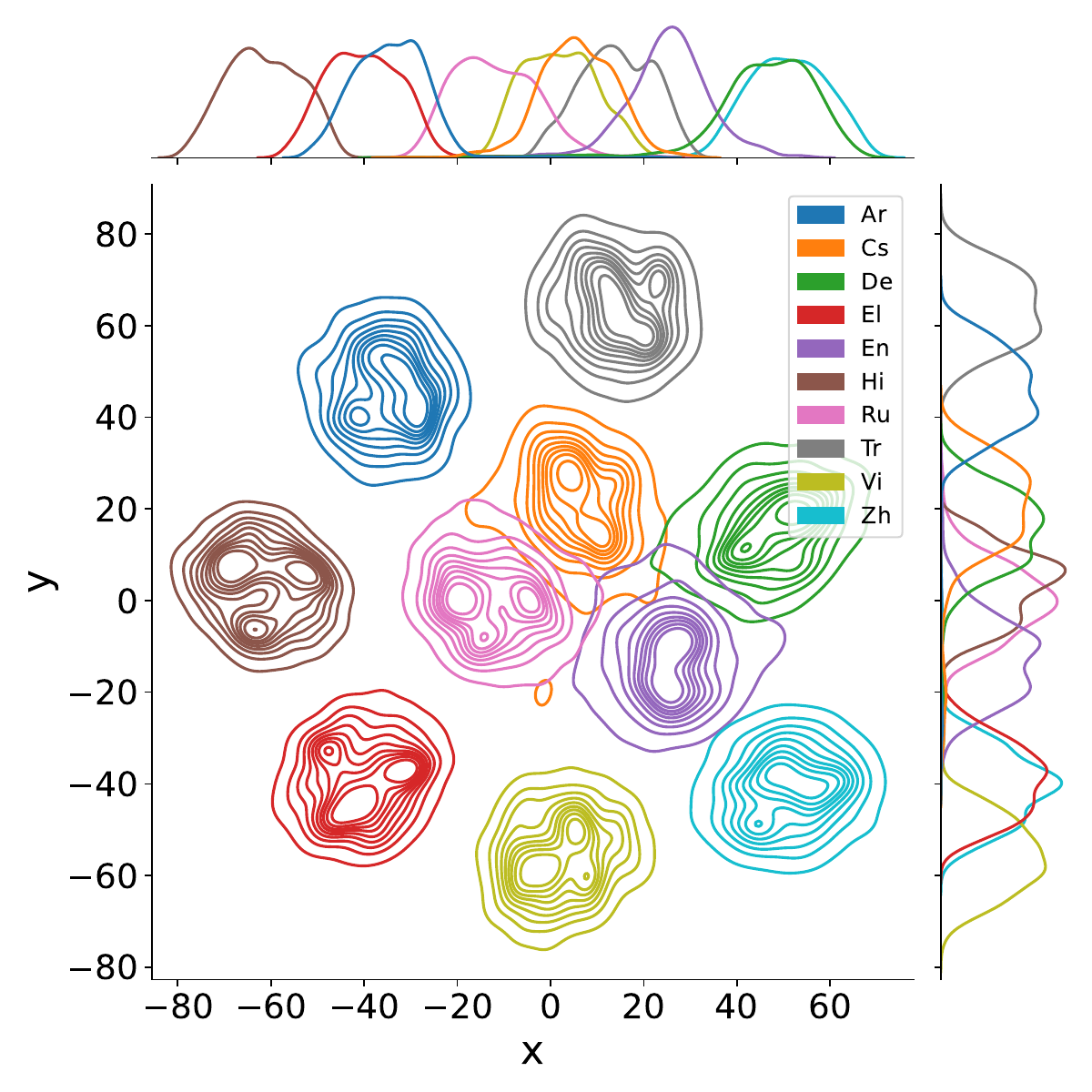}} \\
    % \subfigure[Layer 25]{\includegraphics[width=1.5in]{images/ours-vis_flores_10langs_layer25.pdf}} 
    \subfigure[Layer 26]{\includegraphics[width=1.5in]{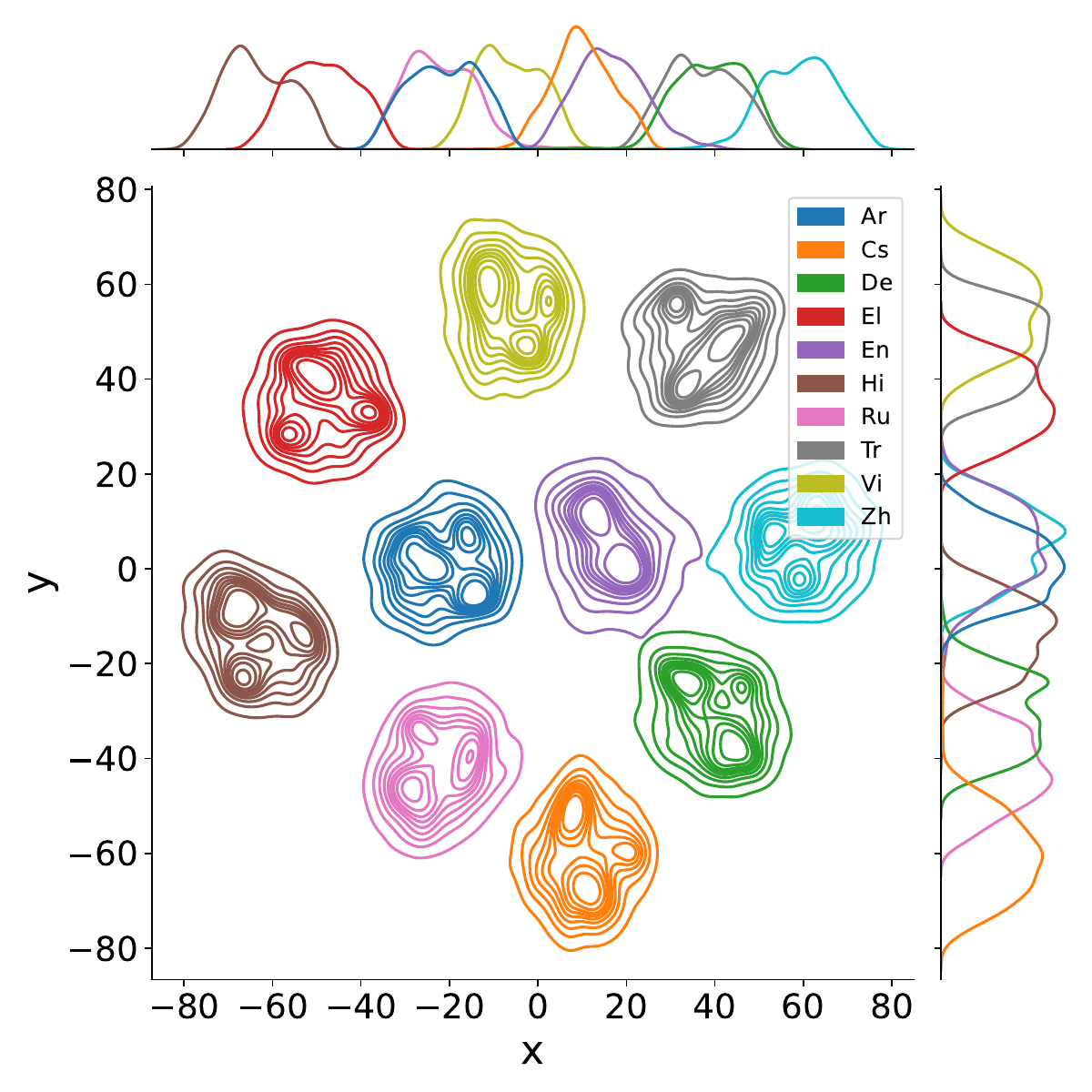}} 
    % \subfigure[Layer 27]{\includegraphics[width=1.5in]{images/ours-vis_flores_10langs_layer27.pdf}}
    \subfigure[Layer 28]{\includegraphics[width=1.5in]{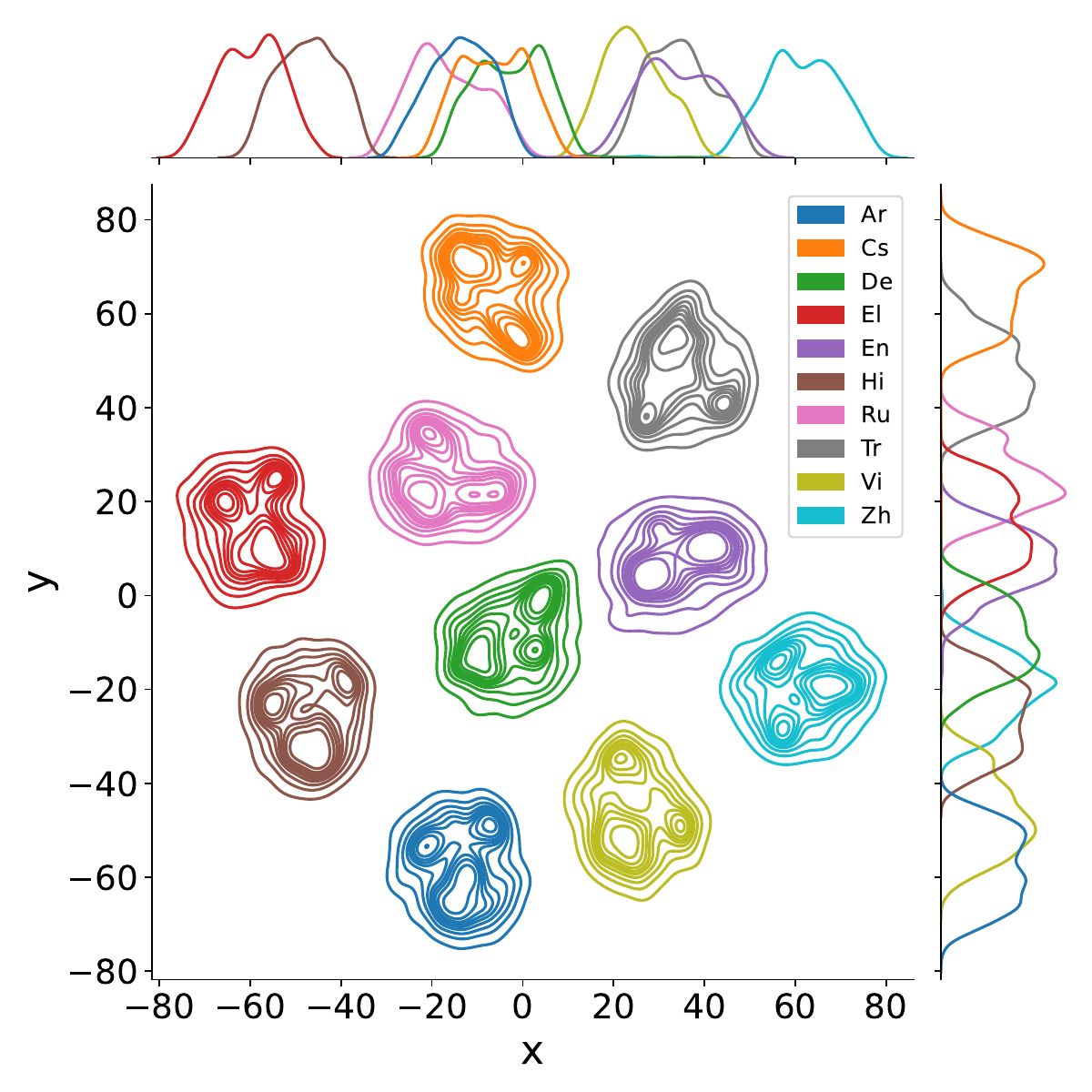}}
    % \subfigure[Layer 29]{\includegraphics[width=1.5in]{images/ours-vis_flores_10langs_layer29.pdf}}
    \subfigure[Layer 30]{\includegraphics[width=1.5in]{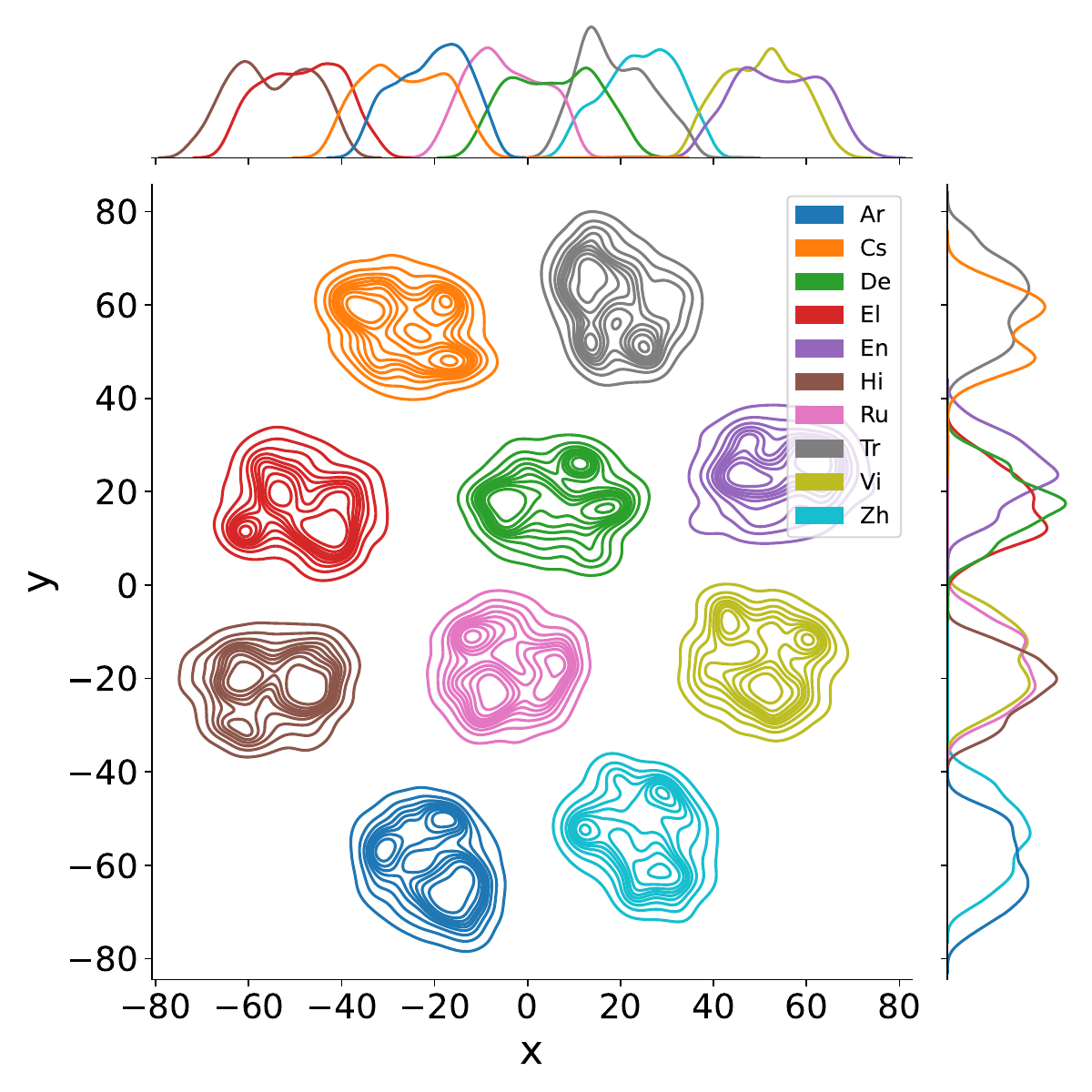}}
    % \subfigure[Layer 31]{\includegraphics[width=1.5in]{images/ours-vis_flores_10langs_layer31.pdf}}
    \subfigure[Layer 32]{\includegraphics[width=1.5in]{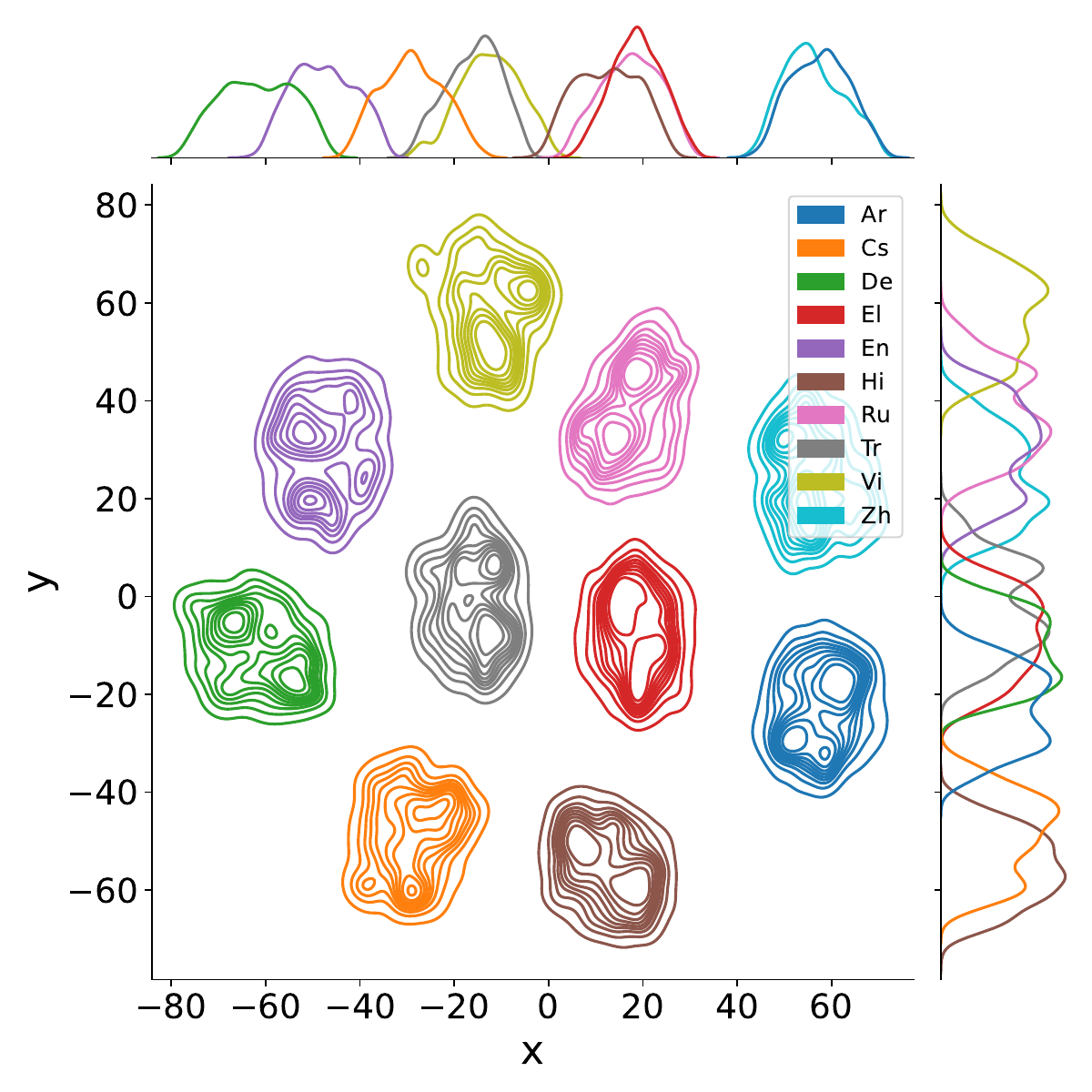}}
    \caption{Visualization of the multilingual representations across layers in AlignX, trained on LLaMA3, after dimension reduction. We leverage the FLORES-101 dev set, which is multi-way parallel.}
    \label{visualization_all_layer_ours}
\end{figure*}

We use the FLORES-101 dev set, with 997 sentences per language. We directly input the sentence into the LLM to obtain the hidden state sequence, perform average pooling to derive the sentence representation, and apply t-SNE~\cite{van2008visualizing} for dimension reduction to 2-dim. Detailed visualization results for both the base LLM and the AlignX LLM, based on LLaMA3-8B, are presented in Figure \ref{visualization_all_layer} and \ref{visualization_all_layer_ours}, respectively.

\section{Detailed Information for Training Dataset}
\label{appendix_detail_information_about_dataset}


\subsection{Data Processing}
\label{appendix_data_processing}
For multilingual translation instruction data, we randomly select English-centric translation pairs from the OPUS-100~\cite{zhang-etal-2020-improving} corpus and sample a translation instruction from the translation instruction set for each pair. For instruction diversity, we prompt ChatGPT to generate a set of ten translation instructions, as shown in Table \ref{translation_set}. For multilingual general instruction data, we primarily draw from the Bactrian-X~\cite{li2023bactrian} dataset, a multilingual version of the Alpaca~\cite{alpaca} dataset available in 52 languages\footnote{Since Greek (El) is not included in Bactrian-X, we obtain the Greek Alpaca dataset from \url{https://github.com/NJUNLP/x-LLM}.}. The instructions are translated using Google Translate, and the responses are generated with GPT-3.5-Turbo. We filter out off-target responses using the \textit{langid} toolkit~\cite{lui-baldwin-2012-langid}. In the first stage, we sample 50k translation pairs per translation direction, resulting in 0.9M translation instruction data in total. In the second stage, we sample 2.5k translation instruction data per translation direction and 10k general instruction data per language, resulting in 145k mixed instruction data in total.

\begin{table*}[htbp]
\centering
\small
\begin{tabular}{c}
\hline
\textbf{Translation Instruction Set} \\ \hline
Translate this text from \{src\_lang\} to \{tgt\_lang\}.\\
Convert this sentence from \{src\_lang\} to \{tgt\_lang\}.\\
Change this paragraph from \{src\_lang\} to \{tgt\_lang\}.\\
Render this message from \{src\_lang\} to \{tgt\_lang\}.\\
Translate this phrase from \{src\_lang\} to \{tgt\_lang\}. \\
Turn this text from \{src\_lang\} to \{tgt\_lang\}.\\
Rewrite this statement from \{src\_lang\} to \{tgt\_lang\}.\\
Provide a translation from \{src\_lang\} to \{tgt\_lang\} for this text.\\
Offer a \{tgt\_lang\} translation for this text from \{src\_lang\}.\\
Give a \{tgt\_lang\} version of this text from \{src\_lang\}.\\ \hline
\end{tabular}
\caption{The translation instruction set in this work. "src\_lang" denotes the source language, and "tgt\_lang" denotes the target language.}
\label{translation_set}
\end{table*}

% 训练语言、语料大小统计
% 翻译指令集
\subsection{Statistics on Corpora Size and Languages}
\label{appendix_statistics_on_corupus_size_and_language}
Table \ref{data_statistics} presents statistics on corpora size for our approach and some typical data-level systems.
%Since many methods have an uneven distribution of training languages, we briefly list the main languages and proportions as follows.
Table \ref{language_statistics} presents information on languages involved in this work.

\begin{table*}[htbp]
\centering
\small
\begin{tabular}{cccc}
\hline
\textbf{Methods}               & \textbf{Languages} & \textbf{Data per Language}     & \textbf{Total Data}           \\ \hline
\textbf{x-LLaMA}~\cite{zhu2023extrapolating}              & 6                  & 736.1K sentences               & 4.4M sentences                \\
\textbf{SDRRL}~\cite{zhang-etal-2024-enhancing-multilingual}                & 15                 & 150K sentences*                & 2.3M sentences*                \\
\textbf{BayLing1}~\cite{zhang2023bayling}             & 4                  & 75.5K sentences                & 302K sentences                \\
\textbf{BayLing2}~\cite{zhang2024bayling}        & 158                & 20.3K sentences                & 3.2M sentences                \\
\textbf{ParroT}~\cite{jiao-etal-2023-parrot}               & 3                  & 67.4K sentences                & 202.2K sentences              \\
\textbf{BigTrans}~\cite{yang2023bigtranslate} & 102                & 880.4M tokens + 2.4k sentences & 89.8B tokens + 241K sentences \\
\textbf{ALMA}~\cite{xu2024a}                 & 6                  & 121.7B tokens + 9.8K sentences & 703B tokens + 58.7K sentences \\ \hdashline
\textbf{AlignX (10langs)}     & 10                 & 104.5K sentences               & 1.0M sentences                \\
\textbf{AlignX (51langs)}     & 51                 & 112.9K sentences               & 5.8M sentences                \\ \hline
\end{tabular}
\caption{Statistics on corpus size for our approach and some typical systems. We list the number of fine-tuned languages, the total corpus size, and the average corpus size per language. “*” indicates that the detailed corpus size is not given in the paper and we estimate a lower bound.}
\label{data_statistics}
\end{table*}

\begin{table*}[htbp]
\centering
\small
\begin{tabular}{cclcclcc}
\cline{1-2} \cline{4-5} \cline{7-8}
\textbf{ISO 639-1} & \textbf{Language} &  & \textbf{ISO 639-1} & \textbf{Language} &  & \textbf{ISO 639-1} & \textbf{Language} \\ \cline{1-2} \cline{4-5} \cline{7-8} 
Af                 & Akrikaans         &  & Hr                 & Croatian          &  & Pl                 & Polish            \\
Ar                 & Arabic            &  & Id                 & Indonesian        &  & Ps                 & Pashto            \\
Az                 & Azerbaijani       &  & It                 & Italian           &  & Pt                 & Portuguese        \\
Bn                 & Bengali           &  & Ja                 & Japanese          &  & Ro                 & Romanian          \\
Cs                 & Czech             &  & Ka                 & Georgian          &  & Ru                 & Russian           \\
De                 & German            &  & Kk                 & Kazakh            &  & Si                 & Sinhala           \\
El                 & Modern Greek      &  & Km                 & Khmer             &  & Sl                 & Slovenian         \\
En                 & English           &  & Ko                 & Korean            &  & Sv                 & Swedish           \\
Es                 & Spanish           &  & Lt                 & Lithuanian        &  & Ta                 & Tamil             \\
Et                 & Estonian          &  & Lv                 & Latvian           &  & Te                 & Telugu            \\
Fa                 & Persian           &  & Mk                 & Macedonian        &  & Th                 & Thai              \\
Fi                 & Finnish           &  & Ml                 & Malayalam         &  & Tr                 & Turkish           \\
Fr                 & French            &  & Mn                 & Mongolian         &  & Uk                 & Ukrainian         \\
Gl                 & Galician          &  & Mr                 & Marathi           &  & Ur                 & Urdu              \\
Gu                 & Gujarati          &  & My                 & Burmese           &  & Vi                 & Vietnamese        \\
He                 & Hebrew            &  & Ne                 & Nepali            &  & Xh                 & Xhosa             \\
Hi                 & Hindi             &  & Nl                 & Dutch             &  & Zh                 & Chinese           \\  \cline{1-2} \cline{4-5} \cline{7-8} 
\end{tabular}
\caption{The languages and corresponding language codes used in this work.}
\label{language_statistics}
\end{table*}

% \begin{table}[htbp]
% \centering
% \begin{tabular}{cccc}
% \hline
% \textbf{Training Stage}                & \textbf{ISO 639-1} & \textbf{Language}      & \textbf{Family}        \\ \hline
% \multirow{10}{*}{Seen}  & Ar        & Arabic        & Afro-Asiatic  \\
%                         & Cs        & Czech         & Indo-European \\
%                         & De        & German        & Indo-European \\
%                         & El        & Greek, Modern & Indo-European \\
%                         & En        & English       & Indo-European \\
%                         & Hi        & Hindi         & Indo-European \\
%                         & Ru        & Russian       & Indo-European \\
%                         & Tr        & Turkish       & Turkic        \\
%                         & Vi        & Vietnamese    & Austroasiatic \\
%                         & Zh        & Chinese       & Sino-Tibetan  \\ \hline
% \multirow{5}{*}{Unseen} & Fr        & French        & Indo-European \\
%                         & He        & Hebrew        & Afro-Asiatic  \\
%                         & It        & Italian       & Indo-European \\
%                         & Ja        & Japanese      & Japonic       \\
%                         & Pl        & Polish        & Indo-European \\ \hline
% \end{tabular}
% \caption{Details of languages in this work. }
% \label{language_statistics}
% \end{table}

\subsection{Details about Evaluation Benchmarks}
\label{appendix_evaluation_benchmarks}
We evaluate multilingual general capabilities and cross-lingual generation capabilities on the following benchmarks:
\begin{itemize}
    \item \textbf{Multilingual TruthfulQA}~\cite{lin-etal-2022-truthfulqa}: This is the multilingual version of the TruthfulQA benchmark and evaluates knowledge and truthfulness capabilities.
    \item \textbf{Multilingual HellaSwag}~\cite{zellers2019HellaSwag}: This is the multilingual version of the HellaSwag benchmark and evaluates commonsense reasoning and contextual understanding capabilities.
    \item \textbf{Cross-lingual Natural Language Inference (XNLI)}~\cite{conneau2018xnli}: This benchmark evaluates  language transfer and cross-lingual sentence classification.
    \item  \textbf{XStoryCloze}~\cite{lin2021few}: This is a multilingual commonsense reasoning benchmark for evaluating story understanding, story generation, and script learning.
    \item \textbf{FLORES-101}~\cite{costa2022no}: This is a multilingual machine translation benchmark and evaluates the cross-lingual generation capability.
\end{itemize}
The multilingual HellaSwag and TruthfulQA benchmarks are obtained from Okapi~\cite{lai-etal-2023-okapi}, translated by ChatGPT. We evaluate the trained languages. In the analysis section, we further experiment on the following benchmarks:
\begin{itemize}
    \item \textbf{IWSLT2017}~\cite{cettolo2017overview}, \textbf{WMT14}\footnote{\url{https://huggingface.co/datasets/wmt/wmt14}} and \textbf{WMT17}\footnote{\url{https://huggingface.co/datasets/wmt/wmt17}}: These benchmarks are classical multilingual translation benchmarks and evaluate cross-lingual generation capabilities of different language subsets.
    \item \textbf{xCSQA}~\cite{lin-etal-2021-common}: This is the multilingual version of CSQA and evaluate cross-lingual commonsense reasoning.
    \item \textbf{xGeo}~\cite{gao-etal-2024-multilingual}: This benchmark evaluates cross-lingual knowledge alignment.
    
\end{itemize}

\subsection{Languages Included in the 51-Language Setup}
\label{appendix_51_languages}
To further validate the effectiveness of AlignX across a broader range of languages, we conduct experiments involving 51 languages. We select languages from the intersection of Bactrian-X and OPUS-100, with the addition of Greek. The languages are: Akrikaans (Af), Arabic (Ar), Azerbaijani (Az), Bengali (Bn), Czech (Cs), German (De), Modern Greek (El), English (En), Spanish (Es), Estonian (Et), Persian (Fa), Finnish (Fi), French (Fr), Galician (Gl), Gujarati (Gu), Hebrew (He), Hindi (Hi), Croatian (Hr), Indonesian (Id), Italian (It), Japanese (Ja), Georgian (Ka), Kazakh (Kk), Khmer (Km), Korean (Ko), Lithuanian (Lt), Latvian (Lv), Macedonian (Mk), Malayalam (Ml), Mongolian (Mn), Marathi (Mr), Burmese (My), Nepali (Ne), Dutch (Nl), Polish (Pl), Pashto (Ps), Portuguese (Pt), Romanian (Ro), Russian (Ru), Sinhala (Si), Slovenian (Sl), Swedish (Sv), Tamil (Ta), Telugu (Te), Thai (Th), Turkish (Tr), Ukrainian (Uk), Urdu (Ur), Vietnamese (Vi), Xhosa (Xh), Chinese (Zh).

\section{More Evaluation Metrics and Detailed Results}
\label{appendix_detailed_results}

\subsection{Multilingual General Benchmarks}
\label{appendix_multilingual_general}
Table \ref{detailed_m_truthfulqa}, \ref{detailed_m_hellaswag} and \ref{detailed_xnli} show results on multilingual TruthfulQA, multilingual HellaSwag, and XNLI benchmark, respectively, and Table \ref{detailed_m_truthfulqa_51langs}, \ref{detailed_m_hellaswag_51langs}, \ref{detailed_xnli_51langs} and \ref{detailed_xstorycloze_51langs} present results on corresponding benchmarks under 51-language setup.

\subsection{Multilingual Translation Benchmark}
\label{appendix_multilingual_translation}
To better evaluate the quality of multilingual translation, we report the results for statistical significance test in Table \ref{table_translation_significance}, and COMET~\cite{rei-etal-2020-comet}\footnote{https://huggingface.co/Unbabel/wmt22-comet-da} metric in Table \ref{table_main_translation_comet}, respectively. We present detailed BLEU results for each base LLMs on the multilingual translation benchmark in Table \ref{detailed_translation_gemma}, \ref{detailed_translation_llama_1}, \ref{detailed_translation_llama_2}, \ref{detailed_translation_llama2} and \ref{detailed_translation_llama3instruct}. We present the averaged BLEU scores under the 51-language setup in Table \ref{table_brief_translation_51langs}.

\section{Case Study for Multilingual Translation}
\label{appendix_case_study}

\begin{figure*}[htbp]
  \centering
  \includegraphics[width=1.0\linewidth]{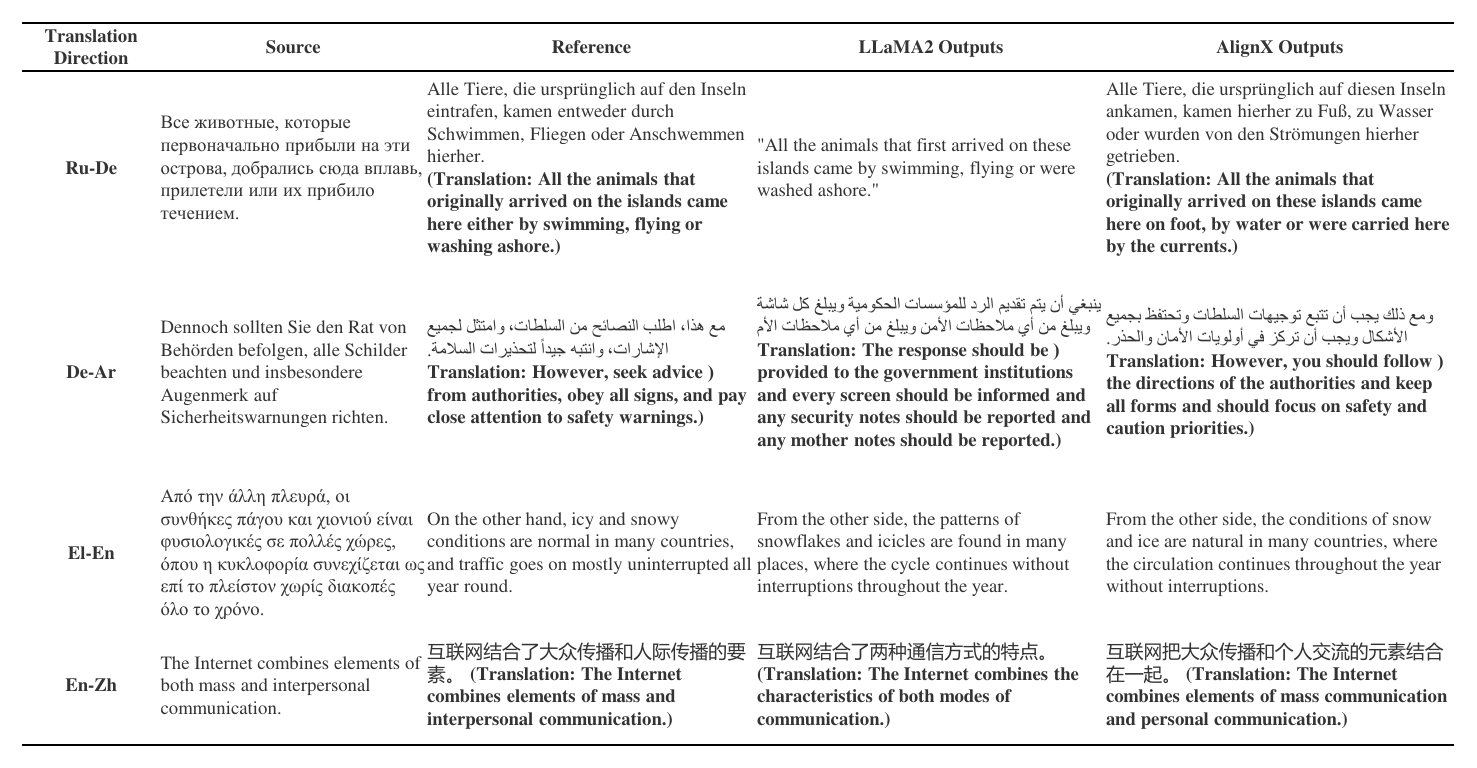}
  \caption{Multilingual translation cases of LLaMA2-7B and AlignX.}
  \label{case_study}
\end{figure*}

To intuitively understand the superiority of AlignX translations, we provide some representative, diverse, and multilingual translation cases of LLaMA2-7B and AlignX. We use online Google Translate to translate non-English text into English for clear understanding. Figure \ref{case_study} presents the cases. While LLaMA2 fails to provide accurate translations, AlignX provides accurate and concise translations. While LLaMA2 fails to provide accurate translations in these cases, including mistranslations, omissions, or even the incorrect target language, AlignX provides accurate and concise translations.

\section{Additional Analysis}
\label{additional_analysis}

\subsection{Comparison with Representation-level AFP}

We further compare our approach with AFP~\cite{li-etal-2024-improving-context}, a SOTA representation-level method. AFP enhances multilingual semantic alignment through an auxiliary contrastive learning objective applied to translation pairs and cross-lingual instruction data. We evaluate AFP on 10 languages, using cross-lingual instructions constructed from approximately 1.15M instances—comparable in scale to AlignX (1.14M instances). Following AFP’s optimal hyperparameters, results in Table~\ref{comparison_between_afp} demonstrate that AlignX outperforms AFP in both multilingual understanding and cross-lingual generation.

We attribute AlignX's superior performance to two key factors. First, AlignX aligns more closely with the LLMs’ inherent “align-then-diverge” pattern by combining contrastive learning with language matching, thereby mitigating the performance degradation in generation when aligning semantic representations alone. Second, AlignX integrates contrastive learning and language matching directly within the instruction format. This encourages the model to naturally learn intra-sentence language shifts, leading to more accurate cross-lingual generation compared to AFP’s cross-lingual instruction tuning.

\begin{table*}[htbp]
\centering
\small
\begin{tabular}{ccccccc}
\hline
                   & \textbf{TruthfulQA} & \textbf{HellaSwag} & \textbf{XNLI}  & \textbf{XStoryCloze} & \textbf{\begin{tabular}[c]{@{}c@{}}FLORES-101 \\ (COMET)\end{tabular}} & \textbf{\begin{tabular}[c]{@{}c@{}}FLORES-101 \\ (OTR)\end{tabular}} \\ \hline

\multicolumn{7}{c}{\textbf{LLaMA-7B   Based}}                                                                                                                   \\ \hline
\textbf{LLaMA-7B}  & 25.91               & 35.44              & 35.70          & 58.75                & 48.25                       & 48.78                     \\
\textbf{AFP}       & 29.00               & 36.18              & 37.30          & 62.64                & 62.17                       & 15.37                     \\
\textbf{AlignX}    & \textbf{31.26}      & \textbf{37.68}     & \textbf{39.24} & \textbf{64.70}       & \textbf{62.78}              & \textbf{8.42}             \\ \hline
\multicolumn{7}{c}{\textbf{LLaMA2-7B   Based}}                                                                                                                   \\ \hline
\textbf{LLaMA2-7B} & 28.35               & 37.79              & 35.92          & 60.50                & 54.34                       & 43.76                     \\
\textbf{AFP}       & 31.79               & 38.35              & 36.31          & 62.66                & 64.11                       & 17.68                     \\
\textbf{AlignX}    & \textbf{32.99}      & \textbf{38.99}     & \textbf{37.76} & \textbf{64.29}       & \textbf{68.64}              & \textbf{8.07}             \\ \hline
\end{tabular}
\caption{Comparison between AFP and AlignX. We report averaged Accuracy, COMET, and OTR (off-target ratio) scores under the 10-language setup. We bold the best results.}
\label{comparison_between_afp}
\end{table*}

\subsection{Efficiency Analysis}

To evaluate the training efficiency, we compare the training cost between AlignX and CPT-then-SFT, using the same two-stage dataset but without the two auxiliary tasks. All experiments are conducted under identical hardware and hyperparameter settings, including 4$\times$A800-80G GPUs, fixed training steps, batch size, and micro-batch size. As shown in Table~\ref{efficiency_analysis}, AlignX incurs only minimal additional cost, as the auxiliary tasks are integrated into the same forward pass during Stage 1. Stage 2 and inference incur no further overhead, as no extra parameters or computations are involved beyond CPT-then-SFT.  

\begin{table}[t]
\centering
\small
\begin{tabular}{cc}
\hline
\textbf{Relative Speed} & \textbf{Training (Stage 1)} \\ \hline
CPT+SFT                 & 1.00$\times$                \\
AlignX                  & 0.95$\times$                \\ \hline
\end{tabular}
\caption{Training efficiency comparison. AlignX introduces negligible additional cost compared to the baseline.}
\label{efficiency_analysis}
\end{table}

\subsection{Detailed GPT-4o Evaluation Results}
\label{details_gpt4o_evaluation}

\begin{figure*}[htbp]
  \centering
  \includegraphics[width=1.0\linewidth]{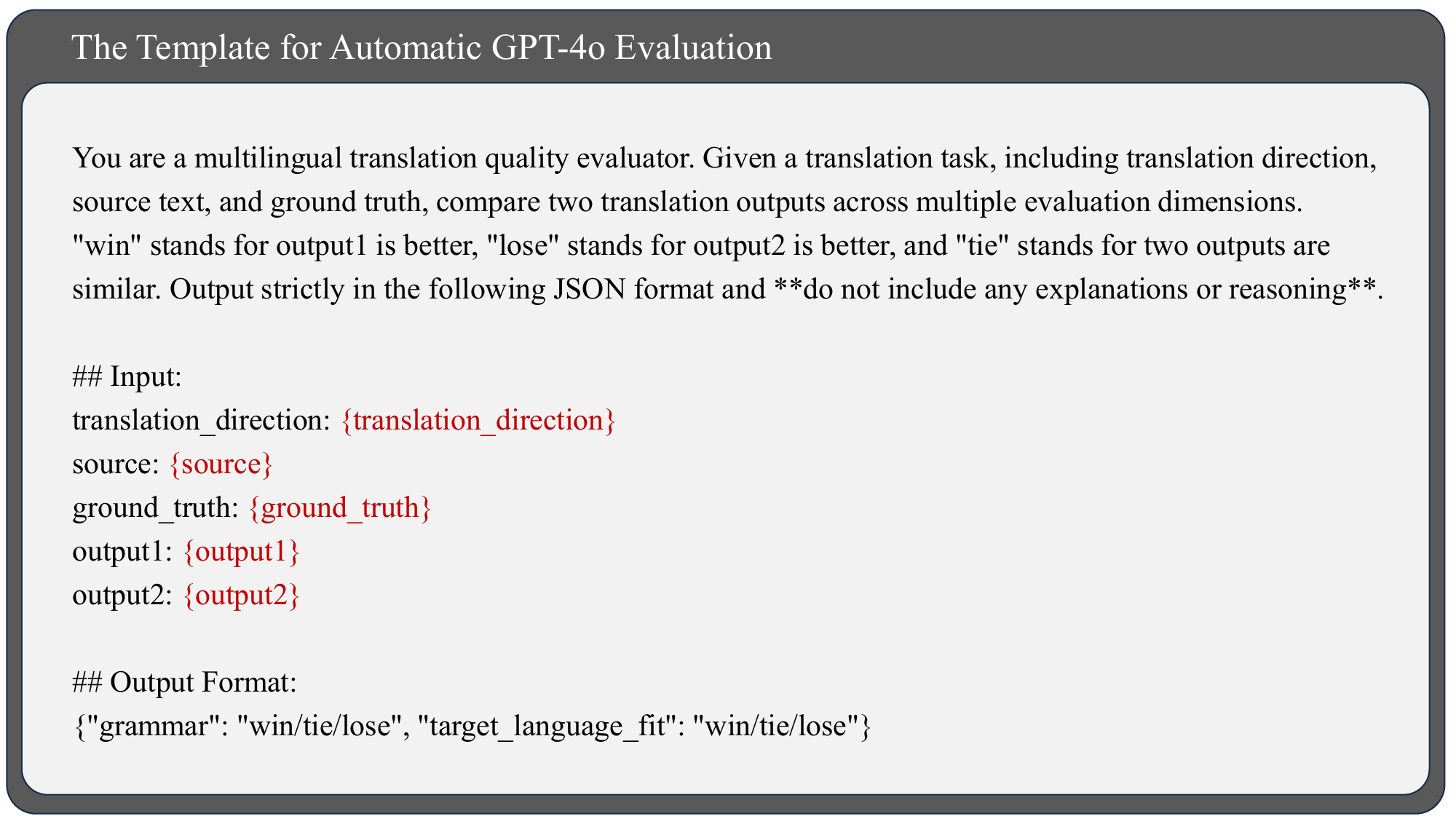}
  \caption{Template used for automatic GPT-4o evaluation. We randomly stack the translation outputs from LLaMA-8B-Instruct and AlignX, treating them as output1 and output2.}
  \label{gpt4o_template}
\end{figure*}

We present the GPT-4o evaluation template in Figure~\ref{gpt4o_template} and provide detailed results in Table~\ref{gpt4o_evaluation_details}, grouped by target language. 
For high-resource languages (e.g., English and German), GPT-4o predominantly produces neutral judgments. 
In contrast, for low-resource languages (e.g., Arabic, Greek, and Czech), GPT-4o shows a clear preference for AlignX outputs. These findings highlight the effectiveness of the language-matching task, which proves particularly beneficial in low-resource scenarios.

\begin{table}[t]
\centering
\small
\begin{tabular}{ccc}
\hline
\textbf{W/T/L} & \textbf{Grammar}  & \textbf{Language Fit} \\ \hline
\textbf{X-En}  & 52 / 394 / 4      & 80 / 359 / 11         \\
\textbf{X-Ar}  & 185 / 191 / 74    & 191 / 165 / 94        \\
\textbf{X-Cs}  & 127 / 264 / 59    & 135 / 248 / 67        \\
\textbf{x-De}  & 103 / 325 / 22    & 102 / 309 / 39        \\
\textbf{X-El}  & 173 / 174 / 103   & 192 / 138 / 120       \\
\textbf{X-Hi}  & 149 / 203 / 98    & 144 / 179 / 127       \\
\textbf{X-Ru}  & 81 / 333 / 36     & 93 / 320 / 37         \\
\textbf{X-Tr}  & 108 / 277 / 65    & 123 / 238 / 89        \\
\textbf{X-Vi}  & 53 / 380 / 17     & 80 / 339 / 31         \\
\textbf{X-Zh}  & 67 / 360 / 23     & 83 / 316 / 51         \\
\textbf{All}   & 1098 / 2901 / 501 & 1223 / 2611 / 666     \\ \hline
\end{tabular}
\caption{Detailed automatic evaluation results using GPT-4o on cross-lingual generation. We compare paired outputs of AlignX and LLaMA3-8B-Instruct, where “W/T/L” indicates that AlignX produces better, comparable, or worse translations, respectively. "X" denotes all other nine languages except the target language.}
\label{gpt4o_evaluation_details}
\end{table}

% 翻译和通用的具体结果
\begin{table*}[htbp]
\centering
\small
\begingroup
\begin{tabular}{ccccccccc}
\hline
\textbf{Methods}            & \textbf{Ar}    & \textbf{De}    & \textbf{En}    & \textbf{Hi}    & \textbf{Ru}    & \textbf{Vi}    & \textbf{Zh}    & \multicolumn{1}{l}{\textbf{Avg.}} \\ \hline
\multicolumn{9}{c}{\textbf{Gemma-2B Based}}                                                                                                                                               \\ \hline
\textbf{Gemma-2B}           & 27.94 & 26.40 & 23.01 & \textbf{26.52} & 26.78 & 27.52 & 26.65 & 26.40 \\
\textbf{CPT-then-SFT}           & 31.18 & \textbf{29.44} & \textbf{27.54} & 26.26 & 30.46 & 29.68 & \textbf{29.44} & \textbf{29.14} \\
\textbf{AlignX}           & \textbf{32.08} & 29.19 & 26.93 & 25.49 & \textbf{30.58} & \textbf{30.06} & 28.81 & 29.02                              \\ \hline
\multicolumn{9}{c}{\textbf{Mistral-7B-v0.3 Based}}                                                                                                                                               \\ \hline
\textbf{Mistral-7B-v0.3}           & 30.79 & 31.22 & 29.01 & 27.30 & 31.22 & 30.19 & 29.31 & 29.86 \\
\textbf{CPT-then-SFT}           & 36.22 & \textbf{37.82} & 36.60 & 29.24 & 38.45 & 35.41 & \textbf{35.03} & 35.54  \\
\hline
\end{tabular}
\endgroup
\caption{Detailed results on multilingual TruthfulQA benchmark. "Avg." denotes average scores on all languages. We bold the highest scores.}
\label{detailed_m_truthfulqa}
\end{table*}

\begin{table*}[htbp]
\centering
\small
\begingroup
\begin{tabular}{ccccccccc}
\hline
\textbf{Methods}            & \textbf{Ar}    & \textbf{De}    & \textbf{En}    & \textbf{Hi}    & \textbf{Ru}    & \textbf{Vi}    & \textbf{Zh}    & \multicolumn{1}{l}{\textbf{Avg.}} \\ \hline
\multicolumn{9}{c}{\textbf{Gemma-2B Based}}                                                                                                                                               \\ \hline
\textbf{Gemma-2B}           & 32.12 & 36.94 & 52.60 & 29.74 & 36.39 & 35.60 & 35.82 & 37.03                              \\
\textbf{CPT-then-SFT}           & 32.14 & 37.50 & 53.01 & 29.85 & 36.65 & 35.93 & 36.37 & 37.35 \\
\textbf{AlignX}           & \textbf{32.74} & \textbf{37.83} & \textbf{53.22} & \textbf{30.10} & \textbf{37.17} & \textbf{36.32} & \textbf{36.68} & \textbf{37.72}                              \\ \hline
\multicolumn{9}{c}{\textbf{Mistral-7B-v0.3 Based}}                                                                                                                                               \\ \hline
\textbf{Mistral-7B-v0.3}           & 32.73 & 43.96 & 61.80 & 29.42 & 43.24 & 35.73 & 40.99 & 41.12 \\
\textbf{CPT-then-SFT}           & 35.21 & 43.83 & 61.82 & 30.77 & 43.01 & 37.58 & 40.47 & 41.81 \\
\hline
\end{tabular}
\endgroup
\caption{Detailed results on multilingual HellaSwag benchmark. "Avg." denotes average scores on all languages. We bold the highest scores.}
\label{detailed_m_hellaswag}
\end{table*}

\begin{table*}[htbp]
\centering
\begingroup
\small
\begin{tabular}{ccccccccccc}
\hline
\textbf{Methods}            & \textbf{Ar}    & \textbf{De}    & \textbf{El}    & \textbf{En}    & \textbf{Hi}    & \textbf{Ru}    & \textbf{Tr}    & \textbf{Vi}    & \textbf{Zh}    & \textbf{Avg.}  \\ \hline
\multicolumn{11}{c}{\textbf{Gemma-2B Based}}                                                                                                                                                             \\ \hline
\textbf{Gemma-2B}           & 33.29 & 41.29 & 33.49 & 44.86 & 36.79 & 38.19 & 36.43 & 36.39 & 35.82 & 37.39           \\
\textbf{CPT-then-SFT}           & \textbf{33.86} & 42.33 & 33.61 & 43.45 & 37.19 & 39.68 & 37.75 & 38.39 & 36.02 & 38.03  \\
\textbf{AlignX}           & 33.49 & \textbf{42.45} & \textbf{34.34} & \textbf{45.38} & \textbf{38.19} & \textbf{40.20} & \textbf{39.32} & \textbf{39.24} & \textbf{36.99} & \textbf{38.84}           \\ \hline
\multicolumn{11}{c}{\textbf{Mistral-7B-v0.3 Based}}                                                                                                                                                             \\ \hline
\textbf{Mistral-7B-v0.3}           & 33.90 & \textbf{43.33} & 33.86 & 43.69 & 36.55 & 38.63 & 36.27 & 36.71 & 37.79 & 37.86 \\
\textbf{CPT-then-SFT}           & 34.66 & 43.45 & 34.34 & 43.45 & 39.44 & 41.04 & 37.91 & 38.96 & 38.39 & 39.07 \\
\hline
\end{tabular}
\endgroup
\caption{Detailed results on XNLI benchmark. "Avg." denotes average scores on all languages. We bold the highest scores.}
\label{detailed_xnli}
\end{table*}

\begin{table*}[htbp]
\centering
\small
\begin{tabular}{ccccccc}
\hline
\textbf{Methods}          & \textbf{Ar}    & \textbf{En}    & \textbf{Hi}    & \textbf{Ru}    & \textbf{Zh}    & \textbf{Avg.}  \\ \hline
\multicolumn{7}{c}{\textbf{Gemma-2B Based}}                                                                                         \\ \hline
\textbf{Gemma-2B}             & 53.61          & 75.65          & 57.71          & 61.35          & 58.70          & 61.40          \\
\textbf{CPT-then-SFT}             & 55.66 & 78.23 & 59.43 & 63.27 & \textbf{61.61} & 63.64 \\
\textbf{AlignX}            & \textbf{55.86} & \textbf{78.42} & \textbf{60.29} & \textbf{63.86} & 61.15 & \textbf{63.92} \\ \hline
\multicolumn{7}{c}{\textbf{Mistral-7B-v0.3 Based}}                                                                                         \\ \hline
\textbf{Mistral-7B-v0.3}             & 51.36 & 80.15 & 54.86 & 65.32 & 63.34 & 63.00 \\
\textbf{CPT-then-SFT}             & \textbf{59.43} & 82.66 & 57.71 & 69.29 & 66.58 & 67.13 \\
\hline
\end{tabular}
\caption{Detailed results on XStoryCloze benchmark. "Avg." denotes average scores on all languages. We bold the highest scores.}
\label{detailed_xstorycloze}
\end{table*}
\begin{table*}[htbp]
\centering
\small
\begin{tabular}{cccccccccc}
\hline
\textbf{TruthfulQA} & \textbf{Ar}    & \textbf{Bn}    & \textbf{De}    & \textbf{En}    & \textbf{Es}    & \textbf{Fr}    & \textbf{Gu}    & \textbf{Hi}          & \textbf{Hr}          \\ \hline
\textbf{LLaMA3-8B-Instruct}      & 34.54          & 29.71          & 34.01          & 35.74          & 36.76          & 35.83          & 27.63          & 30.53                & 33.55                \\
\textbf{AlignX (51langs)}        & \textbf{35.58} & \textbf{30.86} & \textbf{35.66} & \textbf{38.80} & \textbf{37.90} & \textbf{39.01} & \textbf{28.32} & \textbf{30.66}       & \textbf{34.59}       \\ \hline
\multicolumn{1}{l}{}             & \textbf{Id}    & \textbf{It}    & \textbf{Ml}    & \textbf{Mr}    & \textbf{Ne}    & \textbf{Nl}    & \textbf{Pt}    & \textbf{Ro}          & \textbf{Ru}          \\ \hline
\textbf{LLaMA3-8B-Instruct}      & 34.32          & 35.89          & 28.53          & 28.53          & 29.33          & 37.20          & 37.44          & 35.04                & 36.29                \\
\textbf{AlignX (51langs)}        & \textbf{37.66} & \textbf{38.06} & \textbf{28.82} & \textbf{32.33} & \textbf{29.72} & \textbf{38.47} & \textbf{39.09} & \textbf{36.97}       & \textbf{39.85}       \\ \hline
\multicolumn{1}{l}{}             & \textbf{Sv}    & \textbf{Ta}    & \textbf{Te}    & \textbf{Uk}    & \textbf{Vi}    & \textbf{Zh}    & \textbf{Avg.}  & \multicolumn{1}{l}{} & \multicolumn{1}{l}{} \\ \cline{1-8}
\textbf{LLaMA3-8B-Instruct}      & 37.60          & 27.46          & 26.24          & 35.84          & 35.41          & 35.53          & 33.29          & \multicolumn{1}{l}{} & \multicolumn{1}{l}{} \\
\textbf{AlignX (51langs)}        & \textbf{36.05} & \textbf{27.73} & \textbf{26.67} & \textbf{36.49} & \textbf{40.89} & \textbf{37.44} & \textbf{34.90} & \multicolumn{1}{l}{} & \multicolumn{1}{l}{} \\ \cline{1-8}
\end{tabular}
\caption{Detailed results on multilingual TruthfulQA benchmark under the 51-language setup. We bold the highest scores.}
\label{detailed_m_truthfulqa_51langs}
\end{table*}

\begin{table*}[htbp]
\centering
\small
\begin{tabular}{cccccccccc}
\hline
\textbf{HellaSwag} & \textbf{Ar}    & \textbf{Bn}    & \textbf{De}    & \textbf{En}    & \textbf{Es}    & \textbf{Fr}    & \textbf{Gu}    & \textbf{Hi}          & \textbf{Hr}          \\ \hline
\textbf{LLaMA3-8B-Instruct}      & 36.54          & 29.08          & 44.24          & 57.61          & 47.97          & 46.78          & 28.06          & \textbf{33.63}       & 37.73                \\
\textbf{AlignX (51langs)}        & \textbf{37.59} & \textbf{29.61} & \textbf{45.73} & \textbf{59.66} & \textbf{49.79} & \textbf{48.13} & \textbf{28.31} & 33.52                & \textbf{39.78}       \\ \hline
\multicolumn{1}{l}{}             & \textbf{Id}    & \textbf{It}    & \textbf{Ml}    & \textbf{Mr}    & \textbf{Ne}    & \textbf{Nl}    & \textbf{Pt}    & \textbf{Ro}          & \textbf{Ru}          \\ \hline
\textbf{LLaMA3-8B-Instruct}      & 41.73          & 45.34          & 26.43          & 27.86          & 27.86          & 44.79          & 46.05          & 41.45                & 42.76                \\
\textbf{AlignX (51langs)}        & \textbf{43.27} & \textbf{46.25} & \textbf{26.61} & \textbf{28.24} & \textbf{28.66} & \textbf{45.87} & \textbf{48.01} & \textbf{42.62}       & \textbf{43.98}       \\ \hline
\multicolumn{1}{l}{}             & \textbf{Sv}    & \textbf{Ta}    & \textbf{Te}    & \textbf{Uk}    & \textbf{Vi}    & \textbf{Zh}    & \textbf{Avg.}  & \multicolumn{1}{l}{} & \multicolumn{1}{l}{} \\ \cline{1-8}
\textbf{LLaMA3-8B-Instruct}      & 43.83          & \textbf{26.02} & 26.93          & 39.36          & 40.22          & 41.31          & 38.48          & \multicolumn{1}{l}{} & \multicolumn{1}{l}{} \\
\textbf{AlignX (51langs)}        & \textbf{44.89} & 25.97          & \textbf{27.10} & \textbf{40.72} & \textbf{42.20} & \textbf{42.05} & \textbf{39.52} & \multicolumn{1}{l}{} & \multicolumn{1}{l}{} \\ \cline{1-8}
\end{tabular}
\caption{Detailed results on multilingual HellaSwag benchmark under the 51-language setup. We bold the highest scores.}
\label{detailed_m_hellaswag_51langs}
\end{table*}

\begin{table*}[htbp]
\centering
\small
\begin{tabular}{cccccccccc}
\hline
\textbf{XNLI} & \textbf{Ar}    & \textbf{De}    & \textbf{El}    & \textbf{En}    & \textbf{Es}    & \textbf{Fr}    & \textbf{Hi}    & \textbf{Ru}    & \textbf{Sw}    \\ \hline
\textbf{LLaMA3-8B-Instruct}      & 34.86          & 45.18          & \textbf{37.11} & 44.66          & 41.69          & 42.29          & 39.80          & \textbf{42.05} & \textbf{34.94} \\
\textbf{AlignX (51langs)}        & \textbf{36.59} & \textbf{46.10} & 36.18          & \textbf{45.86} & \textbf{44.86} & \textbf{44.54} & \textbf{42.21} & 41.16          & 34.66          \\ \hline
\multicolumn{1}{l}{}             & \textbf{Th}    & \textbf{Tr}    & \textbf{Ur}    & \textbf{Vi}    & \textbf{Zh}    & \textbf{Avg.}  &       &       &       \\ \hline
\textbf{LLaMA3-8B-Instruct}      & 39.28          & 40.72          & \textbf{37.51} & 38.55          & 40.52          & 39.94          &                &                &                \\
\textbf{AlignX (51langs)}        & \textbf{41.73} & \textbf{43.33} & 35.98          & \textbf{41.61} & \textbf{40.84} & \textbf{41.12} &       &       &       \\ \hline
\end{tabular}
\caption{Detailed results on XNLI benchmark under the 51-language setup. We bold the highest scores.}
\label{detailed_xnli_51langs}
\end{table*}

\begin{table*}[htbp]
\centering
\small
\begin{tabular}{ccccccc}
\hline
\textbf{XStoryCloze}        & \textbf{Ar}    & \textbf{En}    & \textbf{Es}    & \textbf{Hi}    & \textbf{Id}    & \textbf{My}    \\ \hline
\textbf{LLaMA3-8B-Instruct} & 61.75          & 80.61          & 72.60          & \textbf{65.25} & 67.31          & \textbf{50.10} \\
\textbf{AlignX (51langs)}   & \textbf{63.47} & \textbf{81.34} & \textbf{73.13} & 64.39          & \textbf{68.56} & 49.04          \\ \hline
                            & \textbf{Ru}    & \textbf{Te}    & \textbf{Zh}    & \textbf{Avg.}  &                &                \\ \hline
\textbf{LLaMA3-8B-Instruct} & 71.01          & 61.02          & 67.90          & 66.40          &                &                \\
\textbf{AlignX (51langs)}   & \textbf{72.47} & \textbf{61.09} & \textbf{69.23} & \textbf{66.97} &                &                \\ \hline
\end{tabular}
\caption{Detailed results on XStoryCloze benchmark under the 51-language setup. We bold the highest scores.}
\label{detailed_xstorycloze_51langs}
\end{table*}

\begin{table*}[htbp]
\centering
\small
\begin{tabular}{cc}
\hline
\textbf{System Comparison}                  & \textbf{Statistical Significance} \\ \hline
\textbf{Gemma-2B v.s. AlignX}            & 84 / 90                     \\
\textbf{Mistral-7B-v0.3 v.s. AlignX}            & 70 / 75                     \\
\textbf{LLaMA-7B v.s. AlignX}           & 81 / 86                     \\
\textbf{LLaMA2-7B v.s. AlignX}          & 90 / 90                     \\
\textbf{LLaMA3-8B-Instruct v.s. AlignX} & 41 / 55                     \\ 
\textbf{LLaMA3-8B-Instruct v.s. AlignX (51langs)} & 79 / 87 \\ \hline
\end{tabular}
\caption{Statistical significance tests for multilingual translations. We present the results in the \textit{a / b} format, indicating that AlignX outperforms corresponding base LLMs in \textit{b} translation directions, where the improvements in \textit{a} translation directions are significant (p\_value < 0.05).}
\label{table_translation_significance}
\end{table*}
\begin{table*}[htbp]
\centering
\small
\begin{tabular}{cccccccccccc}
\hline
\textbf{Methods}             & \textbf{X-En}  & \textbf{X-Ar}  & \textbf{X-Cs}  & \textbf{X-De}  & \textbf{X-El}  & \textbf{X-Hi}  & \textbf{X-Ru}  & \textbf{X-Tr}  & \textbf{X-Vi}  & \textbf{X-Zh}  & \textbf{Avg.}  \\ \hline
\multicolumn{12}{c}{\textbf{Gemma-2B Based}}                                                                                                                                                                           \\ \hline
\textbf{Gemma-2B}           & 84.55          & 58.84          & 65.20          & 65.51          & 61.16          & 47.02          & 58.03          & 59.54          & 66.23          & 68.94          & 63.50          \\
\textbf{CPT-then-SFT}           & \textbf{85.39} & 66.02 & 71.24 & 68.44 & \textbf{75.27} & \textbf{71.04} & \textbf{73.17} & \textbf{74.44} & 72.29 & 53.28 & 71.06 \\
\textbf{AlignX}          & 85.29 & \textbf{67.31} & \textbf{72.18} & \textbf{73.44} & 70.88 & 53.34 & 71.07 & 67.10 & \textbf{76.59} & \textbf{75.71} & \textbf{71.29} \\ \hline
\multicolumn{12}{c}{\textbf{Mistral-7B-v0.3 Based}}                                                                                                                                                                           \\ \hline
\textbf{Mistral-7B-v0.3} & 84.48 & 55.65 & 75.34 & 55.99 & 74.30 & 48.51 & 75.88 & 64.37 & 75.44 & 39.13 & 64.91 \\
\textbf{CPT-then-SFT} & \textbf{86.12} & 67.44 & \textbf{78.00} & 67.01 & \textbf{78.79} & 62.68 & \textbf{77.28} & 72.97 & \textbf{82.13} & 45.18 & 71.76  \\
\hline
\end{tabular}
\caption{The averaged COMET~\cite{rei-etal-2020-comet} scores on FLORES-101. "X" denotes all other training languages except the target language. "Avg." denotes average scores on all translation directions. We bold the highest scores.}
\label{table_main_translation_comet}
\end{table*}
\begin{table*}[htbp]
\centering
\small
\begin{tabular}{c|cccccccccc|c}
\hline
\textbf{Gemma-2B} & \textbf{En} & \textbf{Ar} & \textbf{Cs} & \textbf{De} & \textbf{El} & \textbf{Hi} & \textbf{Ru} & \textbf{Tr} & \textbf{Vi} & \textbf{Zh} & \textbf{Avg.} \\ \hline
\textbf{En}       & -           & 13.18       & 20.98       & 26.93       & 13.48       & 16.34       & 22.70       & 14.71       & 29.78       & 18.80       & 19.66         \\
\textbf{Ar}       & 29.79       & -           & 4.28        & 6.26        & 3.19        & 1.43        & 1.89        & 1.57        & 6.01        & 4.69        & 6.57          \\
\textbf{Cs}       & 34.67       & 5.08        & -           & 10.62       & 3.99        & 0.39        & 7.37        & 3.23        & 2.86        & 7.35        & 8.40          \\
\textbf{De}       & 40.67       & 5.55        & 6.71        & -           & 5.94        & 5.42        & 4.21        & 4.37        & 8.65        & 10.00       & 10.17         \\
\textbf{El}       & 29.11       & 1.16        & 3.37        & 5.85        & -           & 0.73        & 0.91        & 1.18        & 4.51        & 3.11        & 5.55          \\
\textbf{Hi}       & 27.36       & 0.60        & 2.70        & 4.00        & 1.01        & -           & 1.00        & 2.55        & 2.97        & 4.72        & 5.21          \\
\textbf{Ru}       & 31.54       & 4.08        & 9.18        & 10.39       & 6.97        & 2.69        & -           & 3.18        & 6.69        & 8.07        & 9.20          \\
\hline
\end{tabular}

\caption{Detailed BLEU scores of the FLORES-101 benchmark on Gemma-2B.}
\label{detailed_translation_gemma}
\end{table*}

\begin{table*}[htbp]
\small
\centering
\begin{tabular}{c|cccccccccc|c}
\hline
\textbf{Mistral-7B-v0.3} & \textbf{En} & \textbf{Ar} & \textbf{Cs} & \textbf{De} & \textbf{El} & \textbf{Hi} & \textbf{Ru} & \textbf{Tr} & \textbf{Vi} & \textbf{Zh} & \textbf{Avg.} \\ \hline
\textbf{En}              & -           & 6.81        & 27.30       & 32.95       & 4.24        & 6.50        & 28.21       & 11.44       & 17.66       & 17.92       & 17.00         \\
\textbf{Ar}              & 27.37       & -           & 9.83        & 13.82       & 1.54        & 2.06        & 9.39        & 2.71        & 7.27        & 6.00        & 8.89          \\
\textbf{Cs}              & 39.36       & 4.44        & -           & 24.67       & 2.94        & 3.60        & 23.31       & 5.81        & 9.89        & 12.60       & 14.07         \\
\textbf{De}              & 43.69       & 4.88        & 22.01       & -           & 2.88        & 4.17        & 23.23       & 7.58        & 12.35       & 13.55       & 14.93         \\
\textbf{El}              & 26.20       & 3.08        & 10.98       & 14.53       & -           & 2.04        & 12.24       & 3.67        & 7.65        & 6.55        & 9.66          \\
\textbf{Hi}              & 21.61       & 1.94        & 5.76        & 8.91        & 1.04        & -           & 5.69        & 3.73        & 5.04        & 5.73        & 6.61          \\
\textbf{Ru}              & 35.88       & 4.80        & 20.29       & 22.52       & 3.11        & 3.97        & -           & 6.53        & 11.75       & 12.59       & 13.49         \\
\hline

\end{tabular}

\caption{Detailed BLEU scores of the FLORES-101 benchmark on Mistral-7B-v0.3.}
\label{detailed_translation_mistral}
\end{table*}

\begin{table*}[htbp]
\centering
\small
\begin{tabular}{c|cccccccccc|c}
\hline
\textbf{LLaMA-7B} & \textbf{En} & \textbf{Ar} & \textbf{Cs} & \textbf{De} & \textbf{El} & \textbf{Hi} & \textbf{Ru} & \textbf{Tr} & \textbf{Vi} & \textbf{Zh} & \textbf{Avg.} \\ \hline
\textbf{En}       & -           & 2.40        & 19.71       & 27.24       & 2.16        & 2.71        & 21.16       & 3.85        & 3.08        & 5.65        & 9.77          \\
\textbf{Ar}       & 13.56       & -           & 0.77        & 3.00        & 0.10        & 0.10        & 1.41        & 0.17        & 0.22        & 0.15        & 2.16          \\
\textbf{Cs}       & 34.83       & 1.17        & -           & 16.94       & 1.19        & 1.49        & 14.86       & 1.55        & 1.45        & 1.52        & 8.33          \\
\textbf{De}       & 40.18       & 1.44        & 11.38       & -           & 1.45        & 1.79        & 13.55       & 2.31        & 1.97        & 2.86        & 8.55          \\
\textbf{El}       & 16.40       & 0.14        & 1.73        & 4.85        & -           & 0.20        & 2.77        & 0.28        & 0.45        & 0.45        & 3.03          \\
\textbf{Hi}       & 9.89        & 0.09        & 0.59        & 2.29        & 0.11        & -           & 0.80        & 0.25        & 0.15        & 0.06        & 1.58          \\
\textbf{Ru}       & 32.22       & 1.10        & 11.29       & 14.84       & 1.27        & 1.40        & -           & 0.90        & 0.99        & 2.71        & 7.41          \\
\hline \hline

\end{tabular}

\caption{Detailed BLEU scores of the FLORES-101 benchmark on LLaMA-7B (part 1).}
\label{detailed_translation_llama_1}
\end{table*}

\begin{table*}[htbp]
\centering
\small
\begin{tabular}{c|cccccccccc|c}
\hline

\textbf{CPT-then-SFT} & \textbf{En} & \textbf{Ar} & \textbf{Cs} & \textbf{De} & \textbf{El} & \textbf{Hi} & \textbf{Ru} & \textbf{Tr} & \textbf{Vi} & \textbf{Zh} & \textbf{Avg.} \\ \hline
\textbf{En}           & -           & 10.30       & 24.08       & 28.80       & 8.22        & 7.93        & 23.71       & 11.49       & 15.14       & 12.37       & 15.78         \\
\textbf{Ar}           & 23.79       & -           & 6.01        & 7.70        & 0.99        & 0.66        & 6.44        & 2.06        & 3.38        & 2.80        & 5.98          \\
\textbf{Cs}           & 36.37       & 4.49        & -           & 19.09       & 4.68        & 3.38        & 18.49       & 5.82        & 7.65        & 7.81        & 11.98         \\
\textbf{De}           & 41.17       & 4.49        & 14.53       & -           & 5.38        & 3.68        & 16.39       & 7.13        & 8.14        & 8.48        & 12.15         \\
\textbf{El}           & 24.94       & 0.61        & 2.48        & 3.98        & -           & 0.15        & 4.24        & 2.03        & 3.50        & 2.71        & 4.96          \\
\textbf{Hi}           & 17.95       & 0.49        & 3.99        & 5.45        & 0.16        & -           & 4.09        & 1.82        & 2.21        & 1.54        & 4.19          \\
\textbf{Ru}           & 33.53       & 3.84        & 16.94       & 17.33       & 4.45        & 3.00        & -           & 5.55        & 6.92        & 7.20        & 10.97         \\
\hline

\end{tabular}

\caption{Detailed BLEU scores of the FLORES-101 benchmark on LLaMA-7B (part 2).}
\label{detailed_translation_llama_2}
\end{table*}

% llama2
\begin{table*}[htbp]
\centering
\small
\begin{tabular}{c|cccccccccc|c}
\hline
\textbf{LLaMA2-7B} & \textbf{En} & \textbf{Ar} & \textbf{Cs} & \textbf{De} & \textbf{El} & \textbf{Hi} & \textbf{Ru} & \textbf{Tr} & \textbf{Vi} & \textbf{Zh} & \textbf{Avg.} \\ \hline
\textbf{En}        & -           & 3.76        & 23.28       & 30.07       & 3.01        & 4.20        & 22.54       & 6.87        & 24.64       & 14.65       & 14.78         \\
\textbf{Ar}        & 16.20       & -           & 1.93        & 4.62        & 0.14        & 0.05        & 1.08        & 0.33        & 2.62        & 0.39        & 3.04          \\
\textbf{Cs}        & 37.30       & 1.73        & -           & 20.18       & 1.65        & 2.16        & 16.04       & 2.49        & 10.35       & 7.96        & 11.10         \\
\textbf{De}        & 41.71       & 1.99        & 17.75       & -           & 1.98        & 2.32        & 16.12       & 4.35        & 17.50       & 9.61        & 12.59         \\
\textbf{El}        & 16.96       & 0.17        & 2.14        & 4.93        & -           & 0.14        & 1.58        & 0.49        & 2.65        & 0.38        & 3.27          \\
\textbf{Hi}        & 11.31       & 0.05        & 1.87        & 3.20        & 0.11        & -           & 1.04        & 0.71        & 1.43        & 0.27        & 2.22          \\
\textbf{Ru}        & 33.69       & 1.69        & 16.27       & 17.95       & 1.79        & 1.86        & -           & 2.27        & 14.16       & 7.74        & 10.82         \\
\hline
\end{tabular}

\caption{Detailed BLEU scores of the FLORES-101 benchmark on LLaMA2-7B.}
\label{detailed_translation_llama2}
\end{table*}

% llama3-8b-instruct
\begin{table*}[htbp]
\centering
\small
\begin{tabular}{c|cccccccccc|c}
\hline
\textbf{LLaMA3-8B-Instruct} & \textbf{En} & \textbf{Ar} & \textbf{Cs} & \textbf{De} & \textbf{El} & \textbf{Hi} & \textbf{Ru} & \textbf{Tr} & \textbf{Vi} & \textbf{Zh} & \textbf{Avg.} \\ \hline
\textbf{En}                 & -           & 19.84       & 29.14       & 34.97       & 22.59       & 24.18       & 29.12       & 24.19       & 34.37       & 22.71       & 26.79         \\
\textbf{Ar}                 & 34.30       & -           & 17.84       & 21.34       & 15.69       & 13.92       & 20.16       & 14.14       & 23.71       & 14.93       & 19.56         \\
\textbf{Cs}                 & 39.19       & 14.84       & -           & 26.63       & 17.30       & 16.78       & 24.19       & 16.65       & 24.76       & 16.88       & 21.91         \\
\textbf{De}                 & 42.08       & 15.43       & 23.80       & -           & 18.67       & 18.60       & 24.54       & 18.44       & 26.25       & 18.41       & 22.91         \\
\textbf{El}                 & 34.10       & 13.76       & 19.31       & 22.65       & -           & 14.38       & 21.03       & 15.08       & 24.36       & 16.07       & 20.08         \\
\textbf{Hi}                 & 33.51       & 12.02       & 15.59       & 19.43       & 13.87       & -           & 17.10       & 15.59       & 21.47       & 14.84       & 18.16         \\
\textbf{Ru}                 & 34.68       & 14.57       & 21.57       & 23.66       & 16.85       & 15.93       & -           & 15.84       & 24.68       & 17.28       & 20.56         \\
\hline

\end{tabular}

\caption{Detailed BLEU scores of the FLORES-101 benchmark on LLaMA3-8B-Instruct.}
\label{detailed_translation_llama3instruct}
\end{table*}
\begin{table*}[htbp]
\centering
\small
\begin{tabular}{cccccccccc}
\hline
\textbf{FLORES-101}       & \textbf{X-Af}  & \textbf{X-Ar}  & \textbf{X-Az}  & \textbf{X-Bn}  & \textbf{X-Cs}  & \textbf{X-De}  & \textbf{X-El}  & \textbf{X-En}  & \textbf{X-Et}  \\ \hline
\textbf{LLaMA3-8B-Instruct}   & 16.64          & 11.71          & \textbf{7.47}  & 8.10           & 16.21          & 19.48          & 14.15          & 30.30          & 10.65          \\
\textbf{AlignX (51langs)} & \textbf{16.67} & \textbf{13.75} & 7.18           & \textbf{8.41}  & \textbf{17.70} & \textbf{21.29} & \textbf{16.02} & \textbf{34.87} & \textbf{12.77} \\ \hline
                 & \textbf{X-Fa}  & \textbf{X-Fi}  & \textbf{X-Fr}  & \textbf{X-Gl}  & \textbf{X-Gu}  & \textbf{X-He}  & \textbf{X-Hi}  & \textbf{X-Hr}  & \textbf{X-Id}  \\ \hline
\textbf{LLaMA3-8B-Instruct}   & 14.30          & 12.80          & 26.06          & 18.12          & 4.10           & 12.30          & 14.97          & 13.39          & 19.37          \\
\textbf{AlignX (51langs)} & \textbf{14.92} & \textbf{13.06} & \textbf{28.29} & \textbf{19.47} & \textbf{4.48}  & \textbf{13.19} & \textbf{16.60} & \textbf{14.70} & \textbf{21.25} \\ \hline
                          & \textbf{X-It}  & \textbf{X-Ja}  & \textbf{X-Ka}  & \textbf{X-Kk}  & \textbf{X-Km}  & \textbf{X-Ko}  & \textbf{X-Lv}  & \textbf{X-Lt}  & \textbf{X-Ml}  \\ \hline
\textbf{LLaMA3-8B-Instruct}   & 19.11          & 16.27          & \textbf{2.79}  & \textbf{8.00}  & 0.76           & 12.45          & 10.64          & 10.35          & 3.01           \\
\hline
\end{tabular}
\caption{The averaged BLEU scores on FLORES-101 under the 51-language setup. "X" denotes all other training languages except the target language. "Avg." denotes average scores on all translation directions. We bold the highest scores.}
\label{table_brief_translation_51langs}
\end{table*}

\end{document}